\documentclass[final]{article}

\usepackage{neurips_2025}

\usepackage[utf8]{inputenc} 
\usepackage[T1]{fontenc}    
\usepackage{hyperref}       
\usepackage{url}            
\usepackage{booktabs}       
\usepackage{amsfonts}       
\usepackage{nicefrac}       
\usepackage{microtype}      
\usepackage{xcolor}         
\usepackage{graphicx}
\usepackage{booktabs} 
\usepackage{amssymb}  
\usepackage{array}    
\usepackage{caption} 
\usepackage{multirow}
\usepackage{enumitem}
\usepackage{xcolor}
\usepackage{tikz}
\usetikzlibrary{fadings}
\usepackage{xcolor}
\usepackage{contour}
\usepackage{gradient-text}
\usepackage{subcaption} 
\usepackage{tabulary}
\usepackage{tabularx}
\usepackage{adjustbox}
\usepackage{wrapfig}
\usepackage{longtable}
\usepackage{xr}
\usepackage{algorithm}
\usepackage{algpseudocode}
\usepackage{makecell}
\newcommand{\InfoChartQA}{\gradientRGB{InfoChartQA}{27, 177, 171}{140, 118, 250}}
\usepackage[most]{tcolorbox}
\newcommand{\TotfactImages}{5,948}

\newcommand{\TextQA}{50,920}

\newcommand{\Charttype}{54}

\newcommand{\xie}[1]{#1}
\newcommand{\changjian}[1]{#1}
\newcommand{\minzhi}[1]{#1}
\newcommand{\fix}[1]{#1}

\usepackage{hyperref}
\usepackage{xr}
\usepackage{placeins}
\def \etal {{\emph{et al}.}}
\def \eg {{\emph{e.g}.}}

\title{\InfoChartQA: A Benchmark for Multimodal Question Answering on Infographic Charts}

\author{
Tianchi Xie\thanks{Equal contribution.} \\
Tsinghua University\\
xietc24@mails.tsinghua.edu.cn\\
\And
Minzhi Lin\footnotemark[1]\\
Tsinghua University\\
linmz21@mails.tsinghua.edu.cn\\
\And
Mengchen Liu\\ 
Meta\\
simon900314@outlook.com\\
\And
Yilin Ye\\
Hong Kong University of Science and Technology\\
yyebd@connect.ust.hk\\
\And
Changjian Chen\thanks{Corresponding author.}\\
Hunan University\\
changjianchen@hnu.edu.cn\\
\And
Shixia Liu\\
Tsinghua University\\
shixia@tsinghua.edu.cn\\
}

\begin{document}
\maketitle

\begin{abstract}
Understanding infographic charts with pictorial visual elements (\eg, pictograms and icons) requires both visual recognition and reasoning, posing challenges for multimodal large language models (MLLMs).
However, existing visual question answering benchmarks fall short in evaluating these capabilities of MLLMs due to the lack of paired plain charts and visual-element-based questions.
To bridge this gap, we introduce \InfoChartQA{}, a benchmark for evaluating MLLMs on infographic chart understanding.
It includes \TotfactImages{} pairs of infographic and plain charts, each sharing the same underlying data but differing in visual presentations.
We further design visual-element-based questions to capture their unique visual designs and communicative intent.
Evaluation of 20 MLLMs reveals a substantial performance decline on infographic charts, particularly for visual-element-based questions related to metaphors.
The paired infographic and plain charts enable fine-grained error analysis and ablation studies, which highlight new opportunities for advancing MLLMs in infographic chart understanding. We release \InfoChartQA{} at \href{https://github.com/thu-vis/InfoChartQA}{https://github.com/thu-vis/InfoChartQA}.

\end{abstract}

\section{Introduction}
Infographic charts enrich standard chart types such as bar, pie, and line charts by integrating pictorial visual elements such as pictograms, thematic icons, and metaphorical imagery. 
These elements serve not only to convey data but also to enhance visual engagement, reinforce the chart's narrative or emotional tone, and communicate abstract concepts through symbolic visuals. 
Unlike plain charts that present data in a neutral and standardized way, infographic charts often adopt creative pictorial visual elements that reflect their communicative intent. 
As a result, understanding infographic charts requires more than basic visual recognition.
It demands reasoning about heterogeneous visual elements, symbolic metaphors, and the underlying data relationships. 
This poses new challenges for multimodal large language models (MLLMs), whose ability to integrate visual and textual information is still under development. 
A comprehensive benchmark is therefore needed to enable systematic evaluation and guide model improvement, capturing the unique features of infographic charts and supporting controlled comparisons with plain charts.

\begin{figure}[t]
    \centering
    \includegraphics[width=1\textwidth]{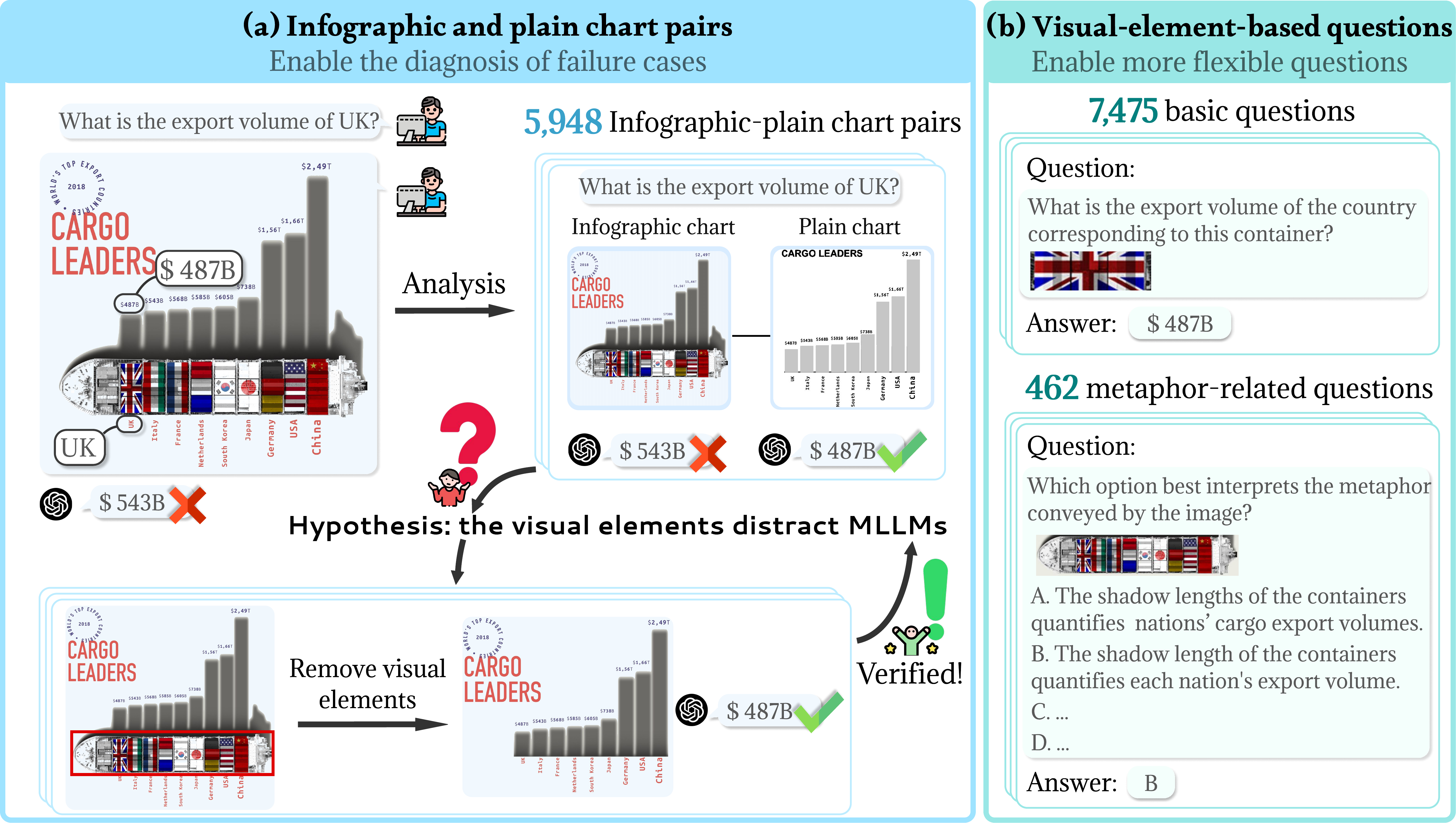
    }
    \caption{Overview of \InfoChartQA{}.}
    \label{fig:teaser}
    \vspace{-1em}
\end{figure}

Many visual question answering benchmarks have been developed to assess the capabilities of MLLMs to jointly understand and reason over both visual and textual information~\cite{masry2022chartqa,mathew2021infographicvqa}. 
However, existing benchmarks face two limitations when it comes to evaluating infographic chart understanding.
First, they lack paired infographic charts and plain chart counterparts constructed from the same underlying data. 
Such pairs are essential for disentangling whether a model's failure stems from the complexity of the data itself or from the additional visual elements used in infographic designs.
For example, the MLLM in Figure~\ref{fig:teaser}(a) answers wrongly on the infographic chart but correctly on the associated plain chart.
By removing the ship image in the infographic chart, the MLLM answers correctly, indicating that the ship image was the main cause of the MLLM's incorrect answer.
Second, most benchmarks do not include visual-element-based questions that specifically target the visual elements in infographic charts, such as pictograms, thematic icons, and metaphorical imagery (\eg, the flag and ship elements in Figure~\ref{fig:teaser}(b)). 
These visual elements are often crucial for conveying data (\eg, the associated value of an icon) or high-level \minzhi{semantics} metaphors (\eg, the metaphor conveyed by the ship).
The absence of such visual-element-based questions limits the benchmarks’ ability to capture the challenges posed by infographic-specific design.



To address these two limitations, we built \InfoChartQA{}, a benchmark for multimodal question answering on infographic charts. 
\InfoChartQA{} comprises \TotfactImages{} paired infographic and plain charts, where each pair shares the same underlying data but differs in visual representation (Figure~\ref{fig:teaser}). 
We built this dataset by collecting a high-quality set of infographic charts, extracting their underlying tabular data, and creating corresponding plain chart counterparts.
These paired charts enable the creation of shared questions based on textual descriptions and tabular data. 
ion to these shared questions, we also design visual-element-based questions.
Such questions include basic ones that target the understanding of visual elements commonly used in infographic charts, and metaphor-related ones that reflect the higher-level semantics conveyed through visual elements.

We conduct a comprehensive evaluation of 6 proprietary and 14 open-source MLLMs on the \InfoChartQA{}. 
The results indicate a significant performance decline
on infographic charts compared with plain charts.
For example, Claude 3.5 Sonnet scores \xie{81.37}\% on plain charts but only \xie{62.80}\% on infographic charts. 
MLLMs perform even poorer (\eg, Claude 3.5 Sonnet only scores 55.33\%) on metaphor-related questions.
The paired infographic and plain charts allow us to diagnose this poor performance through detailed error analysis and ablation studies.
The analysis shows that the proximity between icons and corresponding data values plays a critical role in supporting accurate reasoning.
Moreover, model accuracy tends to decline as visual complexity increases, particularly when more visual elements are present in the infographic chart.
These findings highlight new opportunities for advancing MLLMs in infographic chart question answering.

The key contributions of this paper are: 
\begin{itemize}[noitemsep, topsep=0pt] 
\item We present \InfoChartQA{}, the first benchmark containing paired infographic and plain charts that share the same underlying data but differ in visual representation. 
\item We introduce a rich set of visual-element-based QAs specifically designed for infographic charts to capture their unique visual elements and intended purpose. 
\item We identify and analyze the performance gap of current MLLMs when interpreting infographic charts versus plain charts, despite both being derived from the same data.
\end{itemize}

\section{Related Works}


Many benchmarks have been developed for chart question answering (QA)~\cite{masry2022chartqa,mathew2021infographicvqa,masry2025chartqapro,wang2024charxiv, li2025chartgalaxy}.
According to the types of charts, they can be categorized into plain and infographic chart QA benchmarks.

\textbf{Plain chart QA benchmarks}.
An initial benchmark along this line is FigureQA~\cite{kahou2017figureqa}.
FigureQA synthesized 100,000 charts across five types and generated one million binary questions based on 15 predefined templates, where answers are either "Yes" or "No". 
Subsequently, DVQA expanded the answer options to a fixed vocabulary of 1,000 words or extracted text from the charts~\cite{Kafle2018dvqa}. 
Additionally, the question templates were extended to 74, derived from 7,000 crowd-sourced questions~\cite{Methani2020plotqa}.
Since the synthesized charts and generated questions from templates cannot represent the real-world charts well, later efforts shifted toward collecting real-world charts with open-ended questions. 
OpenCQA collected 7,724 charts from Pew Research (\href{pewresearch.org}{pewresearch.org}) and asked crowdworkers from Amazon Mechanical Turk to create open-ended questions and answers~\cite{kantharaj2022opencqa}. 
ChartQA gathered 20,882 charts from four distinct online sources, along with human-authored QA pairs created through Amazon Mechanical Turk~\cite{masry2022chartqa}.
Since OpenCQA and ChartQA primarily focus on three chart types, ChartBench extended them to nine chart types, resulting in a total of 2,100 charts~\cite{xu2023chartBench}.
Later efforts have been dedicated to collecting more diverse charts and more complex questions.
ChartX covers 18 chart types and questions from 22 disciplinary topics~\cite{xia2025chartx}. 
ChartXiv includes 2,323 real-world charts selected from scientific papers across eight primary subjects published on arXiv~\cite{wang2024charxiv}.
ChartInsights found that most benchmarks focus on high-level chart QA tasks, with less attention given to low-level tasks, leading them to collect 2,000 charts and 22,000 QA for low-level chart QA tasks~\cite{wu2024chartinsights}.

Although these plain chart QA benchmarks are effective in evaluating MLLMs, they overlook infographic charts, which are an important category of charts with the composition of data and \fix{pictorial visual} elements presenting unique challenges to visual understanding and reasoning.
In response, infographic chart QA benchmarks have been proposed.

\textbf{Infographic chart QA benchmarks}.
The first benchmark in this category is InfographicVQA~\cite{mathew2021infographicvqa}, which consists of 30,035 questions across 5,485 infographic charts. 
The questions in this dataset are based on tables, figures, and visualizations, as well as those that require combining multiple cues. 
This makes it particularly challenging for MLLMs.
ChartQAPro~\cite{masry2025chartqapro} contains 1,341 charts from 157 diverse online sources, including 190 infographic charts. 
It features 1,948 questions in various formats, such as multiple-choice, conversational, hypothetical, and unanswerable questions, to better reflect real-world challenges.

Although these benchmarks collect a large number of infographic charts, they do not include the associated plain charts.
These plain charts, which display the same data in simpler visual forms, are crucial for diagnosing the root causes behind the failure of MLLMs.
Moreover, an important characteristic of infographic charts is that they convey rich information by combining a variety of visual elements \cite{yang2024foundation}.
However, the existing benchmarks do not provide such QAs specifically designed to evaluate the understanding of the visual elements in infographic charts.
To fill these gaps, we developed \InfoChartQA, a benchmark for multimodal QAs that includes pairs of infographics and plain charts, covering both data-fact-based and visual-element-based questions.








\section{The \InfoChartQA{} Benchmark}
\begin{figure}[t]
    \centering
    \includegraphics[width=1\textwidth]{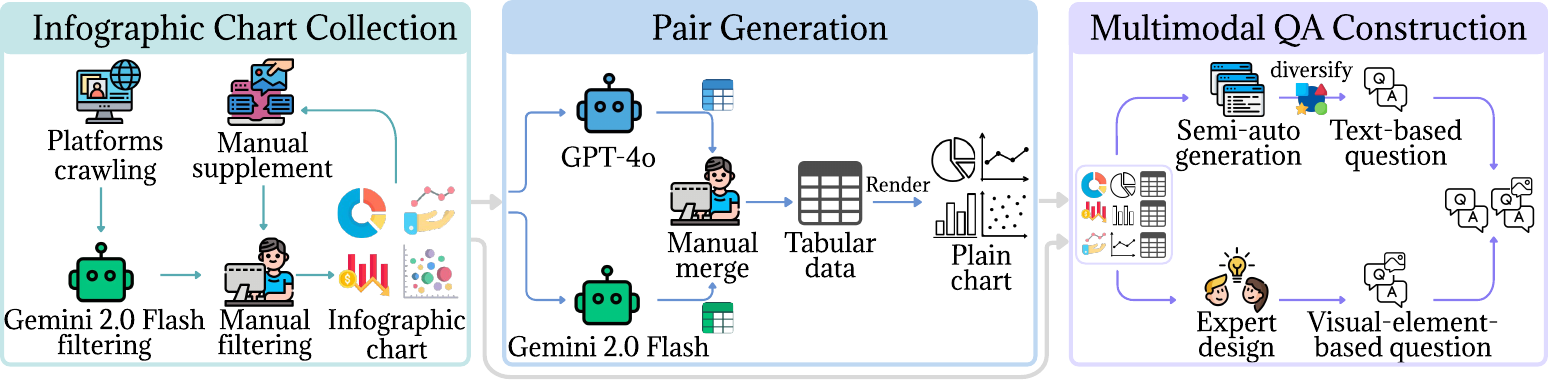}
    \caption{The \InfoChartQA{} benchmark construction pipeline.}
    \label{fig:overview}
\end{figure}

The \InfoChartQA{} benchmark is constructed by three main steps: infographic chart dataset construction, paired infographic and plain chart generation, and multimodal question and answer construction (Figure~\ref{fig:overview}). 
First, the infographic chart dataset construction step collects a diverse set of infographic charts. 
Next, the paired infographic and plain chart generation step creates the corresponding plain chart for each infographic chart.
Finally, the multimodal question and answer construction step creates both text-based and visual-element-based questions that focus on data-related facts and the interpretation of visual elements.

\subsection{Infographic Chart Dataset Construction}
\textbf{Infographic chart source.}
\InfoChartQA{} is collected from 11 real-world mainstream visualization platforms, such as Pinterest, Visual Capitalist, Statista, Behance, and iStock, \changjian{as well as a large-scale infographic chart dataset, ChartGalaxy~\cite{li2025chartgalaxy}}. 
For platforms with high data quality, such as Statista and Visual Capitalist, we collected all publicly available infographic charts up to March 2025. 
For platforms with varying data quality, such as Pinterest and iStock, we manually selected high-quality infographic charts as seeds and utilized the recommendation systems of the associated platforms for identifying more infographic charts.
\changjian{For ChartGalaxy, we selected several high-quality infographic charts from it, following the recommendations of its authors.  }

\textbf{Chart type identification}.
\changjian{To ensure the diversity of the collected infographic charts, one practical way is to ensure the collected charts encompass all major chart types.} 
Although many existing studies~\cite{xu2023chartBench,xia2025chartx,Han2023ChartLlamaAM,Meng2024ChartAssisstantAU} classify charts into around 10 coarse-grained types, large visual differences persist within each type. 
For example, the radial bar chart and polar bar chart are both considered bar charts, yet they differ substantially in appearance.
Therefore, we invited three visualization experts to identify more fine-grained chart types.
Specifically, we first derived a set of over 150 potential types from the Data Viz Project~\cite{datavizproject}.
However, we found that some of these types were not commonly used in infographic charts, such as multi-level donut charts.
Therefore, we used the name of each type to search for infographic charts in all 11 visualization platforms mentioned above.
If the total number of searched infographic charts of one type in all 11 platforms was less than 10, and the visualization experts believed it was not common in infographic charts, we removed it from our benchmark.
\changjian{During our search, we found that, although rare, some infographic charts contain multiple panels (sub-charts)~\cite{liu2019interative, chen2020oodanalyzer}. 
For such charts, questions involving cross-panel reasoning are more challenging than those for single-panel charts. 
To highlight these more challenging cases, we added two data types of multi-panel charts: homogeneous ones, where all panels share the same chart type, and heterogeneous ones, where panels belong to different chart types.}
Finally, a total of \Charttype{} chart types were identified, with details shown in Appendix \ref{ssec:types}.

\textbf{Infographic chart selection}.
Since the \changjian{majority of the} infographic charts were crawled from various platforms automatically, some irrelevant data, such as diagrams, illustrations, and natural images, were also included.
Moreover, the number of infographic charts for certain chart types was limited, which led to an imbalanced benchmark.
To mitigate the low quality and imbalance issues, we developed a semi-automatic selection pipeline. 
First, we applied Gemini 2.0 Flash, one of the most powerful MLLMs, to identify infographic chart candidates.
The prompts can be found in Appendix \ref{ssec:selection_prompt}.
Then, we recruited two experienced graduate students to select infographic charts from the candidate set.
After the selection, we analyzed the distribution of the infographic charts by the \Charttype{} chart types.
For each chart type, if the number of the associated infographic charts was less than 30, we used the chart type name to search for additional infographic charts on the platforms.
The newly added infographic charts were also processed through the semi-automatic pipeline. 
This process was repeated until the number of charts for each type exceeds 30.
The final dataset comprises \TotfactImages{} infographic charts.

\subsection{Paired Infographic and Plain Chart Generation}
Infographic charts enrich plain charts with rich \fix{pictorial} visual elements to better convey information and metaphors. 
Comparing the performance of MLLMs in understanding these two types of charts can provide deeper insights into their visual recognition capabilities.
Therefore, we generate the corresponding plain chart for each infographic chart.
The generation consists of two steps: chart-to-table translation and plain chart rendering.

\textbf{Chart-to-table translation}.
Since only a few platforms provide the associated tabular data for infographic charts, we utilize chart-to-table translation to extract the associated tabular data from the infographic charts.
To ensure more reliable table extraction, we ensembled two MLLMs \changjian{and invited four experts for verification}.
Specifically, for each infographic chart, we employed both Gemini 2.0 Flash and GPT-4o to extract the associated tabular data.
\changjian{Then, the experts merged the tables extracted by the two models and corrected any errors they found to ensure accuracy.}

\textbf{Plain chart rendering}.
Once the tabular data is extracted, the associated plain charts can be rendered easily according to their chart types.
For example, for the vertical bar chart, we directly utilize APIs in Python, including plotly, matplotlib, and seaborn, for rendering when the tabular data is given.

\subsection{Multimodal Question and Answer Construction}

\begin{table}[b]
\footnotesize
\centering
\caption{Comparison of \InfoChartQA{} and existing benchmarks.
}
\resizebox{\textwidth}{!}{
\normalsize
\begin{tabular}{lcccccc}
  \toprule
  \multirow{2}{*}{\textbf{Dataset}} &
  \multirow{2}{*}{\textbf{Chart type}} &
  \multirow{2}{*}{\textbf{\makecell{Infographic\\charts}}} &
  \multirow{2}{*}{\textbf{\makecell{ Text-based\\questions}}} &
  \multirow{2}{*}{\textbf{\makecell{Visual-element-based\\questions}}} &
  \multirow{2}{*}{\textbf{HD-D}} &
  \multirow{2}{*}{\textbf{SD}} \\
  & & & & & & \\
  \midrule 
 ChartQA  & 3 & $\times$ & 2.5K & $\times$ & 0.769	& 0.805\\
 ChartBench  & 42 & $\times$ & 16.8K & $\times$ & 0.630	& 0.743\\
 ChartQAPro  & 9 &  $\checkmark$ & 1.9K & $\times$ & 	0.828	& 0.864\\
 InfographicVQA  & 11 & $\checkmark$ & 3.2K &  1.1K & \textbf{0.837} & \textbf{0.823}\\
 \toprule 
 \InfoChartQA{}  & \textbf{\Charttype{}} & $\checkmark$ & \textbf{50.9K}  &  \textbf{7.9K} & 0.817 & 0.802\\
   \bottomrule
\end{tabular}
}

\label{tab:dataset}

\end{table}


We construct the multimodal question and answer pairs by incorporating generic text-based questions, which are shared between plain and infographic charts, as well as visual-element-based questions unique to infographic charts, as shown in Table~\ref{tab:dataset}.

\textbf{Text-based questions}.
We curate high-quality text-based questions to facilitate comparative analysis of MLLMs' performance on infographic charts and their plain chart counterparts.
The questions of existing chart understanding benchmarks are designed based on heuristics or experience, which may not ensure that all the information conveyed by the chart is covered.
To address this issue, we propose using data facts to guide the design of questions. 
Data facts refer to the numerical or statistical results that the chart is intended to convey.
According to the analysis by Wang~\etal~\cite{wang2020datashot}, there are eleven types of data facts: value, categorization, aggregation, extreme, rank, proportion, distribution, trend, difference, outlier, and association.
Different types of data facts may be suitable for different chart types.
For example, line charts are suitable for showing the trends of the data, but not for showing ranking results.
%

\changjian{Based on the data facts, we utilize a semi-automated method to ensure the difficulty and diversity of questions while minimizing human efforts.
Firstly, four visualization experts manually wrote 1,376 general questions based on 405 infographic samples, covering all chart types and data facts in the dataset.
Then,} for each infographic chart, we selected the suitable templates according to the chart type and data facts to generate questions and their answers, with more detail shown in Appendix \ref{ssec:template}.
\changjian{Finally, we employed Gemini-2.5-Flash and GPT-4o to rewrite all template questions using the experts’ questions as reference to ensure both difficulty and linguistic diversity.}

\xie{While the majority of questions can be reliably generated through our semi-automated \changjian{method}, we observed that the generated questions of multi-panel infographic charts tend to be inaccurate, especially the co-referential ones that require linking entities in different panels to answer.
Therefore, for multi-panel infographic charts, instead of using semi-automatically generated questions, we invited the four visualization experts to design 780 complex co-referential questions.
}

In total, we create \TextQA{} text-based questions for \Charttype{} different chart types.
As shown in Table~\ref{tab:dataset}, our text-based questions surpass existing benchmarks in both scale and chart type diversity, enabling a more comprehensive comparison.
\xie{\changjian{Moreover, our dataset demonstrates a comparable level of semantic diversity, as measured by the \textbf{S}emantic \textbf{D}iversity score~\cite{wang2023expectation} (\textbf{SD}), and vocabulary richness, as measured by the \textbf{H}ypergeometric \textbf{D}istribution-based \textbf{D}ivergence~\cite{masry2025chartqapro} (\textbf{HD-D}), to that of purely human-generated questions in existing benchmarks (\eg, InfographicVQA).}}
\textbf{Visual-element-based questions}.
The visual-element-based questions are a unique type of question we introduce for infographic charts to evaluate more sophisticated visual understanding and reasoning capabilities.
As shown in Figure~\ref{fig:teaser}(b), it includes basic questions and metaphor-related questions.
\begin{itemize}
    \item \textbf{Basic questions}. The basic questions enable intuitive reference to visual elements (\eg, the flag in Figure~\ref{fig:teaser}(b)) related to data even in the absence of text annotations.
    It is an extension of text-based questions by multimodal inputs with infographic-specific elements.
    To construct these questions, we first combine InternImage~\cite{interimage}, a SOTA detection model, with human verification to extract the visual elements. 
    Subsequently, two types of basic questions are derived.
    The first type (2,073 questions) asks about the correspondence between the visual element and the data item (\eg, Figure~\ref{fig:teaser}(b)).
    The second type (5,402 questions) examines the function of the visual elements (\eg, highlighting trends or conveying themes) through multiple-choice.
    \changjian{Similar to generating text-based questions, we generated the basic questions with templates and MLLMs in a semi-automated manner. The templates can be found in Appendix \ref{ssec:visual_sample}}.

    \item \textbf{Metaphor-related questions}. A more subtle type of visual-element-based question is the metaphor-related one.
    Specifically, the metaphor of infographic charts combines visual elements to convey narratives or evoke emotional responses.
    For example, as shown in Figure~\ref{fig:teaser}(b), the ship visual element conveys a metaphor for the export volume.
    Due to the challenge of recognizing metaphors within these charts, it is not feasible to generate metaphor-related questions automatically. 
    Therefore, we invited two visualization experts experienced in metaphor analysis for infographic charts.
    Initially, the two experts reviewed all the infographic charts and identified 143 that convey metaphors. 
    Then, each chart was annotated by one expert who designed metaphor-related questions and the corresponding answers in multiple-choice format.
    Since an infographic chart may contain multiple metaphors, the experts could provide one or more questions.
    After the annotation, the questions and answers of each infographic chart were reviewed by the other expert to ensure the correctness \xie{and neutrality (\eg, avoiding ambiguous or culturally sensitive interpretations).}
    Through this process, we obtained 462 metaphor-related questions.
\end{itemize}

In total, we construct over 7K visual-element-based questions, which is significantly more than those in InfographicVQA (Table~\ref{tab:dataset}).
We show more examples of such questions in Appendix \ref{append:sample}.

\section{Experiments}
\subsection{Experimental Setup}

\begingroup 
\renewcommand{\arraystretch}{0.29}
\begin{table}[!b]
\caption{Evaluation results on {\InfoChartQA} in terms of accuracy. \xie{The best one (except human) is \textbf{bold}, and the runner-up is \underline{underlined}.} Results with (*) are tested on a randomly sampled $10\%$ subset.}

\begin{tabularx}{\textwidth}{>{\arraybackslash}p{118pt}>{\centering\arraybackslash}p{30pt}>{\centering\arraybackslash}p{30pt}>{\centering\arraybackslash}p{30pt}p{3pt}>{\centering\arraybackslash}p{30pt}>{\centering\arraybackslash}p{30pt}>{\centering\arraybackslash}p{30pt}}
\toprule
  \rule[-0.4em]{0pt}{1.2em}\textbf{Model} & \multicolumn{3}{c}{\textbf{Text-based}} & &\multicolumn{3}{c}{\textbf{Visual-element-based}} \\
 \cline{2-4} \cline{6-8}
 &  \rule[-0.25em]{-10pt}{1.3em}\textbf{Infographic} & \textbf{Plain} & $\Delta$ & & \textbf{Basic} & \textbf{Metaphor}  & \textbf{Avg.}\\
\midrule
\multicolumn{8}{c}{\rule[-0.2em]{0pt}{0.8em}\textbf{Baselines}}\\
\midrule
Human                        &    $94.63^*$   &     $95.44^*$    & $0.81$ & & $92.89^*$    &  $88.69$ & $90.79$ \\
\midrule
\multicolumn{8}{c}{\rule[-0.2em]{0pt}{0.8em}\textbf{Proprietary Models}}\\
\midrule
OpenAI O4-mini               &   \underline{$76.23$}     &  \underline{$89.62$}  &  $13.39$ & & $\mathbf{91.42}$    &   $54.76$  & $\underline{73.09}$\\
GPT-4.1                      &   $71.29$     &  $80.81$  &  $9.52$  & & $87.52$    &   $50.87$  & $69.20$\\
GPT-4o                       &   $64.59$     &  $80.60$  &  $16.01$ & & $81.05$    &   $47.19$  & $64.12$\\
Claude 3.5 Sonnet            &   $62.80$     &  $81.37$  &  $18.57$ & & \underline{$89.22$}    &   $55.33$  & $72.28$\\
Gemini 2.5 Pro Preview       &   $\mathbf{79.23}$     &  $\mathbf{91.16}$  &  $11.93$ & & $88.91$    &   $\mathbf{60.42}$  & $\mathbf{74.67}$\\
Gemini 2.5 Flash Preview     &   $72.40$     &  $80.56$  &  $8.16$  & & $81.25$    &   \underline{$56.28$}  & $68.77$\\
\midrule
\multicolumn{8}{c}{\rule[-0.2em]{0pt}{0.8em}\textbf{Open-Source Models}} \\
\midrule
Qwen2.5-VL-72B             &     $61.08$  &   $77.92$   &  $16.84$ & &     $76.71$    & $54.64$  & $65.68$\\
Llama-4 Scout              &     $63.68$  &   $78.84$   &  $15.16$ & &     $81.69$    & $51.89$ & $66.79$ \\
Intern-VL3-78B             &    $63.42$   &   $81.41$   &  $17.99$ & &     $78.80$    & $51.52$ & $65.16$\\
Intern-VL3-8B              &    $46.45$   &   $61.67$   &  $15.22$ & &     $73.62$    & $49.57$ & $61.60$\\
Janus Pro                  &    $27.89$   &   $35.88$   &  $7.99$  & &     $41.22$    & $42.21$ & $41.72$\\
DeepSeek VL2               &    $40.40$   &   $44.44$   &  $4.04$  & &     $58.59$    & $44.54$ & $51.57$\\
Phi-4                      &    $35.47$   &   $54.68$   &  $19.21$ & &     $61.63$    & $38.31$ & $49.97$\\
LLaVA OneVision Chat 72B   &    $44.69$   &   $58.51$   &  $13.82$ & &     $61.82$    & $50.22$ & $56.02$\\
LLaVA OneVision Chat 7B    &    $36.45$   &   $50.47$   &  $14.02$ & &     $60.56$    & $45.67$ & $53.12$\\
Pixtral                    &    $46.61$   &   $59.29$   &  $12.68$ & &     $64.00$    & $50.87$ & $57.44$\\
Ovis1.6-Gemma2-9B          &    $51.69$   &   $58.66$   &  $6.97$  & &     $60.81$    & $34.42$ & $47.62$\\
ChartGemma                 &    $22.42$   &   $33.33$   &  $10.91$ & &     $30.75$    & $33.77$ & $32.26$\\
TinyChart                  &    $24.32$   &   $42.97$   &  $18.65$ & &     $15.35$    & $9.03$  & $12.19$\\
ChartInstruct-LLama2       &    $19.95$   &   $26.87$   &  $6.92$  & &     $34.15$    & $33.12$ & $33.64$\\
\bottomrule

\end{tabularx}
\label{tab:main_res}
\end{table}
\endgroup 

\textbf{Models.} 
We evaluated a diverse set of open-source and proprietary models on \InfoChartQA. 
For open-source models, we tested both general-purpose and domain-specific (in chart understanding) models, including: Qwen2.5-VL~\cite{bai2025qwen2}, Llama 4~\cite{meta2025llama4}, Intern-VL3~\cite{zhu2025internvl3}, Janus Pro~\cite{chen2025janus}, DeepSeek VL2~\cite{wu2024deepseek}, Phi-4~\cite{abdin2024phi}, LLaVA OneVision~\cite{li2024llava}, Pixtral~\cite{agrawal2024pixtral}, Ovis~\cite{lu2024ovis}, ChartGemma~\cite{masry2024chartgemma}, TinyChart~\cite{zhang2024tinychart}, and ChartInstruct~\cite{masry2024chartinstruct}.
For proprietary models, we tested: OpenAI O4-mini~\cite{o4mini}, GPT-4.1~\cite{gpt41}, GPT-4o~\cite{hurst2024gpt4o}, Claude 3.5 Sonnet~\cite{claude35}, Gemini 2.5 Pro Preview~\cite{gemini25pro}, and Gemini 2.5 Flash Preview~\cite{gemini25flash}. 
We provide test configurations for all these models in Appendix \ref{append:config}.

\textbf{Human baseline.} We recruited 15 human participants \changjian{with expertise in deep learning and visualization} and report their performance (\textit{i.e.}, Human) on \InfoChartQA{} as a baseline (see Appendix \ref{app:humanevalinfo} for more information of human evaluation). 
To enable a fair comparison between humans and models, we presented the participants with the same questions and instructions and evaluated their responses using the same criteria as those applied to the models.
\minzhi{
Given the substantial time and cost involved, human performance was evaluated on a $10\%$ subset of text-based and basic visual questions, while all metaphor-related questions were included due to their limited number.}

\textbf{Evaluation metric.} \InfoChartQA{} consists of multiple forms of questions with textual, numeric, and option answers.
For textual answers, answers were considered correct if the ANLS score exceeded $0.8$.
\changjian{The ANLS score evaluates the similarity between the model-generated answer and the ground truth based on the number of edits needed to convert one text into the other~\cite{mathew2021docvqa,li2027levenshtein}.}
For numeric answers, we employed the commonly used relaxed accuracy metric in chart question answering benchmarks~\cite{Methani2020plotqa}.
To avoid errors introduced by different forms of numbers (\eg, ``1K'' and ``1,000''), we normalized the numbers into a unified form, \eg, from ``1K'' to ``1,000''.
For option answers, we considered an answer correct if it exactly matched the ground truth.
\xie{The pseudocode of the evaluation process can be found in Appendix \ref{append:eval code}.}
\subsection{Quantitative MLLM Evaluation Results on the \InfoChartQA{} Benchmark}
We present our main result in Table~\ref{tab:main_res}. The detailed breakdowns, sampled questions, answers, and model responses can be found in Appendix \ref{append:sample}.
Key observations include:


\textbf{The performance of MLLMs degrades on infographic charts compared to plain charts.} 
As shown in Table~\ref{tab:main_res}, the top-performing models demonstrated impressive performance on plain chart benchmarks, sometimes on a par with human performance.
For example, Gemini 2.5 Pro Preview achieved $91.16\%$ on plain charts while the human baseline was \xie{$95.44\%$}.
This result is also consistent with existing studies ~\cite{wang2024charxiv,o4mini}.
However, the performance of all models deteriorated significantly when on infographic charts. 
It shows that there is significant potential for improvement in the infographic chart understanding abilities of MLLMs.


\textbf{Strong performance on text-based questions is foundational to strong performance on visual-element-based questions.}
Visual-element-based questions evaluate not only a model's ability to understand charts but also its visual alignment capability (\eg, align the cropped element with the whole image).
If a model lacks strong chart understanding ability, it is likely to perform poorly on visual-element-based questions.
As shown in Table~\ref{tab:main_res}, models that performed well on visual-element-based questions, such as GPT-4.1 and Gemini 2.5 Pro Preview, generally exhibited strong performance on text-based questions.
Conversely, models that performed poorly on visual-element-based questions, such as ChartGemma and TinyChart, tended to have weaker performance on text-based questions.
\xie{This observation was further supported by a high and statistically significant Spearman correlation between the two sets of results ($\rho=0.895$, $p<0.01$).}


\textbf{Metaphor-related questions are challenging for MLLMs.}
We found that understanding visual metaphors in infographic charts was still challenging for current MLLMs.
Even though some models achieved \xie{approximately} $80\%$ accuracy on text-based questions in infographic charts (\eg, \xie{$79.23\%$} for Gemini 2.5 Pro Preview), their performance dropped by \xie{around} $20\%$ on metaphor-related questions, down to $60.42\%$.
On the other side, the human baseline showed a smaller drop, from $94.63\%$ to $88.69\%$.
This gap indicates that the alignment between abstract concepts and visual elements (\eg, a rising balloon symbolizing hope) needs to be enhanced in current MLLMs.


\subsection{Analysis on Performance Degradation for Infographic Charts}
Since \InfoChartQA{} processes paired infographic and plain charts sharing the same underlying data, it enables us to perform ablation studies to analyze the performance degradation for infographic charts. 

\begin{figure}[!tb]
    \centering
    \subfloat[]{
        \includegraphics[width=0.19\textwidth, page=1]{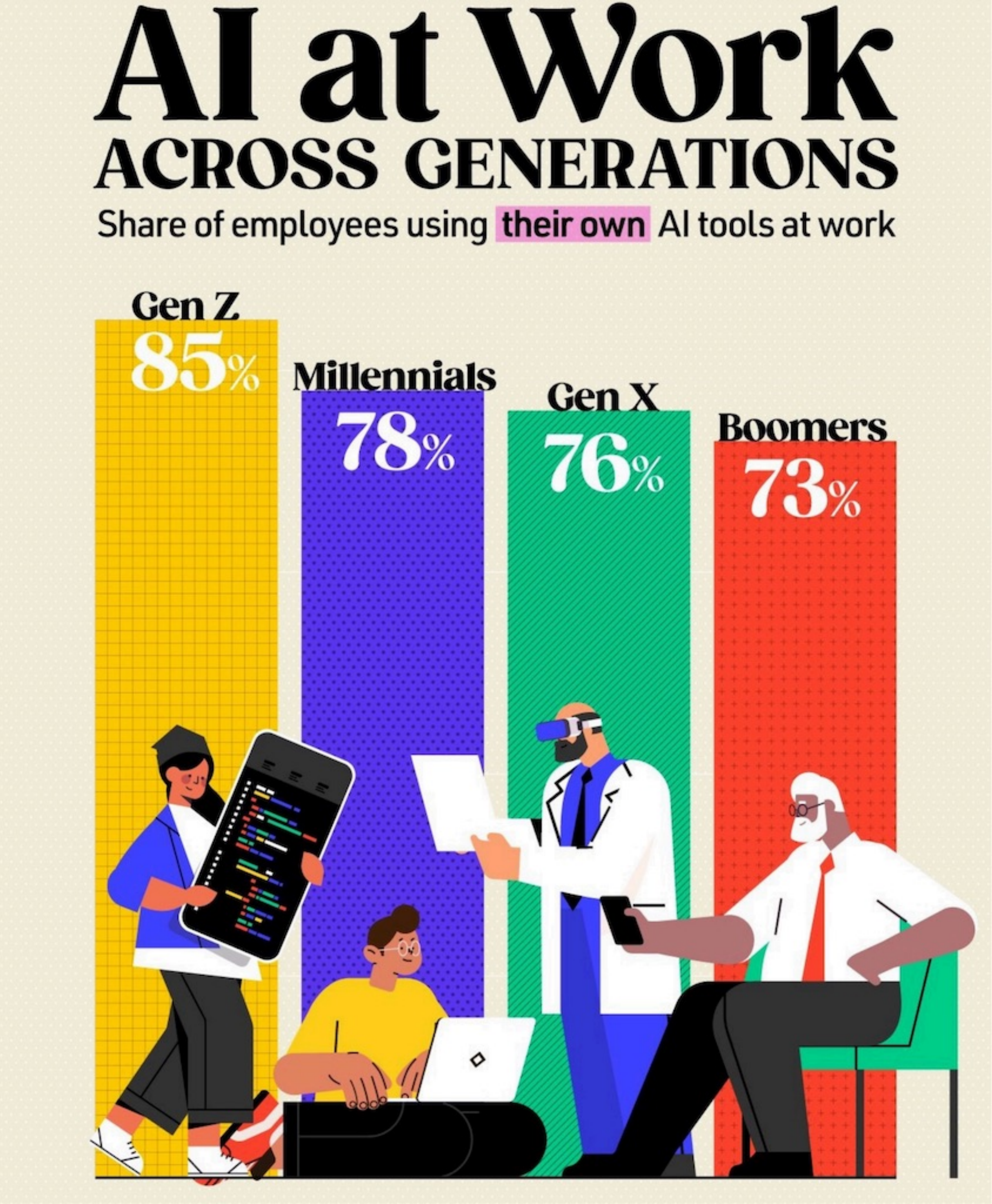}
        \label{fig:subfig1}
    }
    \subfloat[]{
        \includegraphics[width=0.19\textwidth, page=2]{pictures/remove_sample_gra_1.pdf}
        \label{fig:subfig2}
    }
    \subfloat[]{
        \includegraphics[width=0.19\textwidth, page=3]{pictures/remove_sample_gra_1.pdf}
        \label{fig:subfig3}
    }
    \subfloat[]{
        \includegraphics[width=0.19\textwidth, page=4]{pictures/remove_sample_gra_1.pdf}
        \label{fig:subfig4}
    }
    \subfloat[]{
        \includegraphics[width=0.19\textwidth, page=5]{pictures/remove_sample_gra_1.pdf}
        \label{fig:subfig5}
    }
    \caption{Example of \textbf{progressively} removing visual elements from infographic charts.}
    \label{fig:reomve_imagery}
\end{figure}

\subsubsection{Visual elements primarily contribute to the performance degradation}\label{sec:4.3.1}

Unlike plain charts, infographic charts often incorporate a wider variety and higher density of visual elements, such as metaphorical imagery, to convey information. 
\changjian{To understand how these elements affect model performance, we grouped infographic charts by the number of visual elements and compared their performance.
We observed that models performed worse on infographic charts with more visual elements compared to those with fewer elements.
For example, infographic charts with 100 visual elements had 10\% lower accuracy than those with 20 visual elements (details are provided in Appendix \ref{append:sample}).}
This suggests that these elements substantially increase the visual complexity of infographic charts, posing challenges for current MLLMs.

\changjian{Since the experiment results above may be influenced by other factors, we conducted a controlled experiment to distangle the impact of visual elements from these other factors.}
As illustrated in Figure~\ref{fig:reomve_imagery}, we manually selected 300 infographic charts from our dataset that feature rich visual elements. 
These charts were then edited to \textbf{progressively} remove visual elements, resulting in versions of the \textbf{same} infographic with \textbf{different} numbers of visual elements, ranging from $0$ to $n$, where $n$ denotes the original number of visual elements.
Text-based QA was then evaluated on these charts, selecting only those remaining answerable even after all visual elements were removed.
This allowed us to observe changes in model performance on the \textbf{same} infographic chart but with \textbf{different} numbers of visual elements.

The results of GPT-4.1 and TinyChart are shown in Figure~\ref{fig:number_of_elements}(a) and (b). \xie{Results of other models are provided in Appendix \ref{app:supp_431}.}
As we can see, after removing all visual elements, the model's accuracy nearly aligned with that on plain charts.
Our results validate that the \fix{visual} elements are the primary cause of the observed performance drop on infographic charts.

\changjian{To further validate this conclusion, we conducted an experiment for model improvement. 
We revised the prompt instruction to explicitly guide the model to focus
on visualization components rather than decorative elements.
The performance of GPT-4.1 is improved by $2.93\%$.
More details about this experiment can be found in Appendix~\ref{app:supp_model_improvement}.
}

\begin{figure}[!bth]
        
     \centering
    \begin{subfigure}{0.4\textwidth}
    \centering
        \includegraphics[width=1\textwidth]{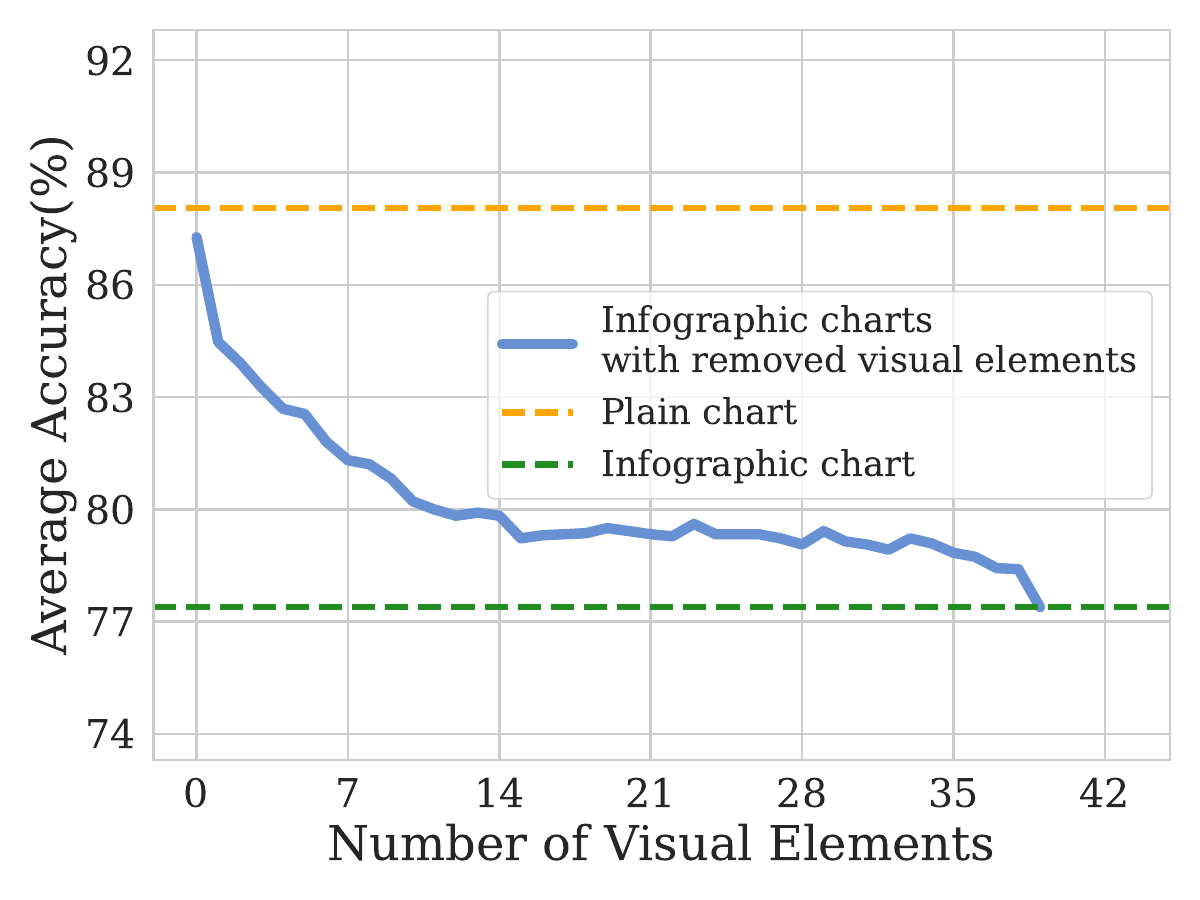}
        \caption{GPT-4.1}
    \end{subfigure}
    \begin{subfigure}{0.4\textwidth}
    \centering
        \includegraphics[width=1\textwidth]{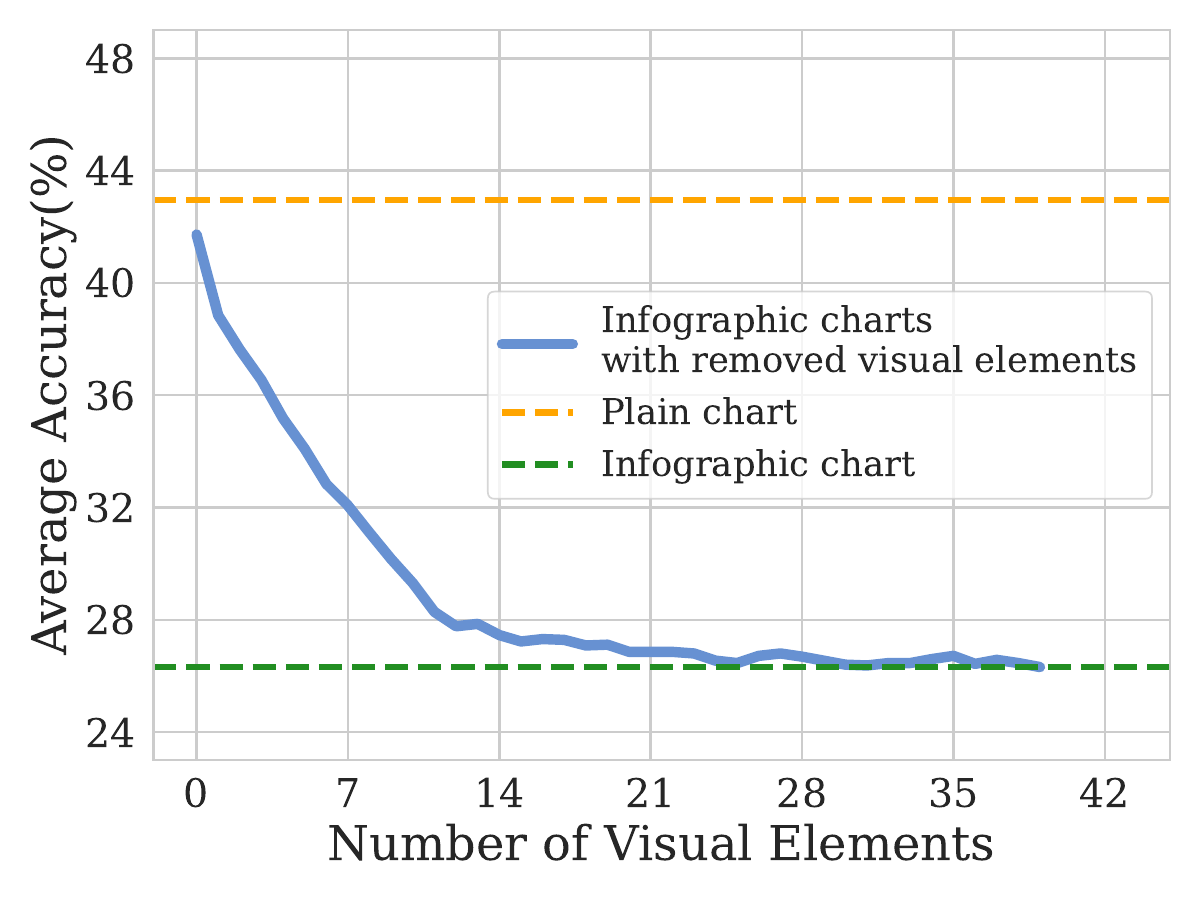}
        \caption{TinyChart}
    \end{subfigure}

   \caption{Model's performance change on the \textbf{same} infographic chart but with \textbf{different} number of visual elements.}
    \label{fig:number_of_elements}
\end{figure}



\newpage
\subsubsection{Clearer connections between text and visual elements improve understanding}

\begin{wraptable}{r}{0.6\textwidth}
\vspace{-10pt}
\centering
\resizebox{1\linewidth}{!}{
\normalsize
\begin{tabular}{lccccc}
\toprule
Model &  Obstructions Removal & Auxiliary Line & Position Perturbation \\
\midrule
GPT-4.1      & $\uparrow0.79$ & $\uparrow0.85$ & $\downarrow2.95$\\
TinyChart  & $\uparrow3.23$       & $\uparrow3.08$ & $\downarrow2.78$\\
\bottomrule
\end{tabular}
}
\caption{Performance changes (\%) of GPT-4.1 and TinyChart under three types of modifications.}
\label{tab:perform_on_three_types}
\end{wraptable}
Knowing that the \fix{visual} elements affect the model's performance, we further investigated how they affect it. 
We discovered that the more \fix{visual} elements were overlaid onto the charts, the lower the performance became. 
Since the overlay disturbs the \textbf{connections} between labels, visualization elements (\eg, bars in bar charts), and numerical annotations,
we hypothesized that the model's ability to understand charts relies on such connections.  
Ambiguities in the connections, like occlusions or positional misalignments, can degrade model performance.

\begin{figure}[!tb]
    \centering
    \begin{subfigure}[b]{0.27\textwidth}
        \includegraphics[width=\textwidth,height=90pt]{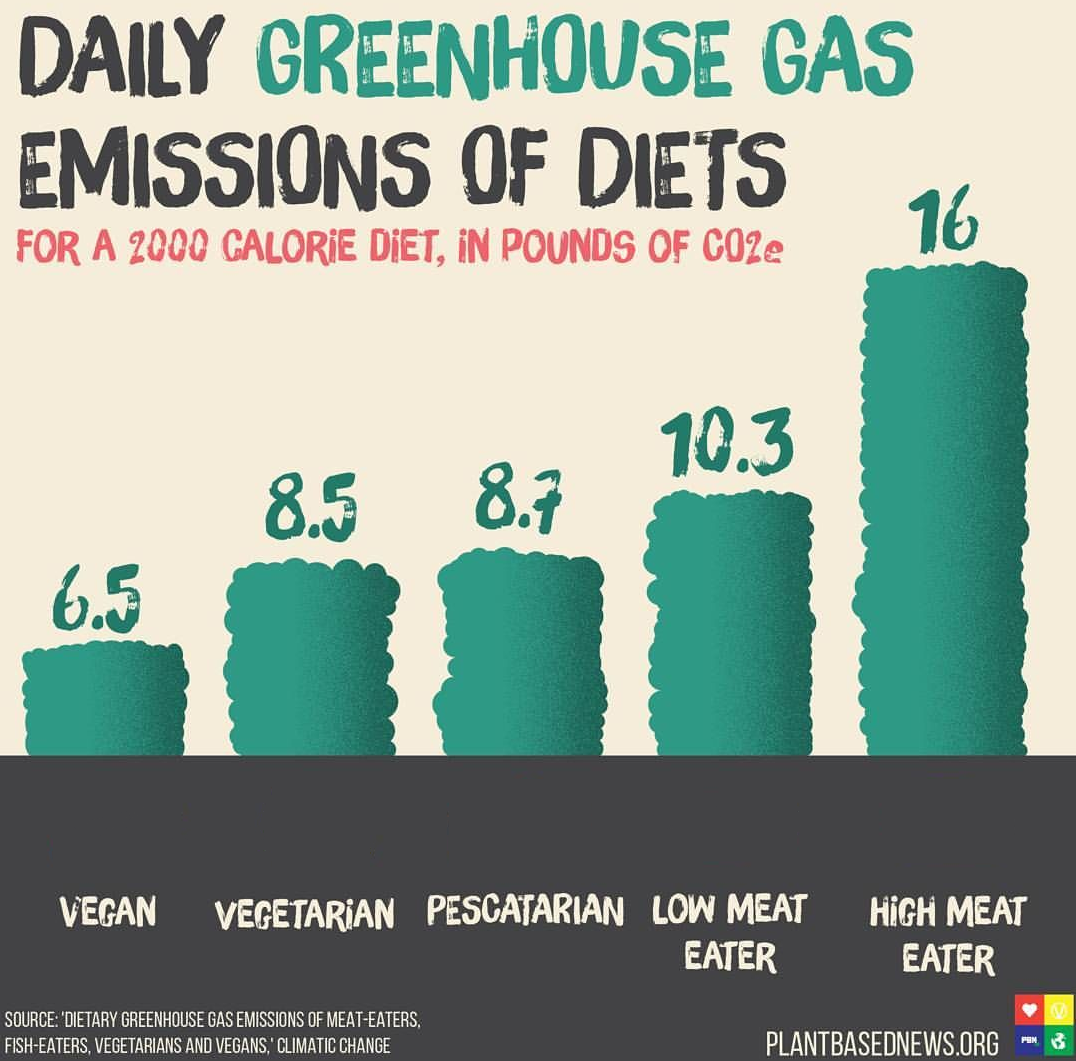}
        \caption{Obstructions Removal}
    \end{subfigure}
    ~ 
    \begin{subfigure}[b]{0.27\textwidth}
        \includegraphics[width=\textwidth,height=90pt]{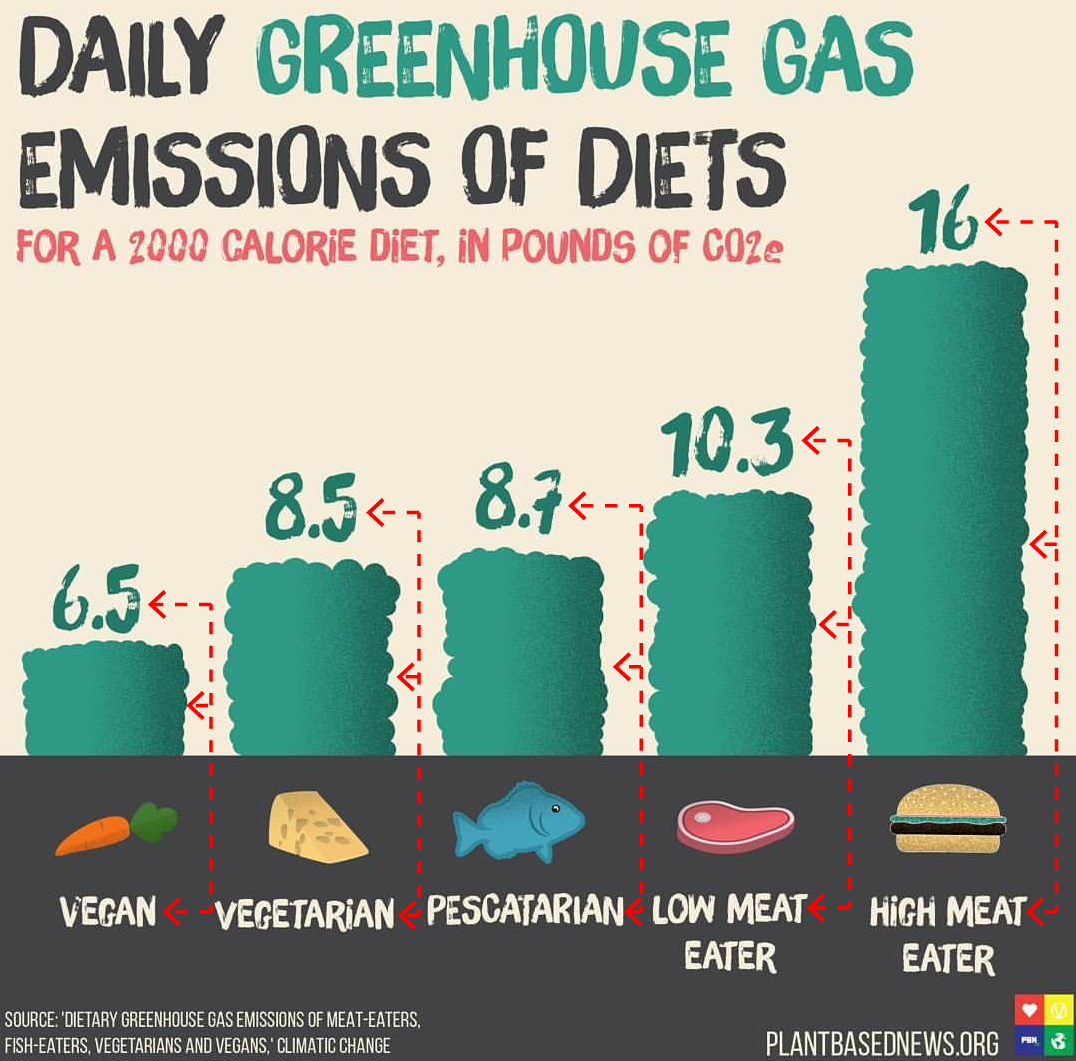}
        \caption{Auxiliary Line}
    \end{subfigure}
    ~
    \begin{subfigure}[b]{0.27\textwidth}
        \includegraphics[width=\textwidth,height=90pt]{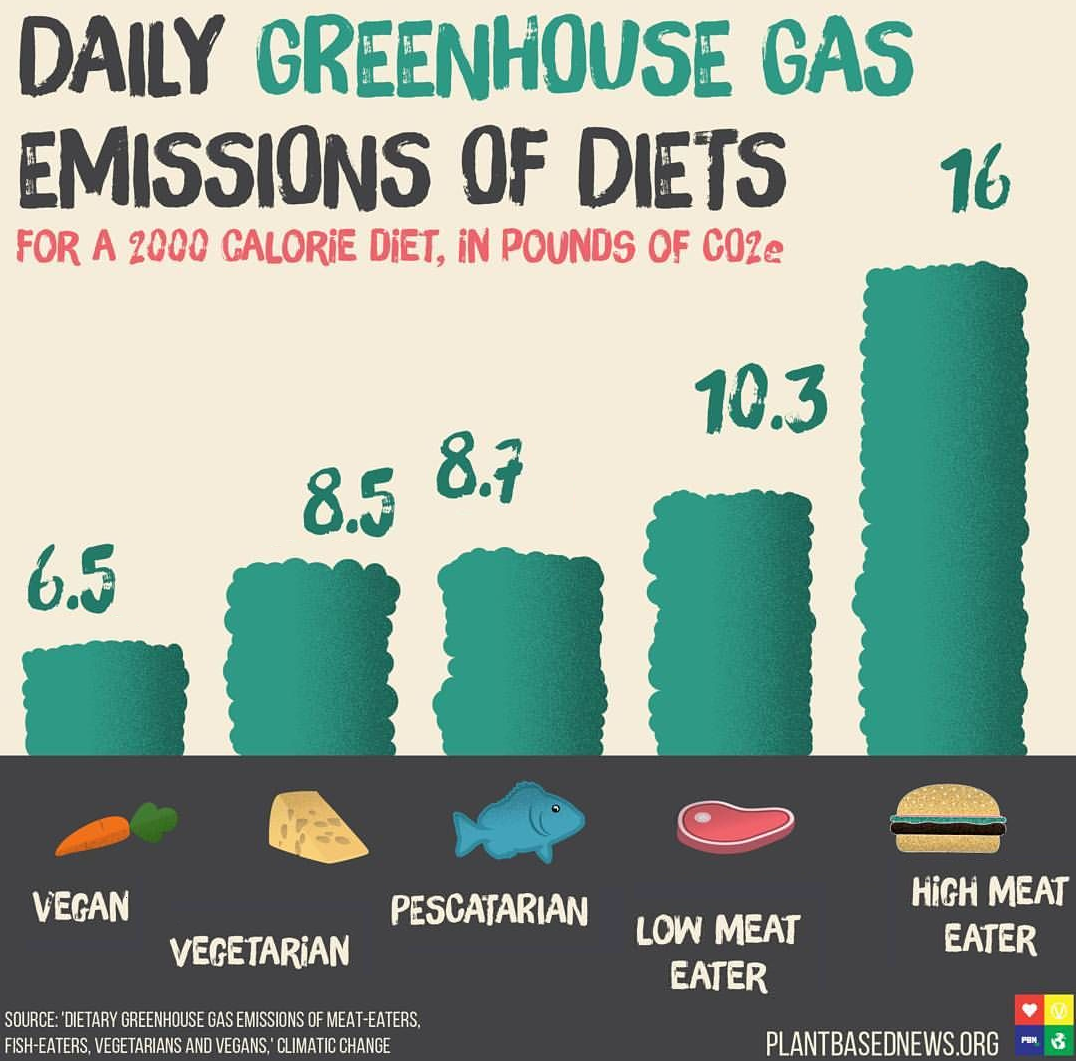}
        \caption{Position Perturbation}
    \end{subfigure}
    \caption{Different modifications on charts.}
    \label{fig:modified_images}
\end{figure}
To validate this hypothesis, we randomly selected 200 images and applied three different types of modifications to introduce varying levels of perturbation to the connections.
As shown in Figure \ref{fig:modified_images}:

1) \textbf{Obstructions Removal.} We eliminated obstructions that hinder the connections. 
2) \textbf{Auxiliary Lines.} We introduced auxiliary lines to explicitly connect texts with their associated visual elements and chart components. 
3) \textbf{Position Perturbation.} We randomly shifted the positions of the bars, labels, and annotations to disrupt the connections.

The result is shown in Table~\ref{tab:perform_on_three_types}. Notably, even simple modifications, as shown in Figure~\ref{fig:modified_images}(a) and (b), to highlight the connection can lead to comparably better performance and vice versa.





\subsubsection{MLLMs are sensitive to the orders of text labels}

Since the majority of QAs in \InfoChartQA{} are designed based on data facts, we further analyzed how different data facts affect model performance.
The accuracy across different data facts is shown in Appendix \ref{sssec:detailed_main_result}. We observed that only the \textit{rank} and \textit{outlier} questions exhibited accuracies below 50\%.
To better understand the underlying causes, we focused on analyzing these two categories.

\begin{figure}[!th]
\vspace{-20pt}
\centering
    \begin{subfigure}{0.35\textwidth}
    \centering
        \includegraphics[width=1\textwidth]{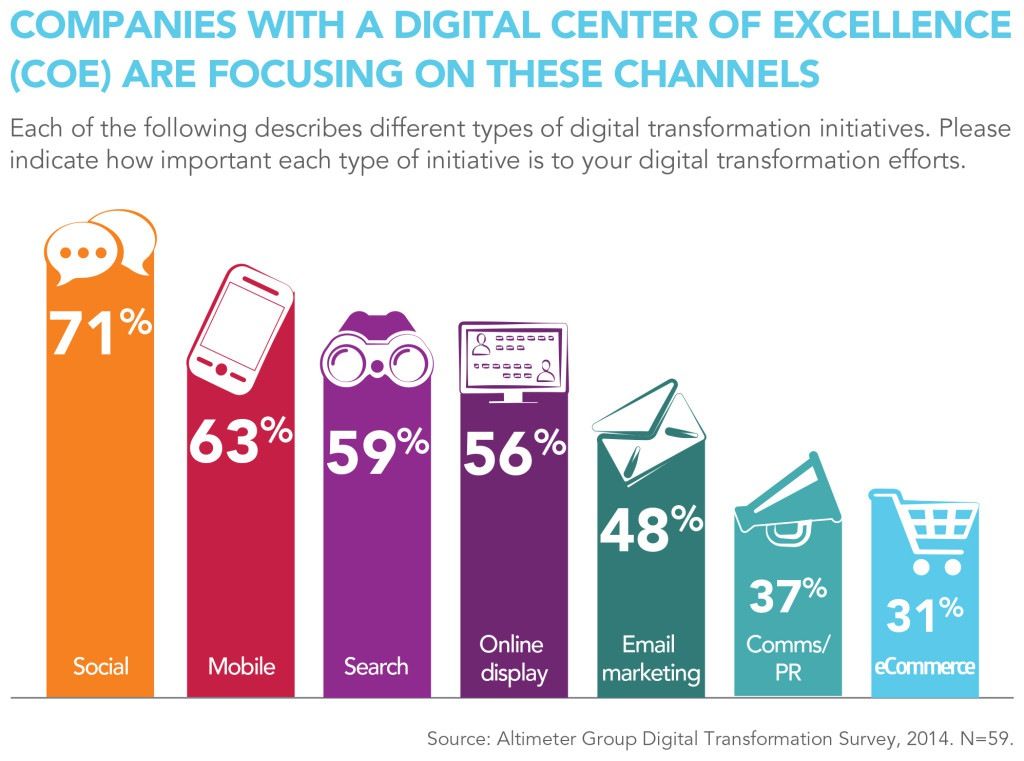}
        \caption{Before shuffle (Acc: $100\%$)}
    \end{subfigure}
    \begin{subfigure}{0.35\textwidth}
    \centering
        \includegraphics[width=1\textwidth]{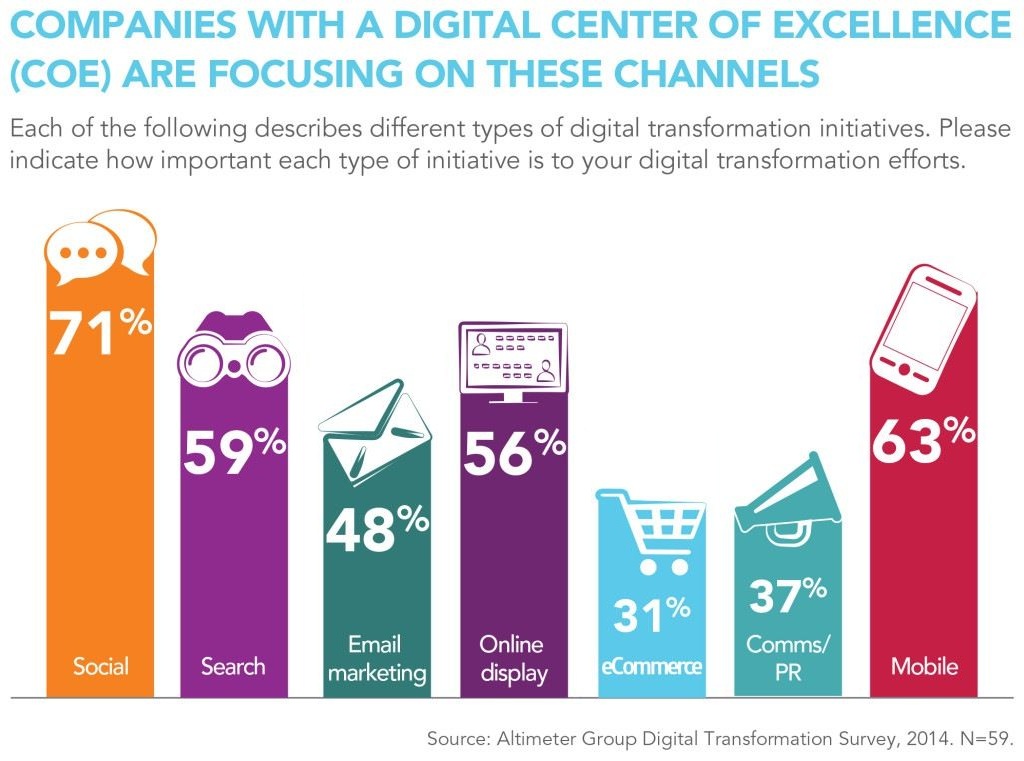}
        \caption{After shuffle (Acc: $76.3\%$)}
    \end{subfigure}
    
    \caption{Sample charts before/after shuffle and corresponding performance.}
    \label{fig:shuffle}
\end{figure}

For \textit{rank} questions, we found that MLLMs tended to 
answer the questions based on the order of the text labels rather than the actual data values. 
Based on this observation, we hypothesized that MLLMs were sensitive to label order.
To validate this, we randomly select $200$ charts on which the model (GPT-4.1) originally achieved correct rankings and apply random spatial permutations to the text labels within each chart, while preserving their semantic content, as shown in Figure~\ref{fig:shuffle}(a) and (b). 
When re-evaluated on these shuffled samples, the model's accuracy dropped from 100\% to 76.3\%. 
This substantial decline strongly suggests that the model relies heavily on superficial cues such as label order, rather than developing a robust understanding of ranking.

For \textit{outlier} questions, answering correctly primarily requires the ability to perceive spatial relationships and contextual dependencies. 
Our results indicate notable limitations in this aspect, suggesting that the model lacks a fine-grained understanding of spatial configurations. 
This observation is consistent with prior work \cite{Yang2024ThinkingIS}, which has also identified shortcomings in MLLMs regarding spatial and relational reasoning capabilities.
\section{Limitations and Conclusion}
\textbf{Limitations}.
While \InfoChartQA{} presents a comprehensive benchmark dedicated to infographic charts understanding with special visual-element-based questions, it still has some limitations, which highlight areas for further research.
First, the difficulty in constructing metaphor-related questions limits the scale of testing for this subtle type of multimodal understanding.  
Increasing the amount of such questions and performing more fine-grained metaphor analysis may elicit more insights into the challenges of infographic charts.
Second, although we actively involved human experts in the creation and verification of questions, some parts of our question generation pipeline rely on templates or large language models.
This may limit the out-of-distribution diversity regarding the textual part of the questions, although the visual part of our questions exhibits superior diversity compared to existing benchmarks with complex real-world infographic charts and a wider range of chart types.
\xie{Third, the participants in our user study consist of 15 students of similar ages, all with expertise primarily in deep learning and visualization. 
As a result, they may not fully represent the broader population with varying ages and areas of expertise.
}
\xie{Moreover, although we briefly discussed how prompt engineering can enhance model performance on infographic charts based on the findings in our ablation studies, exploring how to better leverage these findings to improve MLLMs remains an important direction for future work.}

\textbf{Conclusion}.
In this paper, we present \InfoChartQA{}, a novel benchmark for infographic chart understanding, with particular focus on evaluating MLLMs' reasoning ability on complicated multimodal questions.
It involves a combination of heterogeneous \fix{pictorial visual} elements or metaphors and the underlying data relationships.
In this benchmark, we first construct paired infographic charts and their plain chart counterparts to pinpoint the source of model failure in either data complexity itself or additional infographic elements.
\InfoChartQA{} also extends the QA space by introducing visual-element-based questions unique to infographic charts, enabling more detailed analysis of visual reasoning capabilities.
Experimental results highlight the special challenges of infographics, especially in visual-element-based questions, with further analysis revealing three performance degradation factors, including the impact of visual elements, the ambiguous connection between text and visual elements, and orders of text labels.
We hope that \InfoChartQA{} can provide a new perspective and a reliable foundation for evaluating more complex chart reasoning capabilities. 

\section*{Acknowledgments}
This work was supported by the National Natural Science Foundation of China under grants U21A20469, 61936002, and 62402167, in part by Tsinghua-Kuaishou Institute of Future Media Data, Yuelushan Laboratory Breeding Program under the grant YLS-2025-ZY01015, the Hunan Natural Science Foundation under the grant 2025JJ60419, and the Science and Technology Innovation Program of Hunan Province under the grant 2023ZJ1080.

{\small

\bibliographystyle{unsrt}
\bibliography{reference}
}









\section*{NeurIPS Paper Checklist}

\begin{enumerate}

\item {\bf Claims}
    \item[] Question: Do the main claims made in the abstract and introduction accurately reflect the paper's contributions and scope?
    \item[] Answer: \answerYes{} 
    \item[] Justification: The abstract and introduction accurately describe the proposed benchmark(\InfoChartQA) and summarize the experimental results.
    \item[] Guidelines:
    \begin{itemize}
        \item The answer NA means that the abstract and introduction do not include the claims made in the paper.
        \item The abstract and/or introduction should clearly state the claims made, including the contributions made in the paper and important assumptions and limitations. A No or NA answer to this question will not be perceived well by the reviewers. 
        \item The claims made should match theoretical and experimental results, and reflect how much the results can be expected to generalize to other settings. 
        \item It is fine to include aspirational goals as motivation as long as it is clear that these goals are not attained by the paper. 
    \end{itemize}

\item {\bf Limitations}
    \item[] Question: Does the paper discuss the limitations of the work performed by the authors?
    \item[] Answer: \answerYes{} 
    \item[] Justification: Section 5 discusses limitations, including limited metaphor-related reasoning coverage and potential linguistic diversity constraints from template/LLM reliance.
    \item[] Guidelines:
    \begin{itemize}
        \item The answer NA means that the paper has no limitation while the answer No means that the paper has limitations, but those are not discussed in the paper. 
        \item The authors are encouraged to create a separate "Limitations" section in their paper.
        \item The paper should point out any strong assumptions and how robust the results are to violations of these assumptions (e.g., independence assumptions, noiseless settings, model well-specification, asymptotic approximations only holding locally). The authors should reflect on how these assumptions might be violated in practice and what the implications would be.
        \item The authors should reflect on the scope of the claims made, e.g., if the approach was only tested on a few datasets or with a few runs. In general, empirical results often depend on implicit assumptions, which should be articulated.
        \item The authors should reflect on the factors that influence the performance of the approach. For example, a facial recognition algorithm may perform poorly when image resolution is low or images are taken in low lighting. Or a speech-to-text system might not be used reliably to provide closed captions for online lectures because it fails to handle technical jargon.
        \item The authors should discuss the computational efficiency of the proposed algorithms and how they scale with dataset size.
        \item If applicable, the authors should discuss possible limitations of their approach to address problems of privacy and fairness.
        \item While the authors might fear that complete honesty about limitations might be used by reviewers as grounds for rejection, a worse outcome might be that reviewers discover limitations that aren't acknowledged in the paper. The authors should use their best judgment and recognize that individual actions in favor of transparency play an important role in developing norms that preserve the integrity of the community. Reviewers will be specifically instructed to not penalize honesty concerning limitations.
    \end{itemize}

\item {\bf Theory assumptions and proofs}
    \item[] Question: For each theoretical result, does the paper provide the full set of assumptions and a complete (and correct) proof?
    \item[] Answer: \answerNA{} 
    \item[] Justification: No theoretical
results are given.
    \item[] Guidelines:
    \begin{itemize}
        \item The answer NA means that the paper does not include theoretical results. 
        \item All the theorems, formulas, and proofs in the paper should be numbered and cross-referenced.
        \item All assumptions should be clearly stated or referenced in the statement of any theorems.
        \item The proofs can either appear in the main paper or the supplemental material, but if they appear in the supplemental material, the authors are encouraged to provide a short proof sketch to provide intuition. 
        \item Inversely, any informal proof provided in the core of the paper should be complemented by formal proofs provided in appendix or supplemental material.
        \item Theorems and Lemmas that the proof relies upon should be properly referenced. 
    \end{itemize}

    \item {\bf Experimental result reproducibility}
    \item[] Question: Does the paper fully disclose all the information needed to reproduce the main experimental results of the paper to the extent that it affects the main claims and/or conclusions of the paper (regardless of whether the code and data are provided or not)?
    \item[] Answer: \answerYes{} 
    \item[] Justification: Experiments presented in the paper are described in detail in Appendix~\ref{append:exp}. We also release the data and code needed to reproduce results on \href{https://github.com/CoolDawnAnt/InfoChartQA}{https://github.com/CoolDawnAnt/InfoChartQA}.
    \item[] Guidelines:
    \begin{itemize}
        \item The answer NA means that the paper does not include experiments.
        \item If the paper includes experiments, a No answer to this question will not be perceived well by the reviewers: Making the paper reproducible is important, regardless of whether the code and data are provided or not.
        \item If the contribution is a dataset and/or model, the authors should describe the steps taken to make their results reproducible or verifiable. 
        \item Depending on the contribution, reproducibility can be accomplished in various ways. For example, if the contribution is a novel architecture, describing the architecture fully might suffice, or if the contribution is a specific model and empirical evaluation, it may be necessary to either make it possible for others to replicate the model with the same dataset, or provide access to the model. In general. releasing code and data is often one good way to accomplish this, but reproducibility can also be provided via detailed instructions for how to replicate the results, access to a hosted model (e.g., in the case of a large language model), releasing of a model checkpoint, or other means that are appropriate to the research performed.
        \item While NeurIPS does not require releasing code, the conference does require all submissions to provide some reasonable avenue for reproducibility, which may depend on the nature of the contribution. For example
        \begin{enumerate}
            \item If the contribution is primarily a new algorithm, the paper should make it clear how to reproduce that algorithm.
            \item If the contribution is primarily a new model architecture, the paper should describe the architecture clearly and fully.
            \item If the contribution is a new model (e.g., a large language model), then there should either be a way to access this model for reproducing the results or a way to reproduce the model (e.g., with an open-source dataset or instructions for how to construct the dataset).
            \item We recognize that reproducibility may be tricky in some cases, in which case authors are welcome to describe the particular way they provide for reproducibility. In the case of closed-source models, it may be that access to the model is limited in some way (e.g., to registered users), but it should be possible for other researchers to have some path to reproducing or verifying the results.
        \end{enumerate}
    \end{itemize}

\item {\bf Open access to data and code}
    \item[] Question: Does the paper provide open access to the data and code, with sufficient instructions to faithfully reproduce the main experimental results, as described in supplemental material?
    \item[] Answer: \answerYes{} 
    \item[] Justification: Our benchmark and code are linked after the abstract . The benchmark can be accessed on HuggingFace at \href{https://huggingface.co/datasets/Jietson/InfoChartQA}{https://huggingface.co/datasets/Jietson/InfoChartQA}
    \item[] Guidelines:
    \begin{itemize}
        \item The answer NA means that paper does not include experiments requiring code.
        \item Please see the NeurIPS code and data submission guidelines (\url{https://nips.cc/public/guides/CodeSubmissionPolicy}) for more details.
        \item While we encourage the release of code and data, we understand that this might not be possible, so “No” is an acceptable answer. Papers cannot be rejected simply for not including code, unless this is central to the contribution (e.g., for a new open-source benchmark).
        \item The instructions should contain the exact command and environment needed to run to reproduce the results. See the NeurIPS code and data submission guidelines (\url{https://nips.cc/public/guides/CodeSubmissionPolicy}) for more details.
        \item The authors should provide instructions on data access and preparation, including how to access the raw data, preprocessed data, intermediate data, and generated data, etc.
        \item The authors should provide scripts to reproduce all experimental results for the new proposed method and baselines. If only a subset of experiments are reproducible, they should state which ones are omitted from the script and why.
        \item At submission time, to preserve anonymity, the authors should release anonymized versions (if applicable).
        \item Providing as much information as possible in supplemental material (appended to the paper) is recommended, but including URLs to data and code is permitted.
    \end{itemize}

\item {\bf Experimental setting/details}
    \item[] Question: Does the paper specify all the training and test details (e.g., data splits, hyperparameters, how they were chosen, type of optimizer, etc.) necessary to understand the results?
    \item[] Answer: \answerYes{} 
    \item[] Justification: The settings of the model used in Section 4 are provided in Appendix~\ref{append:config}, and data details are shared in Appendix~\ref{append:data}.
    \item[] Guidelines:
    \begin{itemize}
        \item The answer NA means that the paper does not include experiments.
        \item The experimental setting should be presented in the core of the paper to a level of detail that is necessary to appreciate the results and make sense of them.
        \item The full details can be provided either with the code, in appendix, or as supplemental material.
    \end{itemize}

\item {\bf Experiment statistical significance}
    \item[] Question: Does the paper report error bars suitably and correctly defined or other appropriate information about the statistical significance of the experiments?
    \item[] Answer: \answerNo{} 
    \item[] Justification:  All our experiment runs are quite costly, which limits our capability to do
multiple runs.
    \item[] Guidelines:
    \begin{itemize}
        \item The answer NA means that the paper does not include experiments.
        \item The authors should answer "Yes" if the results are accompanied by error bars, confidence intervals, or statistical significance tests, at least for the experiments that support the main claims of the paper.
        \item The factors of variability that the error bars are capturing should be clearly stated (for example, train/test split, initialization, random drawing of some parameter, or overall run with given experimental conditions).
        \item The method for calculating the error bars should be explained (closed form formula, call to a library function, bootstrap, etc.)
        \item The assumptions made should be given (e.g., Normally distributed errors).
        \item It should be clear whether the error bar is the standard deviation or the standard error of the mean.
        \item It is OK to report 1-sigma error bars, but one should state it. The authors should preferably report a 2-sigma error bar than state that they have a 96\% CI, if the hypothesis of Normality of errors is not verified.
        \item For asymmetric distributions, the authors should be careful not to show in tables or figures symmetric error bars that would yield results that are out of range (e.g. negative error rates).
        \item If error bars are reported in tables or plots, The authors should explain in the text how they were calculated and reference the corresponding figures or tables in the text.
    \end{itemize}

\item {\bf Experiments compute resources}
    \item[] Question: For each experiment, does the paper provide sufficient information on the computer resources (type of compute workers, memory, time of execution) needed to reproduce the experiments?
    \item[] Answer: \answerYes{} 
    \item[] Justification: The paper specifies the hardware used in Appendix~\ref{append:exp}. We accessed proprietary models via API, while open source models testing and analysis experiments were run locally. 
    \item[] Guidelines:
    \begin{itemize}
        \item The answer NA means that the paper does not include experiments.
        \item The paper should indicate the type of compute workers CPU or GPU, internal cluster, or cloud provider, including relevant memory and storage.
        \item The paper should provide the amount of compute required for each of the individual experimental runs as well as estimate the total compute. 
        \item The paper should disclose whether the full research project required more compute than the experiments reported in the paper (e.g., preliminary or failed experiments that didn't make it into the paper). 
    \end{itemize}
    
\item {\bf Code of ethics}
    \item[] Question: Does the research conducted in the paper conform, in every respect, with the NeurIPS Code of Ethics \url{https://neurips.cc/public/EthicsGuidelines}?
    \item[] Answer: \answerYes{} 
    \item[] Justification: The research involves benchmark construction and result analysis, and we assume it conforms to the NeurIPS Code of Ethics.
    Particularly, the infographic chart image data we collected is all publicly available data, and we only release the URLs to avoid potential copyright infringement.
    We also double-checked the image content to ensure that there is no harmful or illegal content.
    \item[] Guidelines:
    \begin{itemize}
        \item The answer NA means that the authors have not reviewed the NeurIPS Code of Ethics.
        \item If the authors answer No, they should explain the special circumstances that require a deviation from the Code of Ethics.
        \item The authors should make sure to preserve anonymity (e.g., if there is a special consideration due to laws or regulations in their jurisdiction).
    \end{itemize}

\item {\bf Broader impacts}
    \item[] Question: Does the paper discuss both potential positive societal impacts and negative societal impacts of the work performed?
    \item[] Answer: \answerNA{} 
    \item[] Justification: The paper focuses on the technical contributions and does not include a specific discussion of broader positive or negative societal impacts.
    \item[] Guidelines:
    \begin{itemize}
        \item The answer NA means that there is no societal impact of the work performed.
        \item If the authors answer NA or No, they should explain why their work has no societal impact or why the paper does not address societal impact.
        \item Examples of negative societal impacts include potential malicious or unintended uses (e.g., disinformation, generating fake profiles, surveillance), fairness considerations (e.g., deployment of technologies that could make decisions that unfairly impact specific groups), privacy considerations, and security considerations.
        \item The conference expects that many papers will be foundational research and not tied to particular applications, let alone deployments. However, if there is a direct path to any negative applications, the authors should point it out. For example, it is legitimate to point out that an improvement in the quality of generative models could be used to generate deepfakes for disinformation. On the other hand, it is not needed to point out that a generic algorithm for optimizing neural networks could enable people to train models that generate Deepfakes faster.
        \item The authors should consider possible harms that could arise when the technology is being used as intended and functioning correctly, harms that could arise when the technology is being used as intended but gives incorrect results, and harms following from (intentional or unintentional) misuse of the technology.
        \item If there are negative societal impacts, the authors could also discuss possible mitigation strategies (e.g., gated release of models, providing defenses in addition to attacks, mechanisms for monitoring misuse, mechanisms to monitor how a system learns from feedback over time, improving the efficiency and accessibility of ML).
    \end{itemize}
    
\item {\bf Safeguards}
    \item[] Question: Does the paper describe safeguards that have been put in place for responsible release of data or models that have a high risk for misuse (e.g., pretrained language models, image generators, or scraped datasets)?
    \item[] Answer: \answerYes{} 
    \item[] Justification: We used manual review and MLLMs filtering to ensure safe images in our dataset.
    \item[] Guidelines:
    \begin{itemize}
        \item The answer NA means that the paper poses no such risks.
        \item Released models that have a high risk for misuse or dual-use should be released with necessary safeguards to allow for controlled use of the model, for example by requiring that users adhere to usage guidelines or restrictions to access the model or implementing safety filters. 
        \item Datasets that have been scraped from the Internet could pose safety risks. The authors should describe how they avoided releasing unsafe images.
        \item We recognize that providing effective safeguards is challenging, and many papers do not require this, but we encourage authors to take this into account and make a best faith effort.
    \end{itemize}

\item {\bf Licenses for existing assets}
    \item[] Question: Are the creators or original owners of assets (e.g., code, data, models), used in the paper, properly credited and are the license and terms of use explicitly mentioned and properly respected?
    \item[] Answer: \answerYes{} 
    \item[] Justification: The paper properly cites the sources for existing assets including models and data sources. The specific licenses and terms of use for these assets are mentioned in Appendix~\ref{append:license}.
    \item[] Guidelines:
    \begin{itemize}
        \item The answer NA means that the paper does not use existing assets.
        \item The authors should cite the original paper that produced the code package or dataset.
        \item The authors should state which version of the asset is used and, if possible, include a URL.
        \item The name of the license (e.g., CC-BY 4.0) should be included for each asset.
        \item For scraped data from a particular source (e.g., website), the copyright and terms of service of that source should be provided.
        \item If assets are released, the license, copyright information, and terms of use in the package should be provided. For popular datasets, \url{paperswithcode.com/datasets} has curated licenses for some datasets. Their licensing guide can help determine the license of a dataset.
        \item For existing datasets that are re-packaged, both the original license and the license of the derived asset (if it has changed) should be provided.
        \item If this information is not available online, the authors are encouraged to reach out to the asset's creators.
    \end{itemize}

\item {\bf New assets}
    \item[] Question: Are new assets introduced in the paper well documented and is the documentation provided alongside the assets?
    \item[] Answer: \answerYes{} 
    \item[] Justification: The primary new asset is our benchmark, which is open source and can be accessed on \href{https://github.com/CoolDawnAnt/InfoChartQA}{https://github.com/CoolDawnAnt/InfoChartQA}. The documentation is provided alongside the benchmark.
    \item[] Guidelines:
    \begin{itemize}
        \item The answer NA means that the paper does not release new assets.
        \item Researchers should communicate the details of the dataset/code/model as part of their submissions via structured templates. This includes details about training, license, limitations, etc. 
        \item The paper should discuss whether and how consent was obtained from people whose asset is used.
        \item At submission time, remember to anonymize your assets (if applicable). You can either create an anonymized URL or include an anonymized zip file.
    \end{itemize}

\item {\bf Crowdsourcing and research with human subjects}
    \item[] Question: For crowdsourcing experiments and research with human subjects, does the paper include the full text of instructions given to participants and screenshots, if applicable, as well as details about compensation (if any)? 
    \item[] Answer: \answerYes{} 
    \item[] Justification: The text of instructions refers to our open-sourced benchmark, and the interface screenshots are provided in Appendix~\ref{app:interface}
    \item[] Guidelines:
    \begin{itemize}
        \item The answer NA means that the paper does not involve crowdsourcing nor research with human subjects.
        \item Including this information in the supplemental material is fine, but if the main contribution of the paper involves human subjects, then as much detail as possible should be included in the main paper. 
        \item According to the NeurIPS Code of Ethics, workers involved in data collection, curation, or other labor should be paid at least the minimum wage in the country of the data collector. 
    \end{itemize}

\item {\bf Institutional review board (IRB) approvals or equivalent for research with human subjects}
    \item[] Question: Does the paper describe potential risks incurred by study participants, whether such risks were disclosed to the subjects, and whether Institutional Review Board (IRB) approvals (or an equivalent approval/review based on the requirements of your country or institution) were obtained?
    \item[] Answer: \answerYes{} 
    \item[] Justification: We provide an equivalent approval/review based on the requirements of our country or institution in the supplementary materials.
    \item[] Guidelines:
    \begin{itemize}
        \item The answer NA means that the paper does not involve crowdsourcing nor research with human subjects.
        \item Depending on the country in which research is conducted, IRB approval (or equivalent) may be required for any human subjects research. If you obtained IRB approval, you should clearly state this in the paper. 
        \item We recognize that the procedures for this may vary significantly between institutions and locations, and we expect authors to adhere to the NeurIPS Code of Ethics and the guidelines for their institution. 
        \item For initial submissions, do not include any information that would break anonymity (if applicable), such as the institution conducting the review.
    \end{itemize}

\item {\bf Declaration of LLM usage}
    \item[] Question: Does the paper describe the usage of LLMs if it is an important, original, or non-standard component of the core methods in this research? Note that if the LLM is used only for writing, editing, or formatting purposes and does not impact the core methodology, scientific rigorousness, or originality of the research, declaration is not required.
    \item[] Answer: \answerYes{} 
    \item[] Justification: Our benchmark is designed for multimodal large language models(MLLMs), so it requires conducting experiments on MLLMs and analyzing the results.
    \item[] Guidelines:
    \begin{itemize}
        \item The answer NA means that the core method development in this research does not involve LLMs as any important, original, or non-standard components.
        \item Please refer to our LLM policy (\url{https://neurips.cc/Conferences/2025/LLM}) for what should or should not be described.
    \end{itemize}

\end{enumerate}

\appendix

\newpage
\section{Dataset Construction}\label{append:data}
This section provides more detail on our dataset construction, including specific chart types~\ref{ssec:types}, infographic chart selection prompts~\ref{ssec:selection_prompt}, text-based question template~\ref{ssec:template}, and visual-element-based question template and examples~\ref{ssec:visual_sample}.

\subsection{Chart Types}\label{ssec:types}

\begin{figure}[!htbp]
    \centering
    \includegraphics[width=1\textwidth]{pictures/appendix_A/chart_distribution.pdf}
    \caption{Distribution of 54 chart types}
    \label{fig:distribution}
\end{figure}

\begin{table}[!thbp]
\centering
\caption{Chart Type and Their Frequencies}
\resizebox{0.7\textwidth}{!}{%
\normalsize
\begin{tabular}{llll}
\hline
\textbf{No.} & \textbf{Chart Type} & \textbf{Note} & \textbf{Count} \\
\hline
1 & Vertical simple bar chart & Bar & 543 \\
2 & Vertical category bar chart & Bar & 30 \\
3 & Vertical grouped bar chart & Bar & 34 \\
4 & Vertical stacked bar chart & Bar & 30 \\
5 & Horizontal simple bar chart & Bar & 877 \\
6 & Horizontal category bar chart & Bar & 43 \\
7 & Horizontal grouped bar chart & Bar & 39 \\
8 & Horizontal stacked bar chart & Bar & 31 \\
9 & Polar simple bar chart & Bar & 33 \\
10 & Polar category bar chart & Bar & 39 \\
11 & Polar stacked bar chart & Bar & 45 \\
12 & Radial simple bar chart & Bar & 33 \\
13 & Radial grouped bar chart & Bar & 38 \\
14 & Radial stacked bar chart & Bar & 49 \\
15 & Spiral simple bar chart & Bar & 33 \\
16 & Spiral complex chart & Bar & 30 \\
17 & Simple line chart & Line & 322 \\
18 & Grouped line chart & Line & 367 \\
19 & Simple area chart & Line & 44 \\
20 & Grouped area chart & Line & 63 \\
21 & Simple sparkline chart & Line & 48 \\
22 & Grouped sparkline chart & Line & 34 \\
23 & Simple spline chart & Line & 35 \\
24 & Grouped spline chart & Line & 32 \\
25 & Simple donut chart & Pie & 560 \\
26 & Simple pie chart & Pie & 417 \\
27 & Simple proportion chart & Proportion & 113 \\
28 & Grouped proportion chart & Proportion & 39 \\
29 & Categorical proportion chart & Proportion & 37 \\
30 & Funnel chart & Funnel/Pyramid & 35 \\
31 & Funnel diagram & Funnel/Pyramid & 54 \\
32 & Pyramid chart & Funnel/Pyramid & 30 \\
33 & Pyramid diagram & Funnel/Pyramid & 60 \\
34 & Angular gauge & Gauge & 75 \\
35 & Solid gauge chart & Gauge & 71 \\
36 & Text-based map & Map & 30 \\
37 & Value-based map & Map & 93 \\
38 & Matrix & Matrix & 30 \\
39 & Simple radar chart & Radar & 75 \\
40 & Grouped radar chart & Radar & 36 \\
41 & Sankey diagram & Sankey & 36 \\
42 & Simple scatter plot & Scatter plot & 206 \\
43 & Grouped scatter plot & Scatter plot & 174 \\
44 & Linear process timeline chart & Timeline & 58 \\
45 & Vertical timeline chart & Timeline & 31 \\
46 & Horizontal timeline chart & Timeline & 39 \\
47 & S-shape timeline chart & Timeline & 33 \\
48 & Convex treemap chart & Treemap & 46 \\
49 & One-layer convex circle treemap chart & Treemap & 32 \\
50 & Multi-layer convex circle treemap chart & Treemap & 41 \\
51 & One-layer treemap chart & Treemap & 58 \\
52 & Multi-layer treemap chart & Treemap & 40 \\
53 & Homogeneous multi-panel chart & Multi-panel & 398 \\
54 & Heterogeneous multi-panel chart & Multi-panel & 129 \\
\hline
&Total&&\TotfactImages\\
\hline
\end{tabular}
}
\end{table}\clearpage\FloatBarrier
\subsection{Infographic Chart Selection Prompts}\label{ssec:selection_prompt}
\begin{tcolorbox}[colframe=black,colback=black!10,title=Infographic Chart Selection Prompt]
You are a professional infographic designer with extensive expertise in infographics and data visualization. Your task is to analyze the given infographic image and provide a detailed assessment in the specified format.\\

\#\#\# Definitions:\\
**Please keep in mind the following definitions.**

1. Visualization types: \\
Funnel Chart, Pyramid Chart, Line Graph, Sankey Diagram, Area Chart, Radar Chart, Radial Bar Chart, Bar Chart, Icicle Diagram, Heat Map, Treemap, Pie Chart, Donut Chart, Scatter Plot, Dot Chart, Bubble Chart, Map, Arc Diagram, Chord Diagram, Matrix Diagram, Boxplot, Timeline, Gauge, Parallel Coordinates, Set Visualization, Contour Plot, Node-link Diagram, Dendrogram......

2. Data types of a visualization include the following:\\
- Single value: only a single value is displayed, such as a gauge or a single proportion or quantity\\
- Tabular data: structured data, such as a bar chart, line chart, or scatter plot\\
- Network data: data that represents relationships between entities, often visualized by a node-link diagram\\
- Hierarchical data: data with a hierarchical structure, primarily a tree structure\\
- Set data: data that represents sets and their relationships, such as a set visualization or a Venn diagram\\
- Geographic data: data that is presented by a map\\
- Descriptive (Textual) data: data that is primarily text-based, such as a word cloud, a timeline, or instructions (steps) for a process\\
3. Composite visualizations combine multiple visual representations of data into a cohesive and aesthetically meaningful layout, utilizing techniques such as juxtaposition, overlay, or nesting. Infographics or posters with multiple titles + charts are often not composite visualizations unless they are in the form of shared axes, connecting lines, cell arrangements, repeating styles, and so on.\\

\#\#\# Task:\\
Please analyze the image and output the results based on the following JSON format.\\

\#\#\# Output Format:\\
Reply in the following JSON format:

\{\\
    "title": , // title of the infographic, if no visible title, summarize one for it\\
    "description": , // describe the infographic\\
    "keywords": [kw1, ...], // give a maximum of five keywords that best describe the detailed theme of the infographic\\
    "domain": , // one-word domain of the infographic\\
    "language": , // language of the infographic\\
    "style": , // design style of the infographic\\
    "vis\_type": ["vis\_type1", ...], // give the different visualization types present in the image: you need to choose from the visualization types given, and can only choose a maximum of **three** answers if there are more than one, answer other if you cannot classify as any of the provided visualization types\\
    "data\_type": ["data\_type1", ...], // give the different data types present in data visualization(s): you need to choose from the data types given\\
    "composite": "yes/no", // analyze if this image contains a composite visualization\\
\}\\

\#\#\# Additional Guidelines:

Ensure your evaluation is concise and follows the format for consistency and accuracy.

\end{tcolorbox}
\clearpage
\subsection{Text-based Question Template}\label{ssec:template}

We designed question templates based on data facts, as shown in Table~\ref{tab:datafact_template}, which are suitable for charts with different data formats, including simple, stacked, grouped, and with-category.

{\small
\begin{longtable}{@{}p{1.6cm}p{3.2cm}p{2.8cm}p{5.2cm}@{}}
\captionsetup{width=1\linewidth}
\caption{Templates for text-based questions} \\
\hline
\toprule
\textbf{Data fact} & \textbf{	
Question type}  &\textbf{Question template} & \textbf{Instructions} \\
\midrule
\endfirsthead
\hline
\toprule
\textbf{Data fact} & \textbf{	
Question type}  & \textbf{Question} & \textbf{Instructions} \\
\midrule
\endhead 
\multicolumn{3}{r@{}}{continued \ldots}\\
\endfoot
\endlastfoot
\multirow{2}{1cm}{Value}& \multirow{2}{2.1cm}{value\_single\_element} & \multirow{2}{3cm}{What is the \{y\_label\} of \{ith\_label\}?} & * Your response should only contain the value of \{y\_label\} corresponding to \{ith\_tick\}. \\
 & & & * If there is an explicit answer in the chart, answer in the same format. \\
\midrule 
\multirow{2}{1cm}{Value} & \multirow{2}{2.1cm}{value\_element\_of\_group} & \multirow{2}{3cm}{What is the \{y\_label\} of \{ith\_label\}'s \{jth\_group\}?} & * Your response should only contain the value of \{y\_label\} corresponding to \{ith\_label\}'s \{jth\_group\}. \\
 &  &  & * If there is an explicit answer in the chart, answer in exactly the same format. \\
\midrule
\multirow{3}{1.6cm}{Difference} & \multirow{3}{2.1cm}{difference\_elements} &\multirow{3}{3cm}{What is the difference between the \{y\_label\} of \{ith\_label\} and \{jth\_label\}?
} & * Your response should only contain the value of the difference between the \{y\_label\} corresponding to \{ith\_label\} and \{jth\_label\}. \\
 &  & & * The answer you give me should be the absolute value. \\
 &  & & * The format of the difference you provide must be consistent with the corresponding data format in the chart. \\
\midrule
\multirow{3}{1.6cm}{Difference} & \multirow{3}{2.1cm}{difference\_group} & \multirow{3}{3cm}{What is the difference between the \{y\_label\} of \{ith\_label\}'s \{kth\_group\} and \{jth\_label\}'s \{kth\_group\}?
} & * Your response should only contain the value of the difference between the \{y\_label\} of \{ith\_label\}'s \{kth\_group\} and \{jth\_label\}'s \{kth\_group\}. \\
 &  & &* The answer you give me should be the absolute value. \\
 &  & &* The format of the difference you provide must be consistent with the corresponding data format in the chart. \\
\midrule
\multirow{3}{1.6cm}{Difference} & \multirow{3}{2.1cm}{difference\_two\_group} & \multirow{3}{3cm}{What is the difference between the \{y\_label\} of \{ith\_label\}'s \{jth\_group\} and \{ith\_label\}'s \{kth\_group\}?
} & * Your response should only contain the value of the difference between the \{y\_label\} of \{ith\_label\}'s \{jth\_group\} and \{ith\_label\}'s \{kth\_group\}. \\
 &  & & * The answer you give me should be the absolute value. \\
 &  &  &* The format of the difference you provide must be consistent with the corresponding data format in the chart. \\
\midrule
\multirow{2}{1.6cm}{Difference} & \multirow{2}{2.1cm}{difference\_yesno} & \multirow{2}{3cm}{Is the \{y\_label\} in \{ith\_label\} less than that in \{jth\_label\}?
} & * If the \{y\_label\} in \{ith\_label\} is less than that in \{jth\_label\}, your response should be 'Yes', otherwise 'No'. \\
 &  & & * Your response should only be 'Yes' or 'No'. \\
\midrule
\multirow{2}{2.1cm}{Difference} &   \multirow{2}{2.1cm}{difference\_in\_group\_yesno} & \multirow{2}{3cm}{Is the \{y\_label\} in \{ith\_label\}'s \{kth\_group\} less than that in \{jth\_label\}'s \{kth\_group\}?
} & * If the \{y\_label\} in \{ith\_label\}'s \{kth\_group\} is less than that in \{jth\_label\}'s \{kth\_group\}, your response should be 'Yes', otherwise 'No'. \\
 &  & & * Your response should only be 'Yes' or 'No'. \\
 \midrule
\multirow{2}{2.1cm}{Difference} &   \multirow{2}{2.1cm}{difference\_groups\_yesno} & \multirow{2}{3cm}{Is the \{y\_label\} in \{ith\_label\}'s \{jth\_group\} less than that in \{ith\_label\}'s \{kth\_group\}?
} & * If the \{y\_label\} in \{ith\_label\}'s \{jth\_group\} is less than that in \{ith\_label\}'s \{kth\_group\}, your response should be 'Yes', otherwise 'No'. \\
 &  & & * Your response should only be 'Yes' or 'No'. \\
\midrule
\multirow{2}{2.1cm}{Proportion} & \multirow{2}{2.1cm}{proportion\_element}  &\multirow{2}{3cm}{What is the proportion of \{ith\_label\} in \{father\_name\}?} & * Your response should only contain the proportion of \{ith\_tick\} in \{father\_name\}. \\
 &  &  &* If there is an explicit answer in the chart, answer in the same format. \\
\midrule 
\multirow{3}{1cm}{Trend} & \multirow{3}{2cm}{trend\_description} & \multirow{3}{3cm}{What is the trend of \{ith\_group\} in this chart?
} & * Your response must be a sequence of trends in chronological order. \\
 &  &  &* Possible trend values: 'increase', 'decrease', 'stable', 'oscillating', 'cyclicality', 'complex'. \\
 &  &  &* Example format: 'increase, decrease, stable' \\
\midrule
\multirow{3}{2cm}{Categorization} & \multirow{3}{2cm}{categorization\_target} &  \multirow{3}{3cm}{Which \{x\_label\}(s) \{['less than \{ith\_label\}','greater than \{ith\_label\}']\}?
} & * Your response should only contain the \{x\_label\} which have \{y\_label\} \{['less than \{ith\_label\}','greater than \{ith\_label\}']\}. \\
 &  & &* Separate the answers with commas. \\
 &  & &* If there is no answer that meets the condition, respond with an empty string. \\
\midrule

\multirow{3}{2cm}{Categorization}& \multirow{3}{2cm}{categorization\_in\_group} & \multirow{3}{3cm}{What is/are the \{x\_label\} which have \{ith\_group\} \{['less','greater']\} than \{jth\_label\}?
} & * Your final answer should only contain the \{x\_label\} which have \{ith\_group\} \{['less','greater']\} than \{jth\_label\}. \\
 &  &  &* Please provide your answer in the order from left to right, top to bottom, as they appear in the chart. \\
 &  &  &* If there is no answer that meets the condition, respond with an empty string. \\
\midrule

\multirow{3}{2cm}{Categorization} & \multirow{3}{2cm}{categorization\_groups} & \multirow{3}{3cm}{Which \{x\_label\} have \{ith\_group\} \{['less','greater']\} than \{jth\_group\}?
} & * Your response should only contain the \{x\_label\} which have \{ith\_group\} \{['less','greater']\} than \{jth\_group\}. \\
 &  & &* Please provide your answer in the order from left to right, top to bottom, as they appear in the chart. \\
 &  & &* If there is no answer that meets the condition, respond with an empty string. \\
\midrule

\multirow{3}{2cm}{Categorization} & \multirow{3}{2cm}{categorization\_category}& \multirow{3}{3cm}{Which \{x\_label\} in \{ith\_category\} have \{y\_label\} \{['less','greater']\} than \{bound\_value\} ?
} & * Your response should only contain the \{x\_label\} in the \{ith\_category\} with \{y\_label\} \{['less','greater']\} than \{bound\_value\}. \\
 &  & &* Please provide your answer in the order from left to right, top to bottom, as they appear in the chart. \\
 &  & &* If there is no answer that meets the condition, respond with an empty string. \\
\midrule

\multirow{3}{2cm}{Categorization} & \multirow{3}{2cm}{categorization\_in\_category} & \multirow{3}{3cm}{What is/are the \{x\_label\} which is/in \{ith\_category\} ?
} & * Your response should only contain the \{x\_label\} is/in \{ith\_category\}. \\
 &  &  &* Please provide your answer in the order from left to right, top to bottom, as they appear in the chart. \\
 &  &  &* If there is no answer that meets the condition, respond with an empty string. \\
\midrule
\multirow{1}{2cm}{Aggregation} & \multirow{1}{2cm}{aggregation\_sum} & \multirow{1}{3cm}{What is the sum of \{y\_label\}?
} & * Your response should only contain the value of the sum of \{y\_label\}. \\
\midrule

\multirow{1}{2cm}{Aggregation} & \multirow{1}{2cm}{aggregation\_average} & \multirow{1}{3cm}{What is the average \{y\_label\} per \{x\_label\}?
} & * Your response should only contain the value of the average of \{y\_label\} per \{x\_label\}. \\
\midrule

\multirow{1}{2cm}{Aggregation} & \multirow{1}{2cm}{aggregation\_median}  & \multirow{1}{3cm}{What is the median \{y\_label\}?
} & * Your response should only contain the value of the median of \{y\_label\}. \\
\midrule
\multirow{1}{1cm}{Aggregation} & \multirow{1}{2cm}{aggregation\_count}   & \multirow{1}{3cm}{How many data points are there?
} & * Your response should only contain the value of the number of data points in the chart. \\
\midrule
\multirow{4}{1cm}{Association} & \multirow{4}{2cm}{association\_correlation}& \multirow{4}{3cm}{What is the correlation between the \{y\_label\} of \{ith\_group\} and \{jth\_group\}?
} & * Your final response should be within a few words, such as "positively correlated", "negatively correlated", or "irrelevant". \\
 &  &  &* "positively correlated" if the correlation coefficient > 0.5, \\
 &  & &* "negatively correlated" if the correlation coefficient < -0.5, \\
 &  & &* "irrelevant" if the correlation coefficient is between -0.5 and 0.5. \\
\midrule

\multirow{4}{2cm}{Association} & \multirow{4}{2cm}{association\_groups}& \multirow{4}{3cm}{Do the distributions of the \{y\_label\} of \{ith\_group\} and \{jth\_group\} exhibit any distinct characteristics?
} & * Your final answer should be within a few words, such as "less", "greater", or "Not Applicable". \\
 &  &  &* If \{ith\_group\} generally less than \{jth\_group\}, Your final answer should be 'less'. \\
 &  & &* If \{ith\_group\} generally greater than \{jth\_group\}, Your final answer should be 'greater'. \\
 &  & &* Otherwise, your final answer should be 'Not Applicable' \\
\midrule
\multirow{2}{2cm}{Extreme} & \multirow{2}{2cm}{extreme\_element}  &\multirow{2}{3cm}{In which \{x\_label\} is the \{y\_label\} \{['minimum', 'maximum']\}?
} & * Your response should only contain the \{x\_label\} where \{y\_label\} is \{['minimum', 'maximum']\}. \\
 &  &   & * If there is an explicit answer in the chart, answer in exactly the same format. \\
\midrule

\multirow{2}{2cm}{Extreme}& \multirow{2}{2cm}{extreme\_value} & \multirow{2}{3cm}{What is the \{['minimum', 'maximum']\} value of \{y\_label\}?
} & * Your response should only contain the numerical value of the \{['minimum', 'maximum']\} \{y\_label\}. \\
 &  &  & * If there is an explicit answer in the chart, answer in exactly the same format. \\
\midrule
\multirow{3}{2cm}{Rank} & \multirow{3}{2cm}{rank\_by\_value} & \multirow{3}{3cm}{What is the order of \{x\_label\} on \{y\_label\} in ['increasing', 'decreasing'] order?
} & * Your final answer should only contain \{x\_label\} on \{y\_label\} in ['increasing',  'decreasing'] order. \\
 &  &  &* Separate the answers with commas. \\
   &  &  &* If there is an explicit answer in the chart, answer in exactly the same format. \\
\midrule
\multirow{2}{2cm}{Outlier} & \multirow{2}{2cm}{outlier\_identification} &\multirow{2}{3cm}{Is there an outlier in this chart? If yes, what is its name?
} & * Respond with 'No' if there is no outlier, otherwise provide the outlier's name." \\
 &  &  &* Your response should only be 'No' or the name of the outlier. \\
\midrule
\multirow{2}{2cm}{Distribution} & \multirow{2}{2cm}{distribution\_identification}&\multirow{2}{3cm}{Does the chart data show a significant statistical distribution? If yes, what type?
} & * Your response should be either 'No' if there's no significant distribution, or '[Distribution Type]' if there is one. \\
 &  &  & * Possible distribution types include: Uniform Distribution, Normal Distribution. \\
\bottomrule
\hline

\label{tab:datafact_template}
\end{longtable}
}

\newpage
\subsection{Visual-element-based Question Template and Examples}\label{ssec:visual_sample}

In this section, we provide additional information about our visual-element-based questions, including the template for generating visual-element-based basic questions (Table~\ref{tab:visual_basic_template}) and more examples of both basic and metaphor-related questions.

{\small
\begin{longtable}{@{}p{3.5cm}p{6.5cm}p{3cm}@{}}
\captionsetup{width=1\linewidth}
\caption{Templates for visual-element-based basic questions} \\
\hline
\toprule
\textbf{Question type} &\textbf{Question} & \textbf{Instructions} \\
\midrule
\endfirsthead
\hline
\toprule
\textbf{Question type} &\textbf{Question} & \textbf{Instructions} \\
\midrule
\endhead 
\multicolumn{2}{r@{}}{continued \ldots}\\
\endfoot
\endlastfoot
visual\_basic\_data\_value & \multirow{2}{6cm}{Read and understand the information presented in Figure 1 (a chart). Then, locate the specified icon in Figure 2. Identify the icon specified and provide the corresponding data point value based on the information from Figure 1. What is the value of the data point in Figure 1 corresponding to the specified icon in Figure 2? \newline [Figure 1: origin chart]\newline [Figure 2: cropped icon]} & * Your response should only contain the value of the data point corresponding to the icon specified in this chart.\\
 &  & * If there is an explicit answer in the chart, answer in exactly the same format. \\ 
\midrule 
visual\_basic\_data\_name & \multirow{2}{6cm}{Read and understand the information presented in Figure 1 (a chart). Then, locate the specified icon in Figure 2. Identify the icon specified and provide the corresponding data point based on the information from Figure 1. What is the name of the data point in Figure 1 corresponding to the specified icon in Figure 2?  \newline [Figure 1: origin chart]\newline [Figure 2: cropped icon]} & * Your response should only contain the name of the data point corresponding to the icon specified in this chart.\\
 &   & * If there is an explicit answer in the chart, answer in exactly the same format. \\
\midrule 
 visual\_basic\_group\_value & \multirow{2}{6cm}{Read and understand the information presented in Figure 1 (a chart). Then, locate the specified icon in Figure 2. Identify the icon specified and the corresponding data group based on the information from Figure 1. What is the \{ith\_label\} value of the data group in Figure 1 that corresponds to the specified icon in Figure 2?  \newline [Figure 1: origin chart] \newline [Figure 2: cropped icon]} & * Your response should only contain the value on/in/at \{ith\_label\} of the data group corresponding to the icon specified in this chart.\\
 &   & * If there is an explicit answer in the chart, answer in exactly the same format. \\
\midrule 
visual\_basic\_difference & \multirow{3}{6cm}{Read and understand the information presented in Figure 1 (a chart). Then, locate the specified icon1 and icon2 in Figure 2 and Figure 3. Identify the icons specified and provide the corresponding data point based on the information from Figure 1. What is the difference between the \{y\_label\} corresponding to icon1 and icon2? \newline [Figure 1: origin chart] \newline[Figure 2: cropped icon1]
\newline[Figure 3: cropped icon2]} & * Your response should only contain the value of the difference between the \{y\_label\} corresponding to icon1 and icon2.\\
 &  &* Your answer should be the absolute value of the difference, and its format must match the corresponding data format shown in the chart. \\
\midrule
visual\_basic\_difference\_yesno & \multirow{3}{6cm}{Read and understand the information presented in Figure 1 (a chart). Then, locate the specified icon1 and icon2 in Figure 2 and Figure 3. Identify the icons specified and provide the corresponding data point based on the information from Figure 1. Is the \{y\_label\} of the icon corresponding to Figure 2 less than that in of the icon corresponding to Figure 3? \newline [Figure 1: origin chart] \newline[Figure 2: cropped icon1]\newline [Figure 3: cropped icon2]} & * If the \{y\_label\} of the icon corresponding to Figure 2 is less than that of of the icon corresponding to Figure 3, your response should be 'Yes', otherwise 'No'.\\
 &  & * Your response should only contain 'Yes' or 'No'.\\
 & \\
\midrule 
visual\_basic\_data\_icon & \multirow{3}{6cm}{Which one of the four icons above best matches \{ith\_label\} based on the chart content? \newline [Figure 1: origin chart], \newline[Figure 2: cropped icon1]\newline [Figure 3: cropped icon2]
\newline[Figure 4: cropped icon3]
\newline[Figure 5: cropped icon4]} & * Think carefully based on the chart and the icons.\\
 &  &* Only output the final answer in the following format: [Number of the best matching icon]\\
 &  &* Do not output anything else besides the answer in the specified format.\\
 \midrule
 visual\_basic\_imagery & \multirow{1}{6cm}{Read and understand the information presented in Figure 1 (a chart). Then, locate the specified icon and Figure 2. What is the correct role of this icon in the chart? \newline (A) [..]\newline(B) [..]\newline(C) [..]\newline(D) [..]}& * Your response should be the letter only (e.g., ’C’). Do not include any explanation or repeat the option text.\\ \\ \\
 & \\
\midrule 
\bottomrule
\label{tab:visual_basic_template}
\end{longtable}
}

\begin{tcolorbox}[colframe=blue,colback=cyan!20,title=Example of visual-element-based (basic) questions]
   \begin{minipage}{0.50\textwidth}
        \centering
        \includegraphics[width=\linewidth,height=190pt]{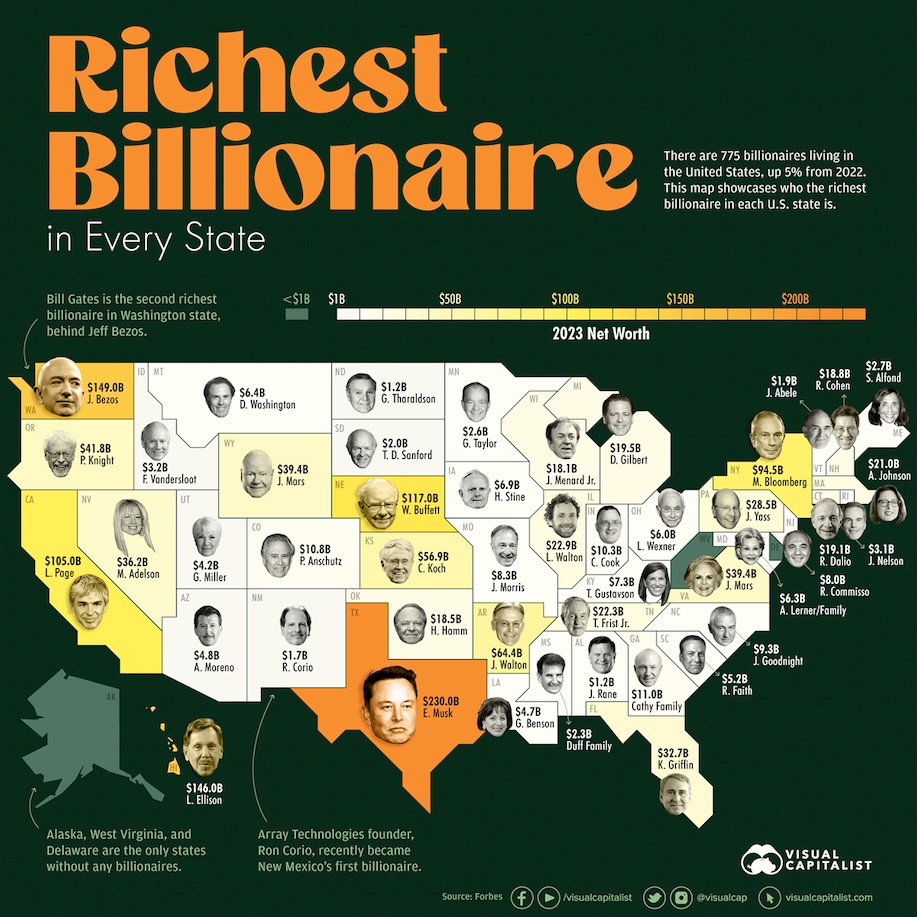}\\
    \end{minipage}
    \hspace{55pt} 
    \begin{minipage}{0.22\textwidth}
        \centering
        \includegraphics[width=\linewidth,height=100pt]{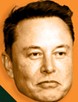}\\
    \end{minipage}
    
    \vspace{15pt}
    \textcolor{blue!60}{\textbf{Question:} }
    \\
    {\setlength{\spaceskip}{0.3em} \ttfamily Examine Figure 1 to familiarize yourself with the chart's details. Next, observe the icon highlighted in Figure 2. Using the information from Figure 1, determine the value associated with this specific icon. What is the data value linked to the icon shown in Figure 2 according to Figure 1?

    \vspace{10pt}
    * Your response should only contain the value of the data point corresponding to the icon specified in this chart.           

    \vspace{10pt}
    * If there is an explicit answer in the chart, answer in exactly the same format.

    \vspace{10pt}
    \textcolor{blue!60}{\textbf{Answer:} }{\ttfamily \$230.0B}

    }
\end{tcolorbox}
\newpage
\begin{tcolorbox}[colframe=blue,colback=cyan!20,title=Example of visual-element-based (basic) questions]
   \begin{minipage}{0.52\textwidth}
        \centering
        \includegraphics[width=\linewidth,height=200pt]{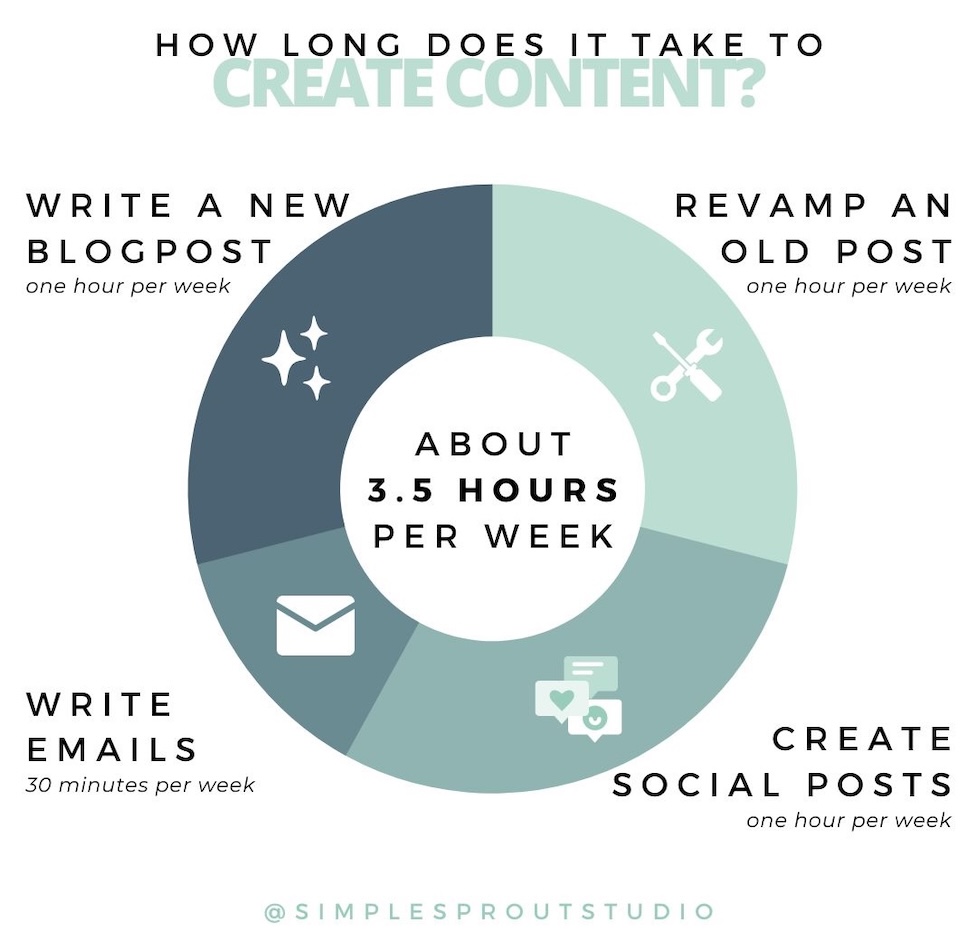}\\
    \end{minipage}
    \hspace{50pt} 
    \begin{minipage}{0.24\textwidth}
        \centering
        \includegraphics[width=\linewidth,height=90pt]{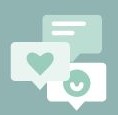}\\
    \end{minipage}
    
    \vspace{15pt}
    \textcolor{blue!60}{\textbf{Question:} }\\
    {\setlength{\spaceskip}{0.3em} \ttfamily Carefully examine the chart shown in Figure 1. Next, observe the icon illustrated in Figure 2 and find its match within Figure 1. What is the data value from Figure 1 that corresponds to the icon presented in Figure 2?
    
    \vspace{10pt}
    * Your response should only contain the value of the data point corresponding to the icon specified in this chart.           

    \vspace{10pt}
    * If there is an explicit answer in the chart, answer in exactly the same format.

    \vspace{10pt}
    \textcolor{blue!60}{\textbf{Answer:} }{\ttfamily one hour per week}

    }
\end{tcolorbox}
\newpage
\begin{tcolorbox}[colframe=blue,colback=cyan!20,title=Example of visual-element-based (basic) questions]
   \begin{minipage}{0.56\textwidth}
        \centering
        \includegraphics[width=\linewidth,height=140pt]{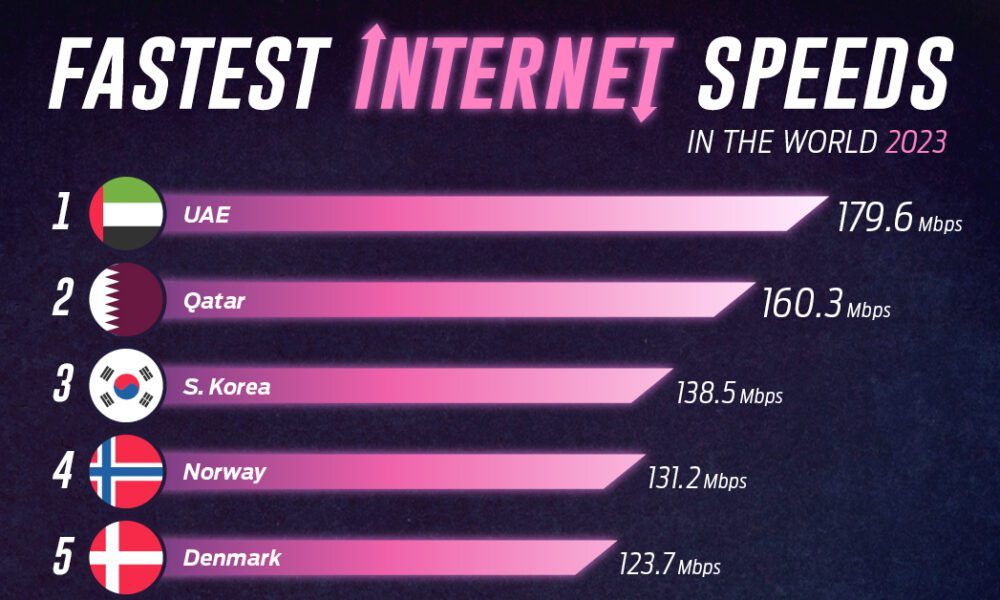}\\
    \end{minipage}
    \hspace{48pt} 
    \begin{minipage}{0.24\textwidth}
        \centering
        \includegraphics[width=\linewidth,height=90pt]{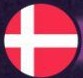}\\
    \end{minipage}
    
    \vspace{15pt}
    \textcolor{blue!60}{\textbf{Question:} }\\{\setlength{\spaceskip}{0.3em} \ttfamily Review the chart in Figure 1 and examine the icon displayed in Figure 2. Match the icon from Figure 2 to its position in Figure 1, then state the data value associated with it as shown in the chart.
    
    \vspace{10pt}
    * Your response should only contain the value of Speed (Mbps) corresponding to the icon specified.

    \vspace{10pt}
    * If there is an explicit answer in the chart, answer in exactly the same format.

    \vspace{10pt}
    \textcolor{blue!60}{\textbf{Answer:} }{\ttfamily 123.7}

    }
\end{tcolorbox}
\newpage
\begin{tcolorbox}[colframe=blue,colback=cyan!20,title=Example of visual-element-based (basic) questions]
   \begin{minipage}{0.48\textwidth}
        \centering
        \includegraphics[width=\linewidth,height=200pt]{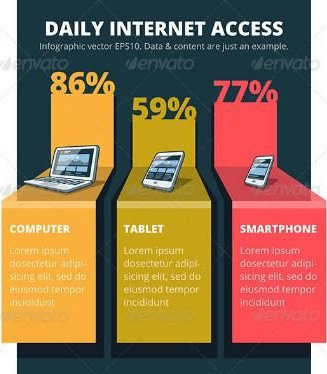}\\
    \end{minipage}
    \hspace{55pt} 
    \begin{minipage}{0.24\textwidth}
        \centering
        \includegraphics[width=\linewidth,height=70pt]{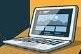}\\
    \end{minipage}
    
    \vspace{15pt}
    \textcolor{blue!60}{\textbf{Question:} }{\setlength{\spaceskip}{0.3em} 
    \\
    \ttfamily First, examine Figure 1 to interpret the chart details. Next, review Figure 2 to find the highlighted icon. Using the chart in Figure 1, determine the data value that matches the icon shown in Figure 2.
    
    \vspace{10pt}
    * Your response should only contain the value of Daily Internet Access (\%) corresponding to the icon specified.
    
    \vspace{10pt}
    * If there is an explicit answer in the chart, answer in exactly the same format.

    \vspace{10pt}
    \textcolor{blue!60}{\textbf{Answer:} }{\ttfamily 86}

    }
\end{tcolorbox}

\newpage
\begin{tcolorbox}[colframe=blue,colback=cyan!20,title=Example of visual-element-based (basic) questions]
   \begin{minipage}{0.50\textwidth}
        \centering
        \includegraphics[width=\linewidth,height=170pt]{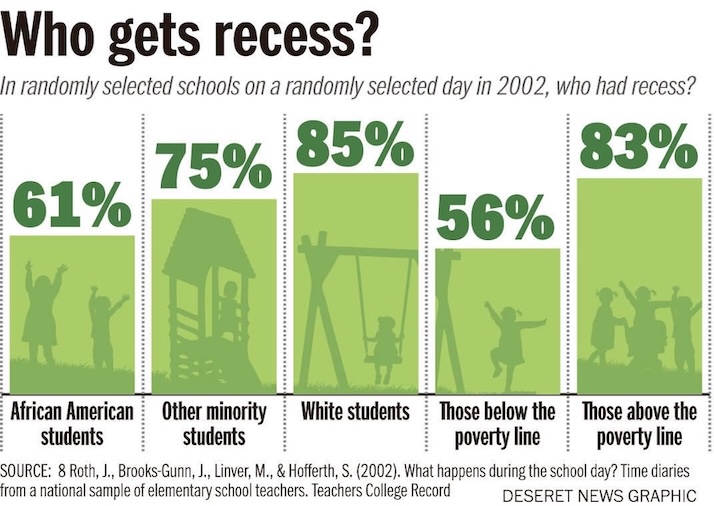}\\
    \end{minipage}
    \hspace{55pt} 
    \begin{minipage}{0.18\textwidth}
        \centering
        \includegraphics[width=\linewidth,height=150pt]{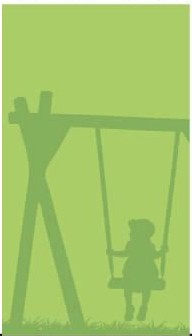}\\
    \end{minipage}
    
    \vspace{15pt}
    \textcolor{blue!60}{\textbf{Question:} }
    \\
    {\setlength{\spaceskip}{0.3em} \ttfamily The left image is a chart, and the right is an image cropped from that chart. What role does this image primarily play within the chart?

    \vspace{10pt}
    (A) To emphasize the chart's central theme concerning disparities in children's access to playtime.

    \vspace{10pt}
    (B) To serve primarily as a decorative background, enhancing the overall visual appeal of the graph.
    
    \vspace{10pt}
    (C) To visually clarify the concept of 'recess' that the 83\% statistic represents for this student group.
    
    \vspace{10pt}
    (D) To symbolically represent the social interaction and group activities common during school recess.
    
    \vspace{10pt}
    * Your response should be the letter only (e.g., 'C'). Do not include any explanation or repeat the option text.      
    
    \vspace{10pt}
    \textcolor{blue!60}{\textbf{Answer:} }{\ttfamily C}

    }
\end{tcolorbox}

\newpage
\begin{tcolorbox}[colframe=blue,colback=cyan!20,title=Example of visual-element-based (basic) questions]
   \begin{minipage}{0.52\textwidth}
        \centering
        \includegraphics[width=\linewidth,height=150pt]{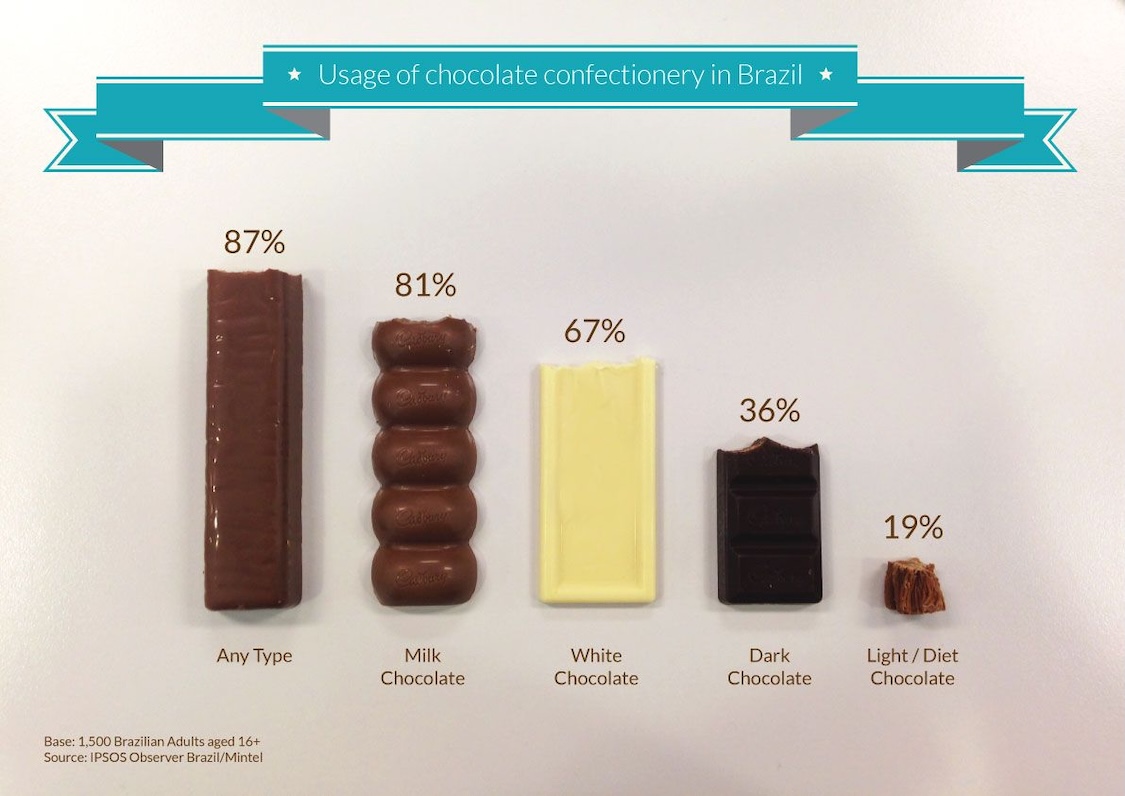}\\
    \end{minipage}
    \hspace{60pt} 
    \begin{minipage}{0.10\textwidth}
        \centering
        \includegraphics[width=\linewidth,height=140pt]{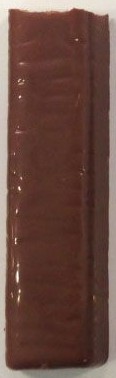}\\
    \end{minipage}
    
    \vspace{15pt}
    \textcolor{blue!60}{\textbf{Question:} }
    \\{\setlength{\spaceskip}{0.3em} \ttfamily The sequence above presents a chart followed by a crop of that chart; within the chart, what primary function does this cropped image serve?

    \vspace{10pt}
    (A) To provide purely aesthetic enhancement and make the chart more visually engaging.

    \vspace{10pt}
    (B) To serve as a direct visual representation for the \"Any Type\" category, clarifying the data point.
    
    \vspace{10pt}
    (C) To establish a specific brand association for the most popular chocolate category.
    
    \vspace{10pt}
    (D) To create an emotional connection by emphasizing the universal appeal of a classic chocolate bar. 
    
    \vspace{10pt}
    * Your response should be the letter only (e.g., 'C'). Do not include any explanation or repeat the option text.      
    
    \vspace{10pt}
    \textcolor{blue!60}{\textbf{Answer:} }{\ttfamily B}

    }
\end{tcolorbox}

\newpage
\begin{tcolorbox}[colframe=blue,colback=cyan!20,title=Example of visual-element-based (metaphor) questions]
   \begin{minipage}{0.48\textwidth}
        \centering
        \includegraphics[width=\linewidth,height=200pt]{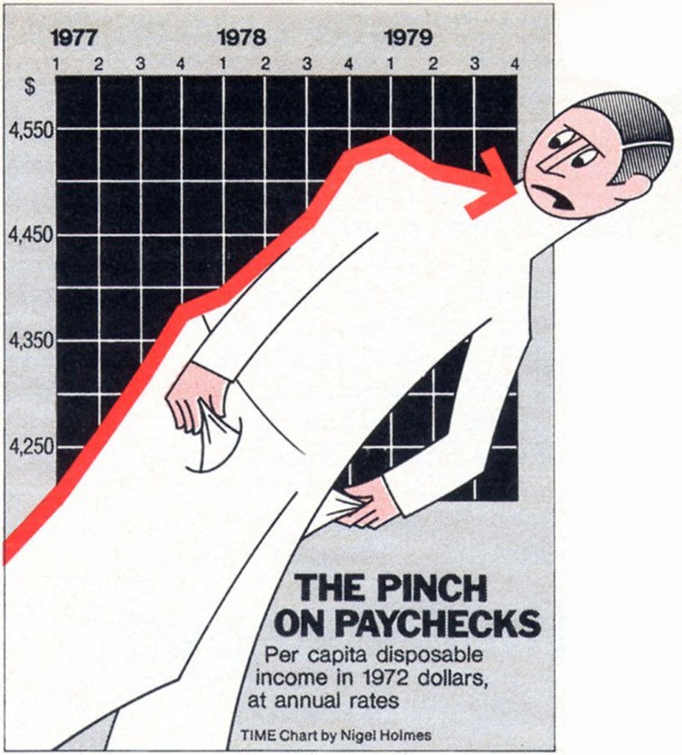}\\
    \end{minipage}
    \hspace{45pt} 
    \begin{minipage}{0.36\textwidth}
        \centering
        \includegraphics[width=\linewidth,height=130pt]{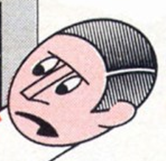}\\
    \end{minipage}
    
    \vspace{15pt}
    \textcolor{blue!60}{\textbf{Question:} }\\
    {\setlength{\spaceskip}{0.3em} \ttfamily Answer this question and choose the most appropriate answer (A, B, C, D).
    
    \vspace{10pt}
    Why might the person in the chart be wearing an unhappy facial expression?
       
    \vspace{10pt}
    (A) To show the person's anger because his packet has been stolen.

    \vspace{10pt}
    (B) To show the person's embarrassment because his packet is empty.
    
    \vspace{10pt}
    (C) To convey the anxiety associated with decreasing disposable income.
    
    \vspace{10pt}
    (D)To make the chart look more friendly.
    
    \vspace{10pt}
    * Your response should only contain 'A'/'B'/'C'/'D'. If your response contains multiple options, you should separate them by ','. Do not include any explanation, analysis, or repeat the option text.
    
    \vspace{10pt}
    \textcolor{blue!60}{\textbf{Answer:} }{\ttfamily C}

    }
\end{tcolorbox}

\newpage
\begin{tcolorbox}[colframe=blue,colback=cyan!20,title=Example of visual-element-based (metaphor) questions]
   \begin{minipage}{0.48\textwidth}
        \centering
        \includegraphics[width=\linewidth,height=200pt]{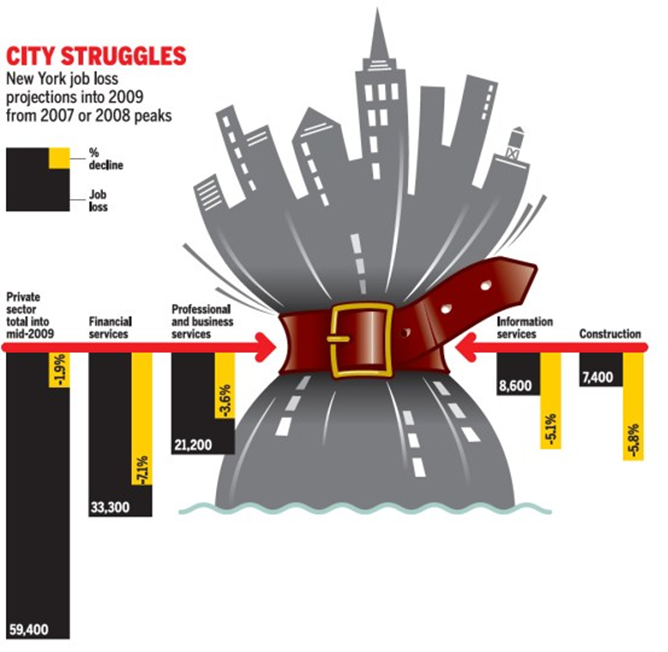}\\
    \end{minipage}
    \hspace{45pt} 
    \begin{minipage}{0.30\textwidth}
        \centering
        \includegraphics[width=\linewidth,height=150pt]{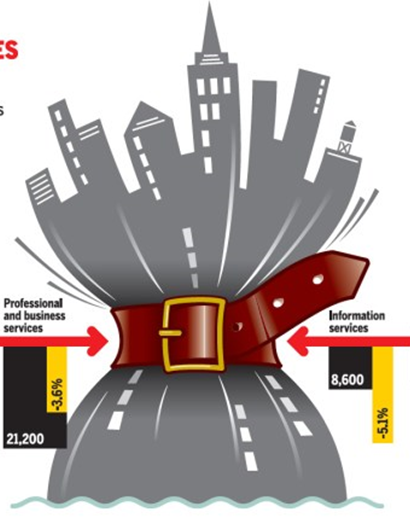}\\
    \end{minipage}
    
    \vspace{15pt}
    \textcolor{blue!60}{\textbf{Question:} 
    \\
    }{\setlength{\spaceskip}{0.3em} \ttfamily Answer this question and choose the most appropriate answer (A, B, C, D).
    
    \vspace{10pt}
    According to the chart, by drawing the buildings in the chart, what is the main idea the chart tries to convey?

    \vspace{10pt}
    (A) The job loss in New York is making it struggle.

    \vspace{10pt}
    (B) New York City is trying to ignore the damage caused by the job loss.
    
    \vspace{10pt}
    (C) The job loss is being solved due to the city's effort.
    
    \vspace{10pt}
    (D) "Private sector total into mid-2009" is facing the most job loss.
    
    \vspace{10pt}
    * Your response should be the letter only (e.g., 'C'). Do not include any explanation or repeat the option text.
    
    \vspace{10pt}
    \textcolor{blue!60}{\textbf{Answer:} }{\ttfamily A}

    }
\end{tcolorbox}
\newpage
\section{Additional Experimental Settings and Results}\label{append:exp}


\subsection{Test Configurations for MLLMs}\label{append:config}
We present our running configurations for each MLLM in Table~\ref{tab:run_config}. We conduct all experiments on a single server equipped with 8 $\times$ NVIDIA RTX 4090 GPUs.
\begin{table}[!htbp]
  \centering
  \caption{Running configurations for different MLLMs. Temp. denotes temperature.}
    \resizebox{1\textwidth}{!}{
    \normalsize
    \begin{tabular}{llcccc}
    \toprule
    \textbf{Model} & \textbf{Version/HF Checkpoint} & \textbf{Do Sample} & \textbf{Max New Tokens} &
    Temp. & Top-P \\
    \midrule
    GPT-4.1 & gpt-4.1 &       & 1024  & 0     & 1 \\       
    GPT-4o & gpt-4o-2024-08-06 &       & 1024  & 0     & 1 \\
    Claude 3.5 Sonnet  & claude-3-5-sonnet-20241022 &       & 1024  & 0     & 1 \\
    Gemini 2.5 Pro Preview & gemini-2.5-pro-preview-03-25 &       & 1024  & 0     & 1 \\
    Gemini 2.5 Flash Preview  & gemini-2.5-flash-preview-04-17 &       & 1024  & 0     & 1 \\
    Qwen2.5-VL-72B & Qwen/Qwen2.5-VL-72B-Instruct & False & 1024  & 0     & 1 \\
    Llama-4 Scout & meta-llama/Llama-4-Scout-17B-16E-Instruct & False & 1024  & 0     & 1 \\
    Intern-VL3 78B & OpenGVLab/InternVL3-78B & False & 1024  & 0     & 1 \\
    Intern-VL3 8B & OpenGVLab/InternVL3-8B & False & 1024  & 0     & 1 \\
    Janus Pro & deepseek-ai/Janus-Pro-7B & False & 512   & 0     & 1 \\
    DeepSeek VL2 & deepseek-ai/deepseek-vl2 & False & 512   & 0     & 1 \\
    Phi-4 & microsoft/Phi-4-multimodal-instruct & False & 1000  & 0     & 1 \\
    LLaVA OneVision Chat 78B  & llava-hf/llava-onevision-qwen2-72b-ov-chat-hf & False & 1024  & 0     & 1 \\
    LLaVA OneVision Chat 7B & llava-hf/llava-onevision-qwen2-7b-ov-chat-hf & False & 1024  & 0     & 1 \\
    Pixtral & mistralai/Pixtral-12B-2409 & False & 8192  & 0     & 1 \\
    Ovis1.6-Gemma2-9B & AIDC-AI/Ovis1.6-Gemma2-9B & False & 1024  & 0     & 1 \\
    ChartGemma & ahmed-masry/chartgemma & False & 512   & 0     & 1 \\
    TinyChart & mPLUG/TinyChart-3B-768 & False & 512   & 0     & 1 \\
    ChartInstruct-LLama2 & ChartInstruct-LLama2 & False & 512   & 0     & 1 \\
    \bottomrule
    \end{tabular}%
    }
  \label{tab:run_config}%
\end{table}%

\xie{
\subsection{Pseudocode for Evaluation}\label{append:eval code}
We provide the pseudocode for answer validation in this section, as shown in Algorithm~\ref{alg:app_evaluation}.
}

\subsection{Additional Results}\label{append:sample}
In this section, we present additional experimental results, including \xie{detailed breakdowns of the \changjian{quantitative MLLM evaluation results}~(\ref{sssec:detailed_main_result}) and more performance degradation results \changjian{when removing visual elements}~(\ref{app:supp_431}).}

\xie{
\subsubsection{Detailed Breakdowns of the Quantitative MLLM Evaluation Results}\label{sssec:detailed_main_result}
In this section, we provide detailed breakdowns of the \changjian{quantitative MLLM evaluation results}. 
We first categorize the infographic charts into three difficulty levels based on the average accuracy of Gemini 2.5 Pro Preview and GPT-4.1: Easy (above 80\%), Moderate (40–80\%), and Hard (below 40\%). 
The difficulty level of each sample is provided in the metadata of the dataset card.
This yields 46.7\% easy, 32.4\% moderate, and 20.9\% hard charts. 
Table~\ref{tab:app_main_res_levels} and Table~\ref{tab:app_main_res_levels_vis} report the performance of each model across the three difficulty levels. 
As shown, all models perform poorly on hard charts, while even domain-specific models such as ChartInstruct-LLama2 struggle to achieve strong results on the easy charts.
}

\xie{
\changjian{We also provide quantitative MLLM evaluation results by answer types and data-fact types}. 
For the answer types, we divide the answers into three categories: textual, numeric, and multiple-choice. 
The corresponding results are presented in Table~\ref{tab:app_main_res_types}.
For the data-fact types, we provide the performance of GPT4.1 on different data-fact types as shown in Table~\ref{tab:app_performance_on_qa_types}.
}

\begin{algorithm}
\caption{Answer Normalization and Evaluation}
\label{alg:app_evaluation}
\begin{algorithmic}[1]
\Require Predicted answer $a_{pred}$, ground truth $a_{gt}$
\Ensure Correct / Incorrect

\If{$a_{pred}$ is \textbf{text}}
    \State Convert all characters to lowercase
    \State Standardize special symbols (e.g., replace single quotes with double quotes)
    \State Remove leading, trailing, and consecutive whitespace
    \State Compute ANLS score between $a_{pred}$ and $a_{gt}$
    \If{ANLS $> 0.8$}
        \State \Return Correct
    \Else
        \State \Return Incorrect
    \EndIf
\ElsIf{$a_{pred}$ is \textbf{numeric}}
    \State Extract the numerical part
    \State Remove units (e.g., ``\$'')
    \State Convert into unified numerical format
    \State Compute relative error $\frac{|a_{pred} - a_{gt}|}{a_{gt}}$
    \If{relative error $< 0.05$}
        \State \Return Correct
    \Else
        \State \Return Incorrect
    \EndIf
\ElsIf{$a_{pred}$ is \textbf{multiple-choice}}
    \If{$a_{pred}$ option = $a_{gt}$ option}
        \State \Return Correct
    \Else
        \State \Return Incorrect
    \EndIf
\EndIf
\end{algorithmic}
\end{algorithm}


\begingroup 
\begin{table}[!b]
\caption{Evaluation result of text-based questions on \InfoChartQA. 
}
\resizebox{\textwidth}{!}{
\normalsize

\begin{tabular}{lccccccccc}
\toprule
  \rule[-0.5em]{0pt}{0.5em}\textbf{Model} & \multicolumn{9}{c}{\textbf{Text-based}} \\
 \cline{2-10}
 \rule[0.5em]{0pt}{0.6em} &  \multicolumn{4}{c}{\textbf{Infographic}} & \multicolumn{4}{c}{\textbf{Plain}} &  \\
  \cmidrule(r){2-5} \cmidrule(l){6-9}
\rule[0em]{0pt}{1em} &  \textbf{Easy} & \textbf{Moderate} & \textbf{Hard} & \textbf{Overall} & \textbf{Easy} & \textbf{Moderate} & \textbf{Hard} & \textbf{Overall} & $\Delta$ \\
\midrule
\multicolumn{10}{c}{\rule[-0.2em]{0pt}{0.8em}\textbf{Proprietary Models}}\\
\midrule
OpenAI O4-mini               &  85.68 & 63.45 & 29.00 & 76.23  &  97.46  & 81.81 & 46.66 & 89.62 & 13.39 \\
GPT-4.1                      &  83.17 & 60.38 & 14.28 & 71.29  &  89.24  & 71.42 & 40.11 & 80.81 & 9.52 \\
GPT-4o                       &  74.27 & 48.35 & 24.41 & 64.59  &  87.91  & 73.83 & 38.33 & 80.60 & 16.01 \\
Claude 3.5 Sonnet            &  75.32 & 44.16 & 26.67 & 62.80  &  91.01  & 67.89 & 45.71 & 81.37 & 18.57 \\
Gemini 2.5 Pro Preview       &  87.53 & 69.34 &	32.75 & 79.23  &  98.30  & 84.52 & 57.33 & 91.16 & 11.93 \\
Gemini 2.5 Flash Preview     &  84.81 & 57.14 & 19.98 & 72.40  &  87.91  & 75.13 & 31.67 & 80.56 & 8.16 \\
\midrule
\multicolumn{10}{c}{\rule[-0.2em]{0pt}{0.8em}\textbf{Open-Source Models}} \\
\midrule
Qwen2.5-VL-72B            &  74.70 & 47.84 & 22.64 & 61.08  &  85.30  & 67.93 & 39.59 & 77.92 & 16.84 \\
Llama-4 Scout             &  70.80 & 48.67 & 27.32 & 63.68  &  85.71  & 72.50 & 28.18 & 78.84 & 15.16 \\
Intern-VL3-78B            &  69.68 & 49.71 & 29.25 & 63.42  &  90.91  & 72.96 & 31.14 & 81.41 & 17.99 \\
Intern-VL3-8B             &  56.25 & 37.80 & 18.29 & 46.45  &  71.36  & 54.84 & 33.73 & 61.67 & 15.22 \\
Janus Pro                 &  32.97 & 21.20 & 20.33 & 27.89  &  40.80  & 29.17 & 24.27 & 35.88 & 7.99 \\
DeepSeek VL2              &  44.94 & 36.36 & 13.33 & 40.40  &  49.37  & 40.26 & 13.33 & 44.44 & 4.04  \\
Phi-4                     &  40.80 & 25.27 & 16.67 & 35.47  &  62.35  & 42.53 & 29.77 & 54.68 & 19.21 \\
LLaVA OneVision Chat 78B  &  54.91 & 33.19 & 21.91 & 44.69  &  66.58  & 49.57 & 39.66 & 58.51 & 13.82 \\
LLaVA OneVision Chat 7B   &  43.63 & 27.92 & 21.84 & 36.45  &  57.56  & 42.86 & 36.03 & 50.47 & 14.02 \\
Pixtral                   &  56.64 & 34.86 & 26.05 & 46.61  &  59.85  & 58.00 & 35.00 & 59.29 & 12.68 \\
Ovis1.6-Gemma2-9B         &  63.09 & 39.21 & 25.02 & 51.69  &  67.50  & 49.75 & 34.52 & 58.66 & 6.97 \\
ChartGemma                &  25.51 & 18.18 & 18.26 & 22.42  &  36.23  & 29.22 & 19.60 & 33.33 & 10.91 \\
TinyChart                 &  28.23 & 18.38 & 21.66 & 24.32  &  48.34  & 35.55 & 36.05 & 42.97 & 18.65 \\
ChartInstruct-LLama2      &  22.18 & 16.10 & 20.20 & 19.95  &  31.22  & 20.44 & 22.95 & 26.87 & 6.92  \\
\bottomrule
\end{tabular}
}
\label{tab:app_main_res_levels}
\end{table}
\endgroup 

\begingroup 
\begin{table}[!b]
\centering
\caption{Evaluation result of visual-element-based basic questions on \InfoChartQA.
}
\resizebox{0.51\textwidth}{!}{

\normalsize
\centering
\begin{tabular}{lccc}
\toprule
  \textbf{Model} & \textbf{Easy} & \textbf{Moderate} & \textbf{Hard} \\
\midrule
\multicolumn{4}{c}{\rule[-0.2em]{0pt}{0.8em}\textbf{Proprietary Models}}\\
\midrule
OpenAI O4-mini               &  92.07 & 91.83 & 90.33 \\
GPT-4.1                      &  88.25 & 87.69 & 86.84 \\
GPT-4o                       &  82.39 & 79.45 & 78.92 \\
Claude 3.5 Sonnet            &  90.45 & 89.12 & 89.98 \\
Gemini 2.5 Pro Preview       &  90.56 & 89.39 & 89.36 \\
Gemini 2.5 Flash Preview     &  82.03 & 81.54 & 80.91 \\
\midrule
\multicolumn{4}{c}{\rule[-0.2em]{0pt}{0.8em}\textbf{Open-Source Models}} \\
\midrule
Qwen2.5-VL-72B            &  79.09 & 77.95 & 75.50 \\
Llama-4 Scout             &  82.32 & 79.86 & 76.88  \\
Intern-VL3-78B            &  79.82 & 81.33 & 77.66 \\
Intern-VL3-8B             &  77.71 & 74.99 & 71.93 \\
Janus Pro                 &  40.77 & 40.22 & 46.39 \\
DeepSeek VL2              &  63.33 & 60.06 & 56.71 \\
Phi-4                     &  63.27 & 61.66 & 59.81 \\
LLaVA OneVision Chat 72B  &  62.12 & 62.08 & 58.53 \\
LLaVA OneVision Chat 7B   &  60.78 & 60.54 & 56.55 \\
Pixtral                   &  60.07 & 67.36 & 73.09 \\
Ovis1.6-Gemma2-9B         &  62.42 & 60.89 & 57.61 \\
ChartGemma                &  30.96 & 28.28 & 36.45 \\
TinyChart                 &  17.47 & 13.71 & 15.69 \\
ChartInstruct-LLama2      &  32.80 & 33.10 & 40.36 \\
\bottomrule
\end{tabular}
}
\label{tab:app_main_res_levels_vis}
\end{table}
\endgroup 

\begin{table}[!b]
\centering
\caption{Evaluation result on \InfoChartQA with different answer types. 
}
\resizebox{0.6\textwidth}{!}{
\centering
\begin{tabular}{lccc}
\toprule
\textbf{Model} & \textbf{Numeric} & \textbf{Textual} & \textbf{Multiple-choice} \\
\midrule
\multicolumn{4}{c}{\textbf{Proprietary Models}} \\
\midrule
OpenAI O4-mini           & 83.38 & 78.75 & 73.36 \\
GPT-4.1                  & 74.51 & 67.85 & 70.29 \\
GPT-4o                   & 65.01 & 58.00 & 63.71 \\
Claude 3.5 Sonnet        & 61.52 & 64.65 & 72.26 \\
Gemini 2.5 Pro Preview   & 80.79 & 81.17 & 72.58 \\
Gemini 2.5 Flash Preview & 73.04 & 70.71 & 62.25 \\
\midrule
\multicolumn{4}{c}{\textbf{Open-Source Models}} \\
\midrule
Qwen2.5-VL-72B           & 62.85 & 57.48 & 62.48 \\
Llama-4 Scout            & 60.03 & 68.83 & 56.49 \\
Intern-VL3-78B           & 65.12 & 59.22 & 66.21 \\
Intern-VL3-8B            & 51.79 & 38.74 & 62.94 \\
Janus Pro                & 31.55 & 19.81 & 36.96 \\
DeepSeek VL2             & 43.28 & 35.35 & 47.72 \\
Phi-4                    & 23.29 & 43.88 & 48.43 \\
LLaVA OneVision Chat 72B & 48.91 & 35.48 & 57.04 \\
LLaVA OneVision Chat 7B  & 41.58 & 26.28 & 57.17 \\
Pixtral                  & 52.37 & 34.38 & 59.69 \\
Ovis1.6-Gemma2-9B        & 49.74 & 54.63 & 57.31 \\
ChartGemma               & 28.01 & 11.13 & 30.51 \\
TinyChart                & 30.58 & 11.17 & 13.75 \\
ChartInstruct-LLama2     & 24.73 & 9.59  & 32.29 \\
\bottomrule
\end{tabular}
}
\label{tab:app_main_res_types}
\end{table}

\begin{table}[h]
    \centering
    \caption{Models' performance on different data-fact types.}
    \resizebox{1\textwidth}{!}{
    \begin{tabular}{lcccccccccccc}
    \toprule
    \textbf{Model} & 
    \textbf{Outlier} & 
    \textbf{Extreme} & 
    \textbf{Association} & 
    \textbf{Trend} & 
    \textbf{Value} & 
    \textbf{Rank} & 
    \textbf{Difference} & 
    \textbf{Categorization} & 
    \textbf{Distribution} & 
    \textbf{Aggregation} & 
    \textbf{Proportion} \\
    \midrule
    GPT-4.1 & 28.1 & 75.2 & 51.0 & 53.8 & 81.6 & 37.6 & 84.1 & 58.6 & 93.3 & 64.2 & 95.9 \\
    \bottomrule
    \end{tabular}%
    }
    \label{tab:app_performance_on_qa_types}%
\end{table}

\subsubsection{Additional Performance Results When Removing \fix{Visual} Elements}\label{app:supp_431}

\begin{figure}[!htbp]
\centering
    \begin{subfigure}{0.35\textwidth}
    \centering
        \includegraphics[width=1\textwidth]{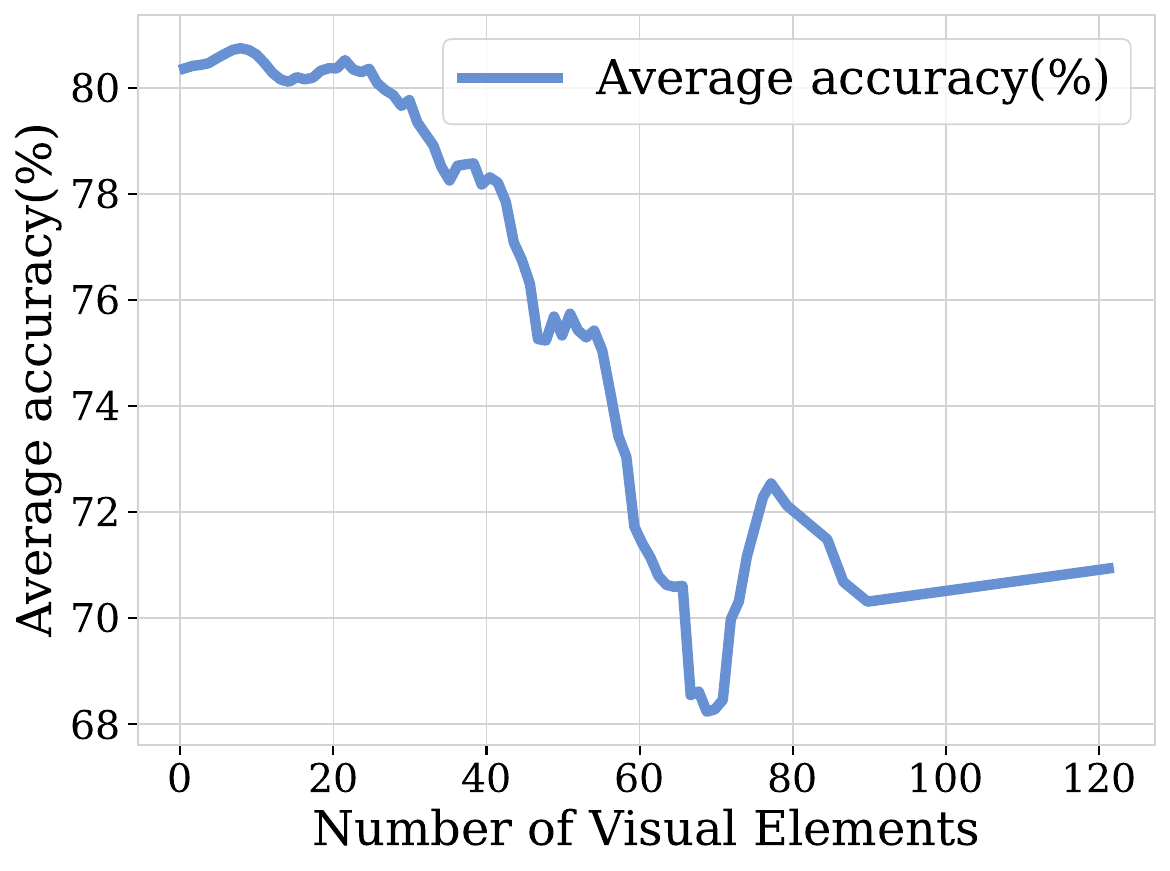}
        \caption{GPT-4.1}
    \end{subfigure}
    \begin{subfigure}{0.35\textwidth}
    \centering
        \includegraphics[width=1\textwidth]{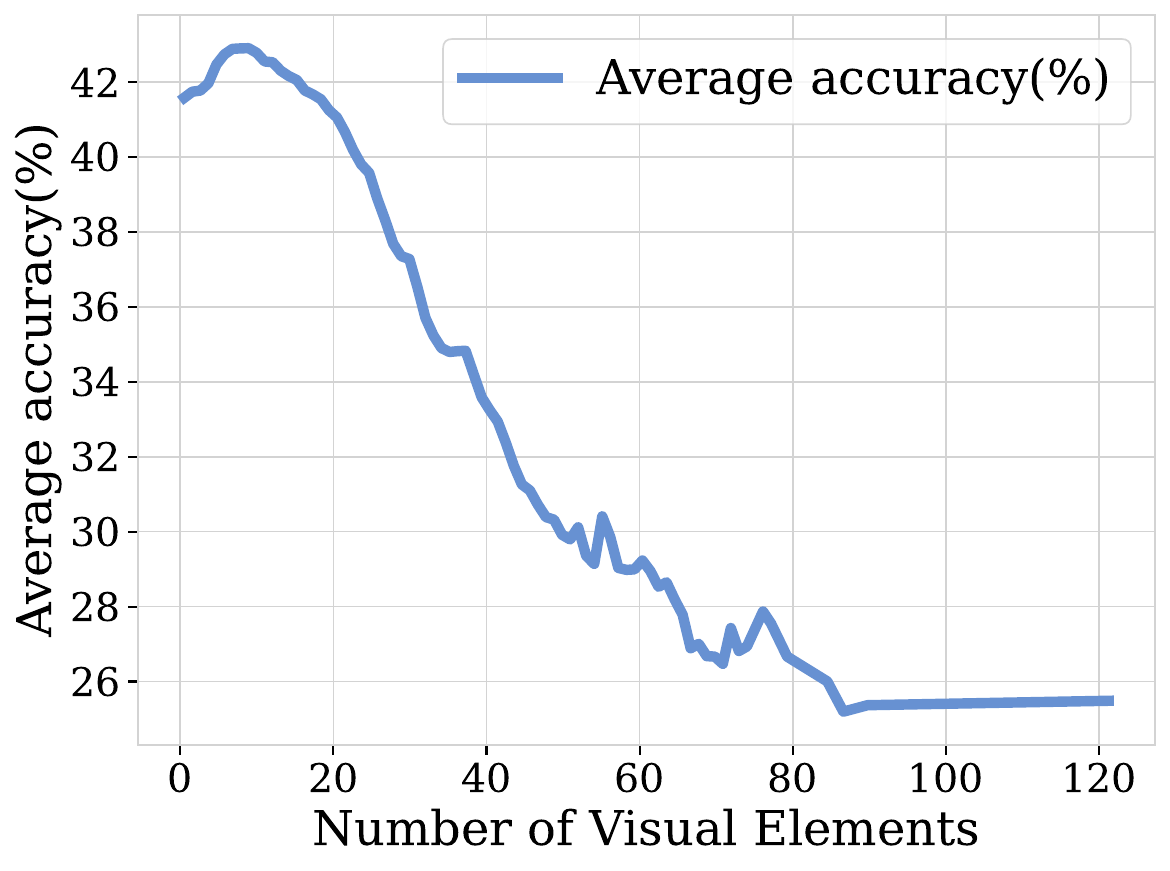}
        \caption{TinyChart}
    \end{subfigure}
    \caption{Average accuracy with respect to the number of \fix{visual} elements.}
    \label{fig:app_acc_num_images}
\end{figure}

We calculate the average accuracy of GPT-4.1 and TinyChart with respect to the number of \fix{visual} elements, as shown in Figure~\ref{fig:app_acc_num_images} (Left: GPT-4.1, right: TinyChart). 
Both models exhibit a clear performance drop as the number of visual elements increases. For example, GPT-4.1's accuracy decreases by more than $10\%$ once the number of such elements exceeds $100$, and a similar trend is observed in TinyChart. 

\xie{
To better understand which elements contribute most to this degradation, we further analyze the effect of element removal strategies. 
Specifically, elements are removed either in descending order of size---since larger elements typically carry more prominent information---or in a randomly shuffled order. 
As shown in Table~\ref{tab:app_removal_order}, both strategies lead to similar degradation trends, suggesting that the removal order plays only a minor role. 
}
\xie{
Building on these observations, we further evaluate models on de-decorated infographic charts, where \fix{visual} elements are removed while preserving the core chart structure.
We test GPT-4.1, Gemini-2.5 Pro Preview, TinyChart, and Qwen2.5-VL-72B under these settings and compare their performance against both infographic and plain charts. 
The results in Table~\ref{tab:app_barebone} align with the findings in Section~\ref{sec:4.3.1}, further confirming that removing unnecessary visual elements consistently improves model performance. 
}

\begin{table}[ht]
\centering
\caption{GPT-4.1's performance under different removal orders and numbers of visual elements.}
\resizebox{1\textwidth}{!}{
\normalsize
\begin{tabular}{lccccccccc}
\toprule
\textbf{Removal order / \# of visual elements} & \textbf{0} & \textbf{5} & \textbf{10} & \textbf{15} & \textbf{20} & \textbf{25} & \textbf{30} & \textbf{35} & \textbf{40} \\
\midrule
By Size & 87.27 & 82.55 & 80.22 & 79.42 & 78.34 & 79.35 & 79.14 & 78.84 & 77.39 \\
Random  & 87.27 & 82.11 & 80.56 & 78.99 & 78.87 & 78.99 & 78.12 & 78.01 & 77.39 \\
\bottomrule
\end{tabular}
}
\label{tab:app_removal_order}
\end{table}

\begin{table}[h]
\centering
\caption{Performance comparison on infographic, de-decorated infographic, and plain charts.}
\resizebox{0.8\textwidth}{!}{
\begin{tabular}{lccc}
\toprule
\textbf{Model} & \textbf{Infographic Chart} & \textbf{De-decorated Infographic Chart} & \textbf{Plain Chart} \\
\midrule
GPT-4.1                 & 77.39 & 82.27 & 83.05 \\
Gemini-2.5 Pro Preview  & 81.31 & 90.36 & 91.28 \\
TinyChart               & 26.32 & 41.72 & 42.94 \\
Qwen2.5-VL-72B          & 60.51 & 76.34 & 79.32 \\
\bottomrule
\end{tabular}
}
\label{tab:app_barebone}
\end{table}

\subsection{Interface for Human Answering}\label{app:interface}
\begin{figure}[!h]
    \centering
    \includegraphics[width=0.8\linewidth]{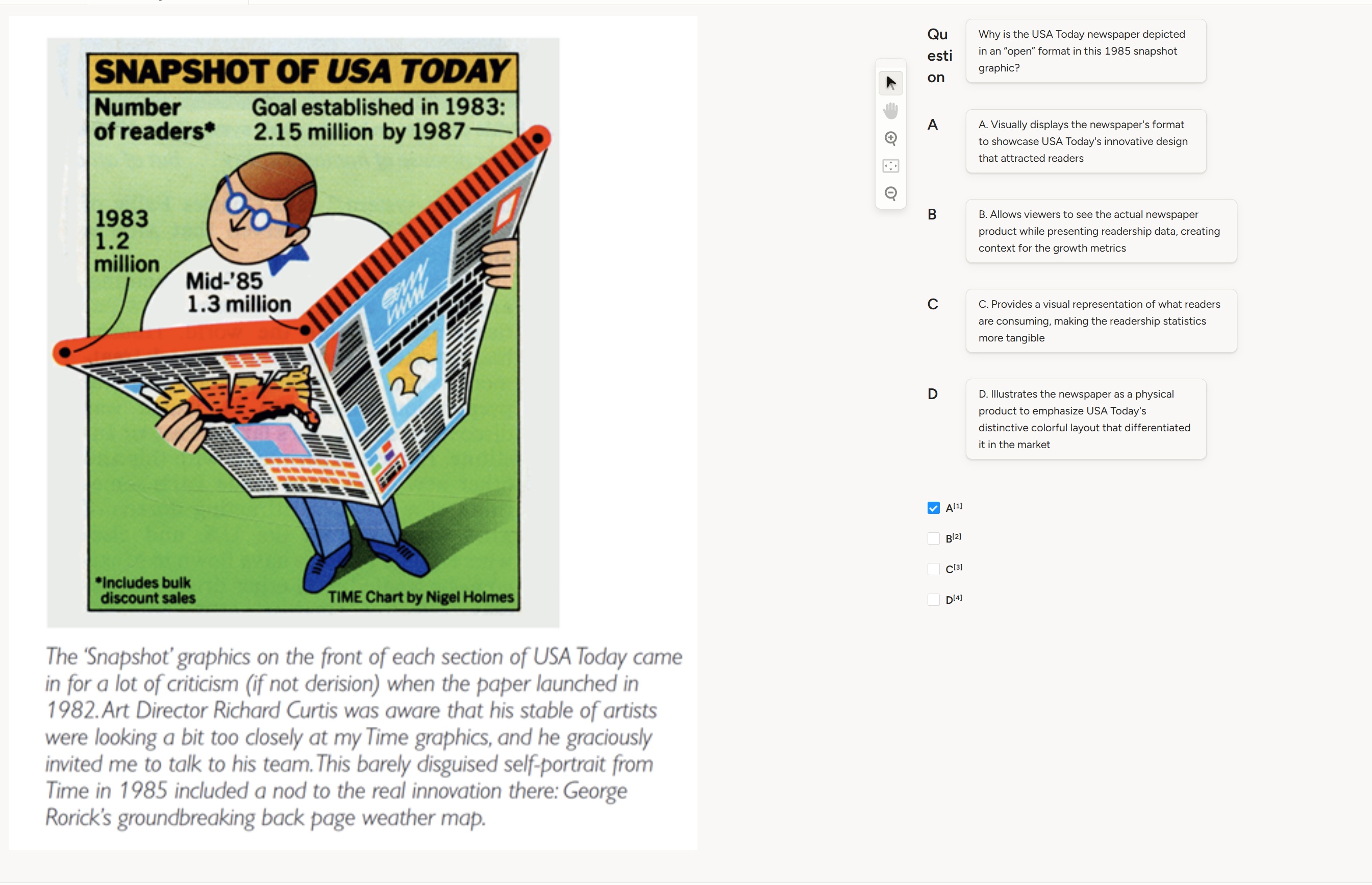}
    \caption{The interface for human answering.}
    \label{fig:app_interface}
\end{figure}

We provide a screenshot of the interface for human answering (Figure~\ref{fig:app_interface}).

\xie{
\subsection{More Detailed Information for Human Evaluation}\label{app:humanevalinfo}
We provide more detailed information for human evaluation.
Our human evaluation involved 15 participants, including 14 males and 1 female, all of whom are native Chinese speakers. They were graduate students (aged 20–30) from universities in China, with expertise in deep learning and visualization. 
None of the participants is the author of this paper.
}

\xie{
\subsection{Model Improvement Based on the Performance Degradation Analysis}
\label{app:supp_model_improvement}
In Sec. 4.3.1, we found that the visual elements primarily contributed to the infographic chart degradation. Therefore, we revised the prompt instruction to explicitly guide the model to focus on visualization components rather than decorative elements:
}

\textit{Instruction prompting: "Hint: Please be aware that this is an infographic chart. To obtain a more accurate answer, do not be influenced by its decorative elements; ignore such elements and focus your attention on the data visualization components."}

\xie{
We conducted this experiment on GPT-4.1 with 20\% of the dataset for quick evaluation. Table \ref{tab:prompting_comparison} shows that performance improves when such instructions are added.
}

\begin{table}[ht]
\centering
\caption{Performance comparison of GPT-4.1 under different prompting strategies.}
\resizebox{0.5\linewidth}{!}{
\normalsize
\begin{tabular}{lcc}
\toprule
\textbf{Model} & \textbf{Vanilla prompting} & \textbf{Instruction prompting} \\
\midrule
GPT-4.1 & 70.97 & 73.90 \\
\bottomrule
\end{tabular}
}
\label{tab:prompting_comparison}
\end{table}

\xie{
\subsection{Finetuning Evaluation}
We randomly selected 30k questions from the synthetic portion as the training set and used the remaining questions as the test set. The training set was then used to fine-tune the LLaVA OneVision Chat 7B model with LoRA. The experimental results show that its performance on text-based questions improved from the original 38.41\% to 50.19\% on the test set, which demonstrates the usefulness of the synthetic portion.
}

\subsection{Sample Questions and Answers}
We provide sample answers for text-based~\ref{sssec:text_sample}, visual-element-based basic~\ref{sssec:visual_basic_sample} and metaphor~\ref{sssec:visual_metaphor_sample} questions in this section.
\newpage
\subsubsection{Sample Answers of Text-based Questions}\label{sssec:text_sample}
\tcbset{
    mybox/.style={
        colback=red!5, 
        colframe=red, 
        left=10pt, 
        right=10pt, 
        top=10pt, 
        bottom=10pt, 
        arc=2pt, 
        boxrule=0pt, 
        borderline west={4pt}{0pt}{red}, 
        fonttitle=\bfseries, 
        title=#1 
    }
}

\begin{tcolorbox}[colframe=blue,colback=cyan!20,title=Example of text-based questions]
   \begin{minipage}{0.48\textwidth}
        \centering
        \includegraphics[width=\linewidth,height=200pt]{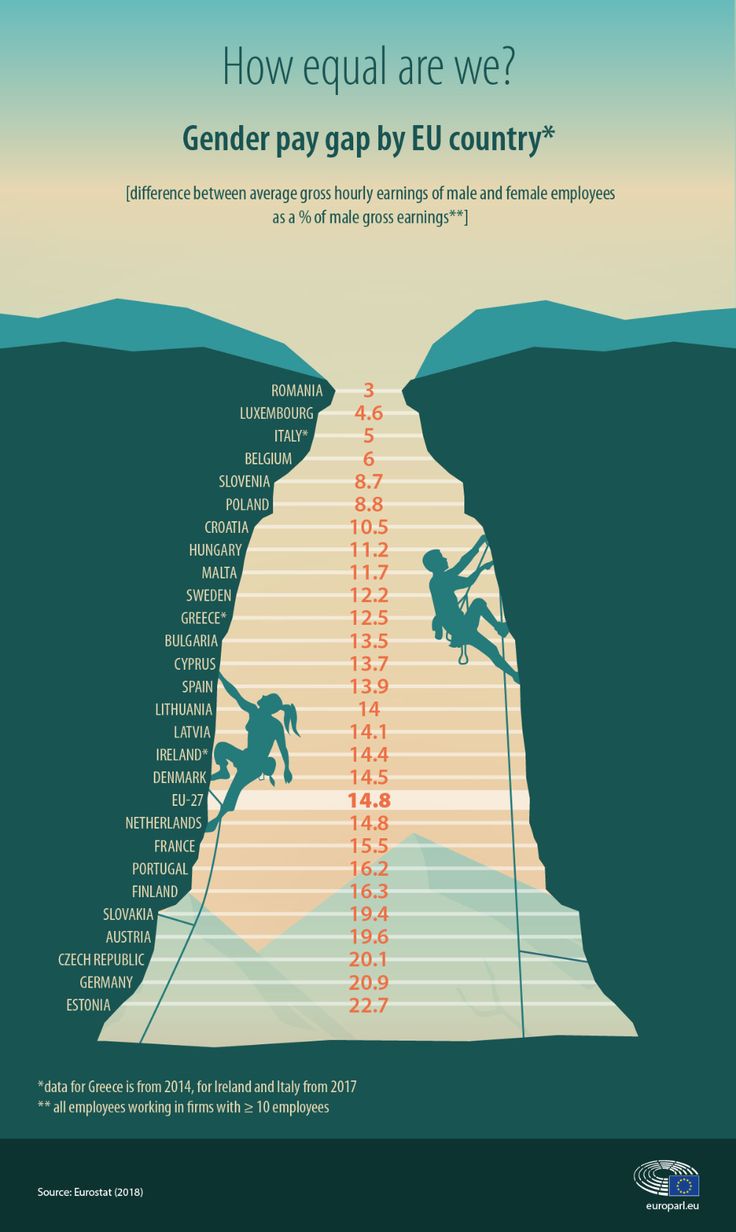}\\
    \end{minipage}
    \hfill 
    \begin{minipage}{0.48\textwidth}
        \centering
        \includegraphics[width=\linewidth,height=200pt]{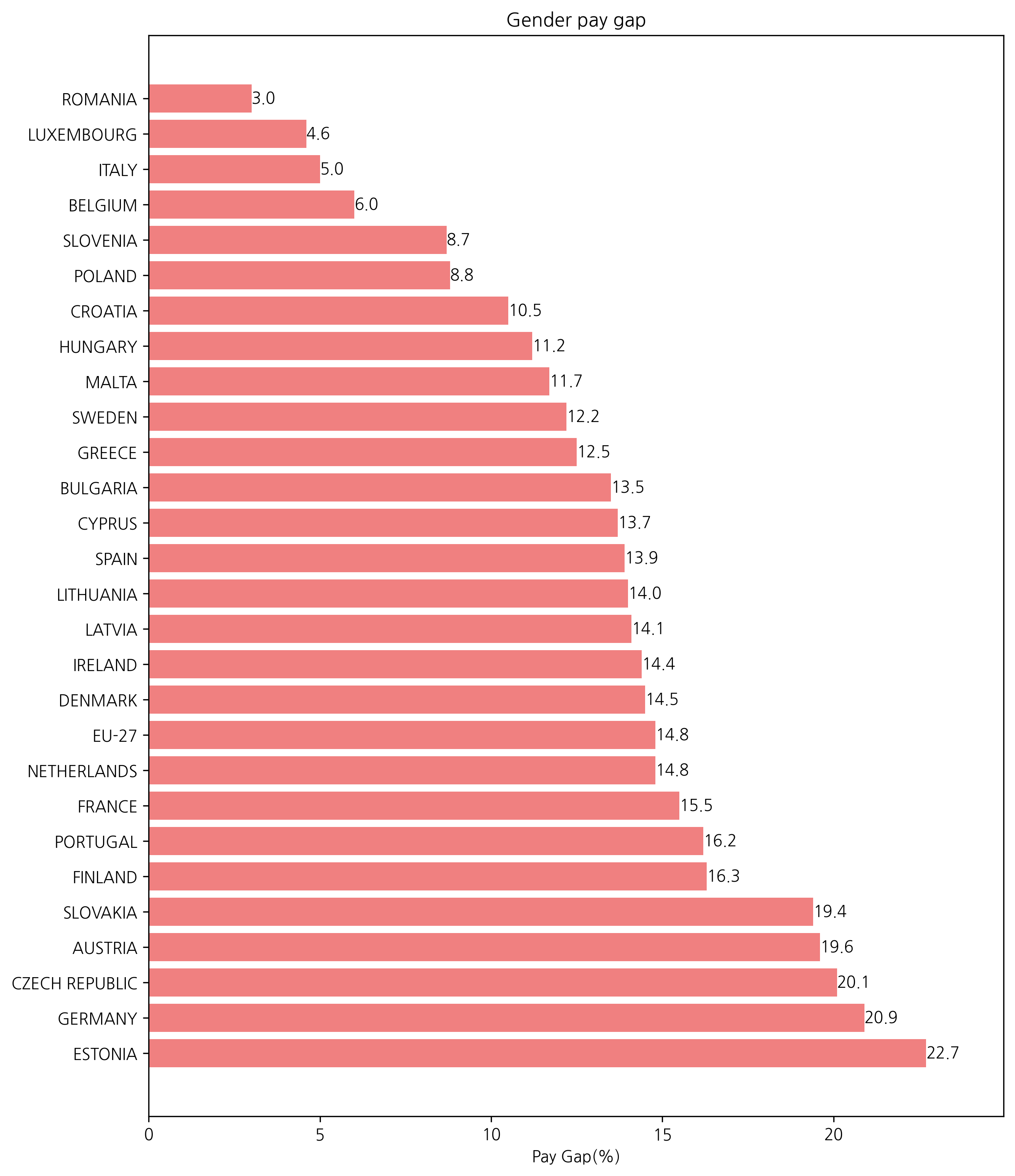}\\
    \end{minipage}
    
    \vspace{15pt}
    \textcolor{blue!60}{\textbf{Question:} }
    \\
    {\setlength{\spaceskip}{0.3em} \ttfamily By how much does the Pay Gap(\%) of MALTA differ from that of POLAND?

    \vspace{10pt}
    * If the Pay Gap(\%) of MALTA is less than the Pay Gap(\%) of POLAND, your response should be Yes, otherwise 'No'.

    \vspace{10pt}
    * Your response should only contain 'Yes' or 'No'.}

    \vspace{10pt}
    \textcolor{blue!60}{\textbf{Answer:} }{\ttfamily No}
\end{tcolorbox}
\begin{tcolorbox}[mybox=Model's answers]
    \textbf{o4-mini:} \textcolor{red}{Yes} (Info) 
    
    \qquad\qquad No (Plain)
    
    \textbf{Gemini 2.5 pro preview:} No (Info) 
    
    \hspace{107pt} No (Plain)
    
    \textbf{InternVL 3 78B:} \textcolor{red}{Yes} (Info) 
    
    \hspace{74pt}No (Plain)
    
    \textbf{TinyChart:} No (Info) 
    
    \hspace{52pt} No (Plain)
\end{tcolorbox}
\newpage

\begin{tcolorbox}[colframe=blue,colback=cyan!20,title=Example of text-based questions]
   \begin{minipage}{0.48\textwidth}
        \centering
        \includegraphics[width=\linewidth,height=200pt]{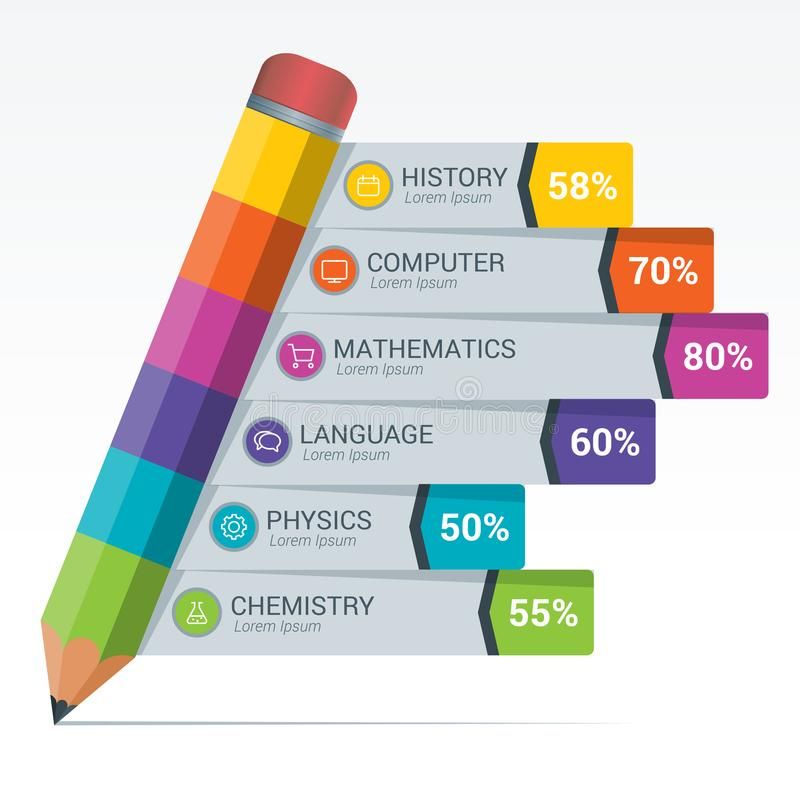}\\
    \end{minipage}
    \hfill 
    \begin{minipage}{0.50\textwidth}
        \centering
        \includegraphics[width=\linewidth,height=150pt]{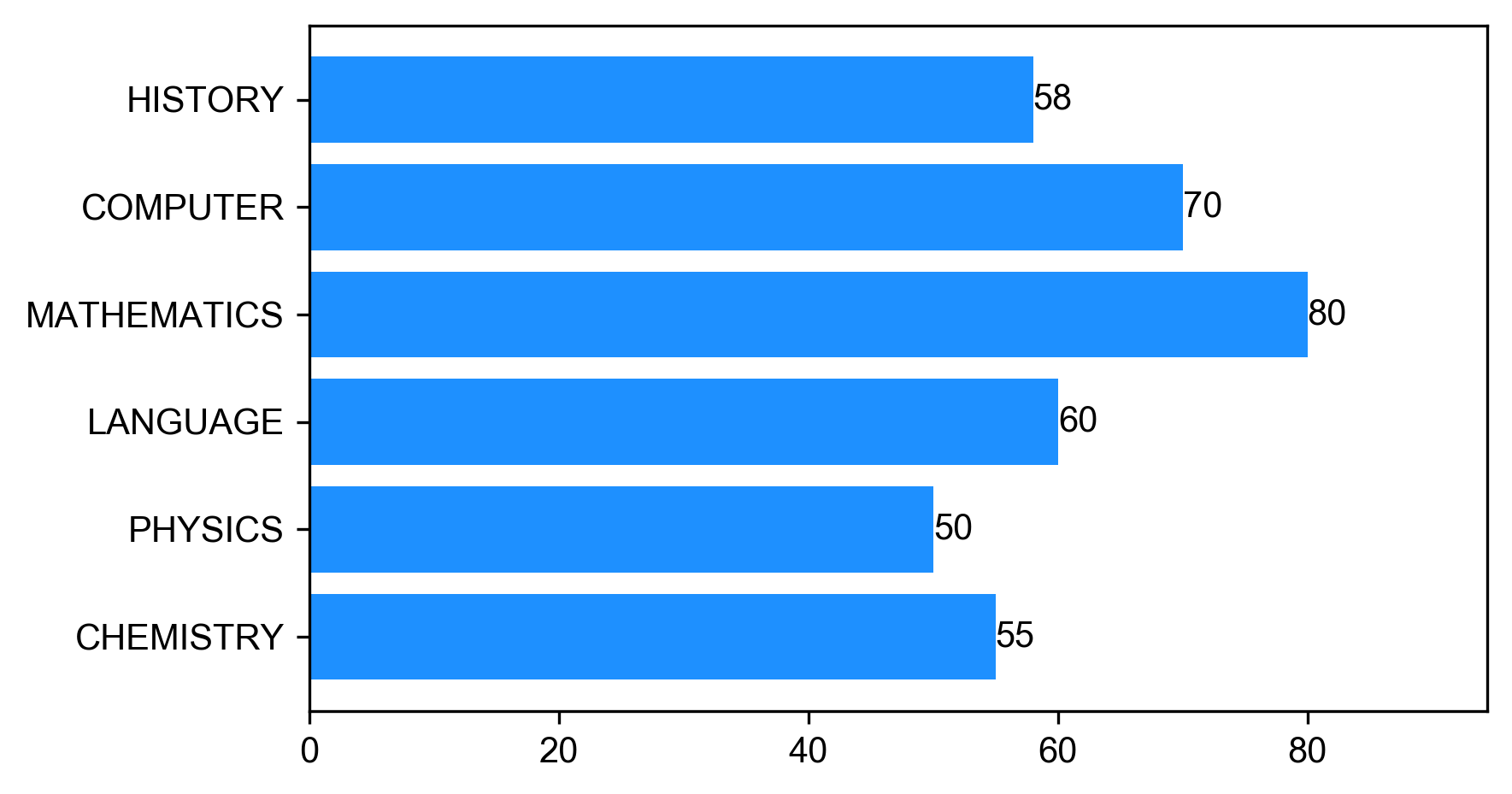}\\
    \end{minipage}
    
    \vspace{15pt}
    \textcolor{blue!60}{\textbf{Question:} }
    \\
    {\setlength{\spaceskip}{0.3em} \ttfamily What is the result if we sum up all Percentage(\%)?

    \vspace{10pt}
    * Your response should only contain the value of the sum of Percentage(\%).}
    
    \vspace{10pt}
    \textcolor{blue!60}{\textbf{Answer:}}{\ttfamily 373}
\end{tcolorbox}
\begin{tcolorbox}[mybox=Model's answers]
    \textbf{o4-mini:} 373 (Info) 
    
    \qquad\qquad373 (Plain)
    
    \textbf{Gemini 2.5 pro preview:} 373 (Info) 
    
    \hspace{106pt}373 (Plain)
    
    \textbf{InternVL 3 78B:} \textcolor{red}{383} (Info) 
    
    \hspace{74pt}373 (Plain)
    
    \textbf{TinyChart:} \textcolor{red}{85\%} (Info) 
    
    \hspace{51pt}373 (Plain)
\end{tcolorbox}
\newpage

\begin{tcolorbox}[colframe=blue,colback=cyan!20,title=Example of text-based questions]
   \begin{minipage}{0.48\textwidth}
        \centering
        \includegraphics[width=\linewidth,height=160pt]{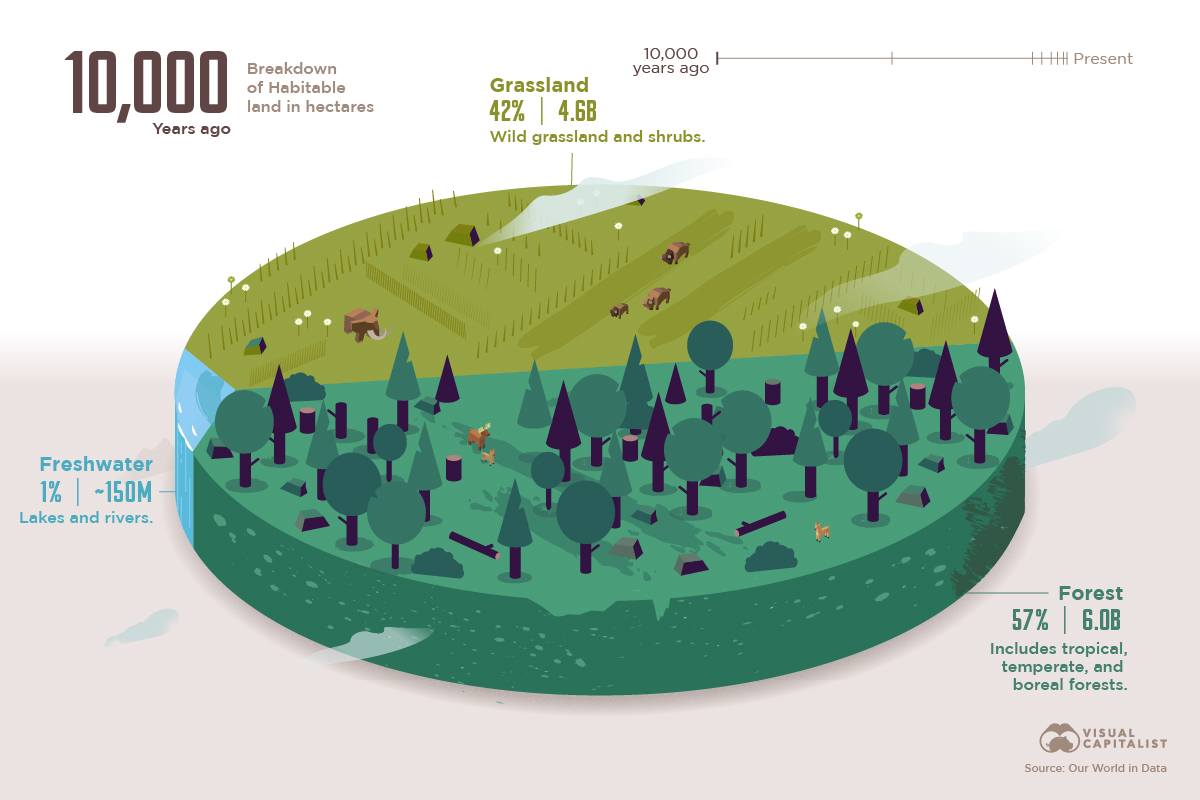}\\
    \end{minipage}
    \hfill 
    \begin{minipage}{0.48\textwidth}
        \centering
        \includegraphics[width=\linewidth,height=110pt]{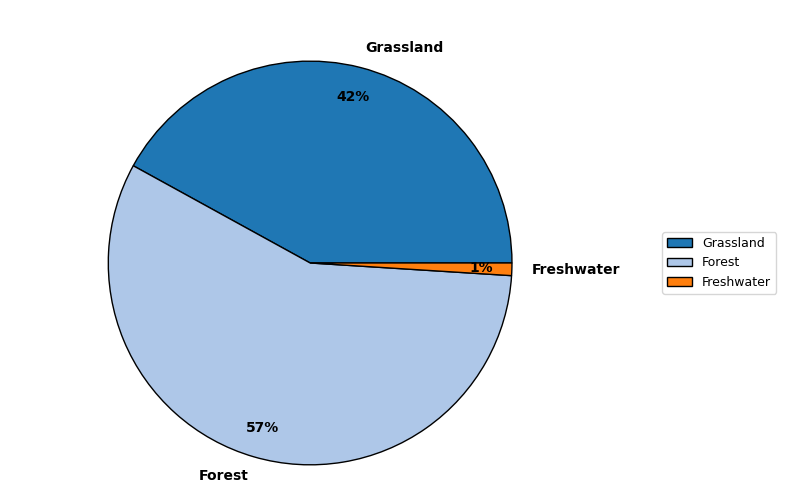}\\
    \end{minipage}
    
    \vspace{15pt}
    \textcolor{blue!60}{\textbf{Question:} }\\
    {\setlength{\spaceskip}{0.3em} \ttfamily What is the value difference between the Freshwater and Forest percentages?

    \vspace{10pt}
    * Your response should only contain the value of the difference between the Percentage corresponding to Freshwater and the Percentage corresponding to Forest.

    \vspace{10pt}
    * The answer you give me should be the absolute value.  
    }

    \vspace{10pt}
    \textcolor{blue!60}{\textbf{Answer:} }{\ttfamily 56\%}
\end{tcolorbox}
\begin{tcolorbox}[mybox=Model's answers]
    \textbf{o4-mini:} 56 (Info) 
    
    \qquad\qquad56 (Plain)
    
    \textbf{Gemini 2.5 pro preview:} 56 (Info) 
    
    \hspace{107pt}56 (Plain)

    \textbf{InternVL 3 78B:} 56\% (Info) 
    
    \hspace{74pt}56\% (Plain)
    
    \textbf{TinyChart:} \textcolor{red}{10.0} (Info)
    
    \hspace{52pt}\textcolor{red}{46\%} (Plain)
\end{tcolorbox}
\newpage

\begin{tcolorbox}[colframe=blue,colback=cyan!20,title=Example of Text-based questions]
   \begin{minipage}{0.48\textwidth}
        \centering
        \includegraphics[width=\linewidth,height=200pt]{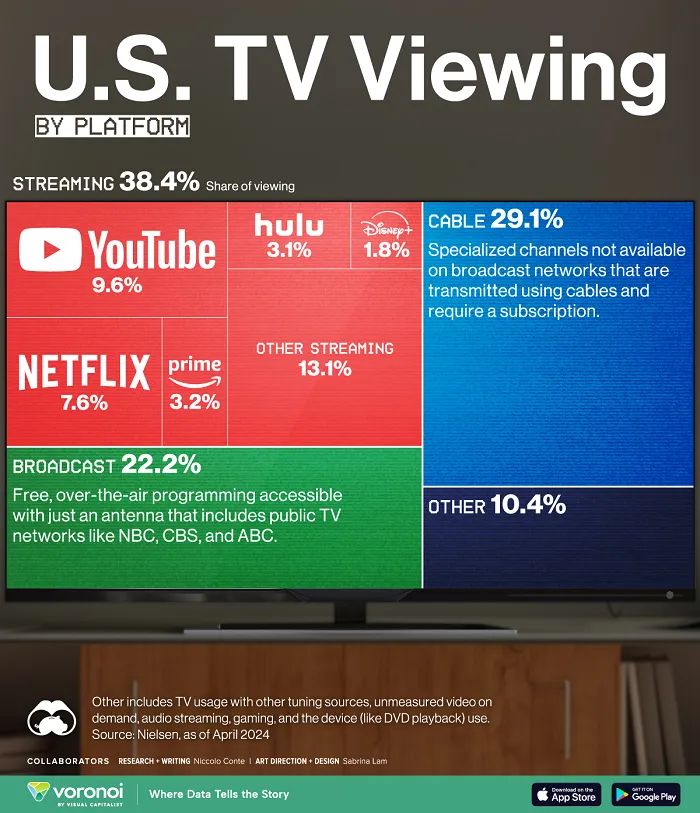}\\
    \end{minipage}
    \hfill 
    \begin{minipage}{0.48\textwidth}
        \centering
        \includegraphics[width=\linewidth,height=150pt]{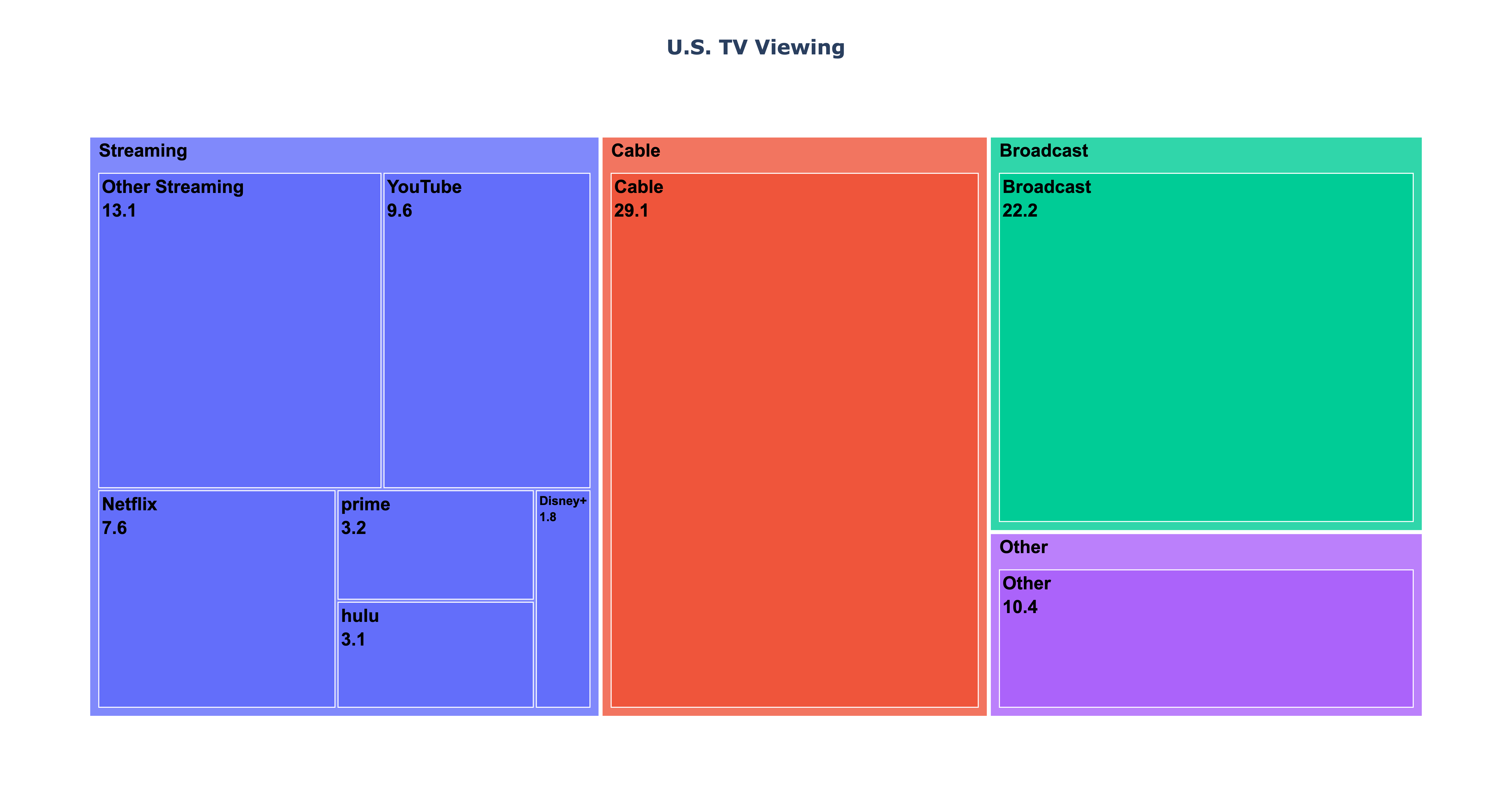}\\
    \end{minipage}
    
    \vspace{15pt}
    \textcolor{blue!60}{\textbf{Question:} }\\{\setlength{\spaceskip}{0.3em} \ttfamily What are the nodes categorized as Streaming?

    \vspace{10pt}
    * Your response should only contain the node is/in Streaming.

    \vspace{10pt}
    *Please provide your answer in the order from left to right, top to bottom, as they appear in the chart.
    
    \vspace{10pt}
    *If there is no answer that meets the condition, respond with an empty string.
    }

    \vspace{10pt}
    \textcolor{blue!60}{\textbf{Answer:} }{\ttfamily YouTube, Netflix, prime, hulu, Disney+, Other Streaming}
\end{tcolorbox}
\begin{tcolorbox}[mybox=Model's answers]
    \textbf{o4-mini:} 
    
    YouTube, Netflix, prime, hulu, Disney+, Other Streaming (Info) 
    
    YouTube, Netflix, prime, hulu, Disney+, Other Streaming (Plain)
    
    \textbf{Gemini 2.5 pro preview:} 
    
    YouTube, Netflix, prime, hulu, Disney+, Other Streaming, \textcolor{red}{Streming} (Info)

    YouTube, Netflix, prime, hulu, Disney+, Other Streaming (Plain)

    \textbf{InternVL 3 78B:} YouTube, Netflix, Hulu, Prime, Other Streaming (Info) 
    
    YouTube, Netflix, prime, hulu, Disney+, Other Streaming (Plain)
    
    \textbf{TinyChart:} 
    
    Streaming (Info) 
    
    Streaming (Plain)
\end{tcolorbox}
\newpage

\begin{tcolorbox}[colframe=blue,colback=cyan!20,title=Example of text-based questions]
   \begin{minipage}{0.48\textwidth}
        \centering
        \includegraphics[width=\linewidth,height=200pt]{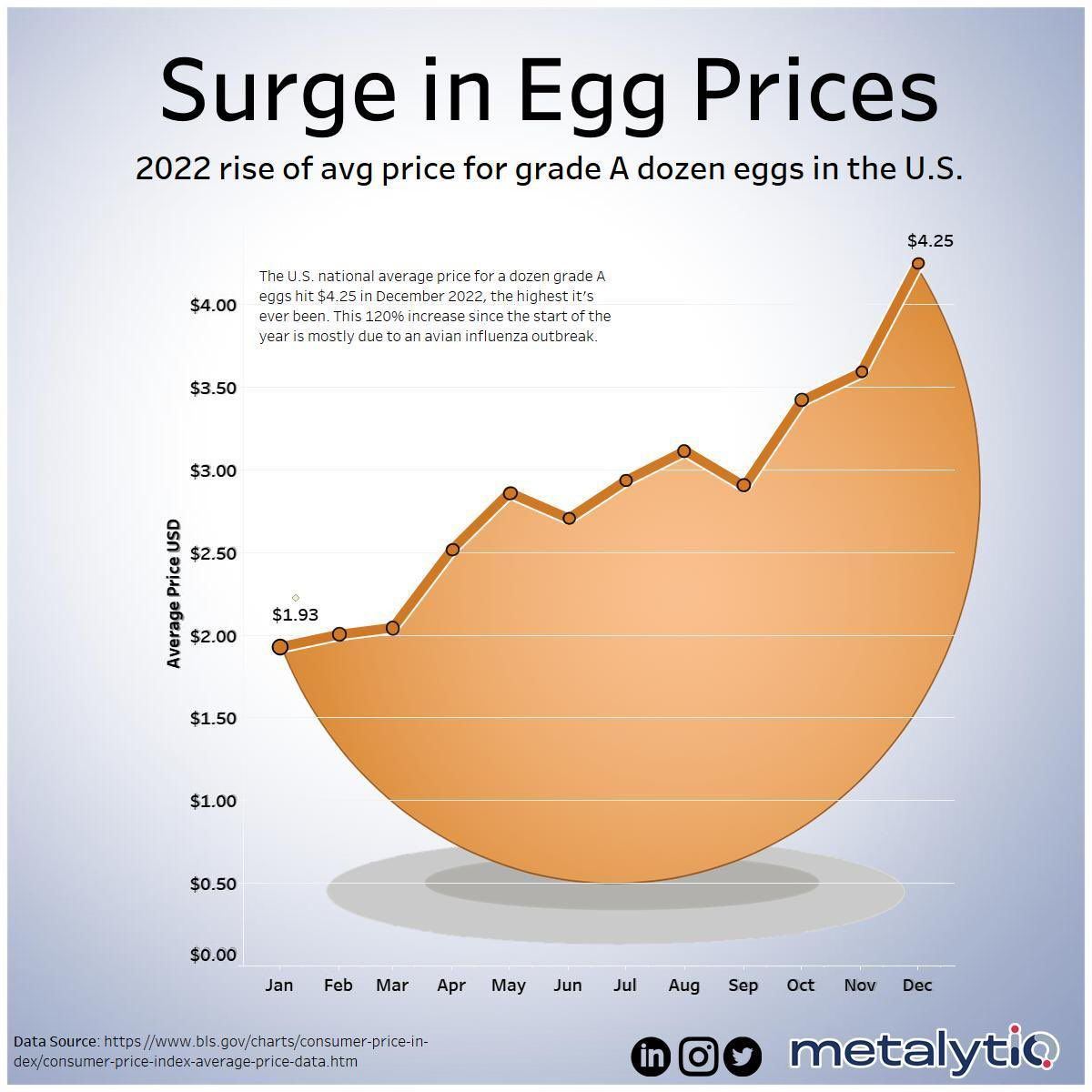}\\
    \end{minipage}
    \hfill 
    \begin{minipage}{0.48\textwidth}
        \centering
        \includegraphics[width=\linewidth,height=180pt]{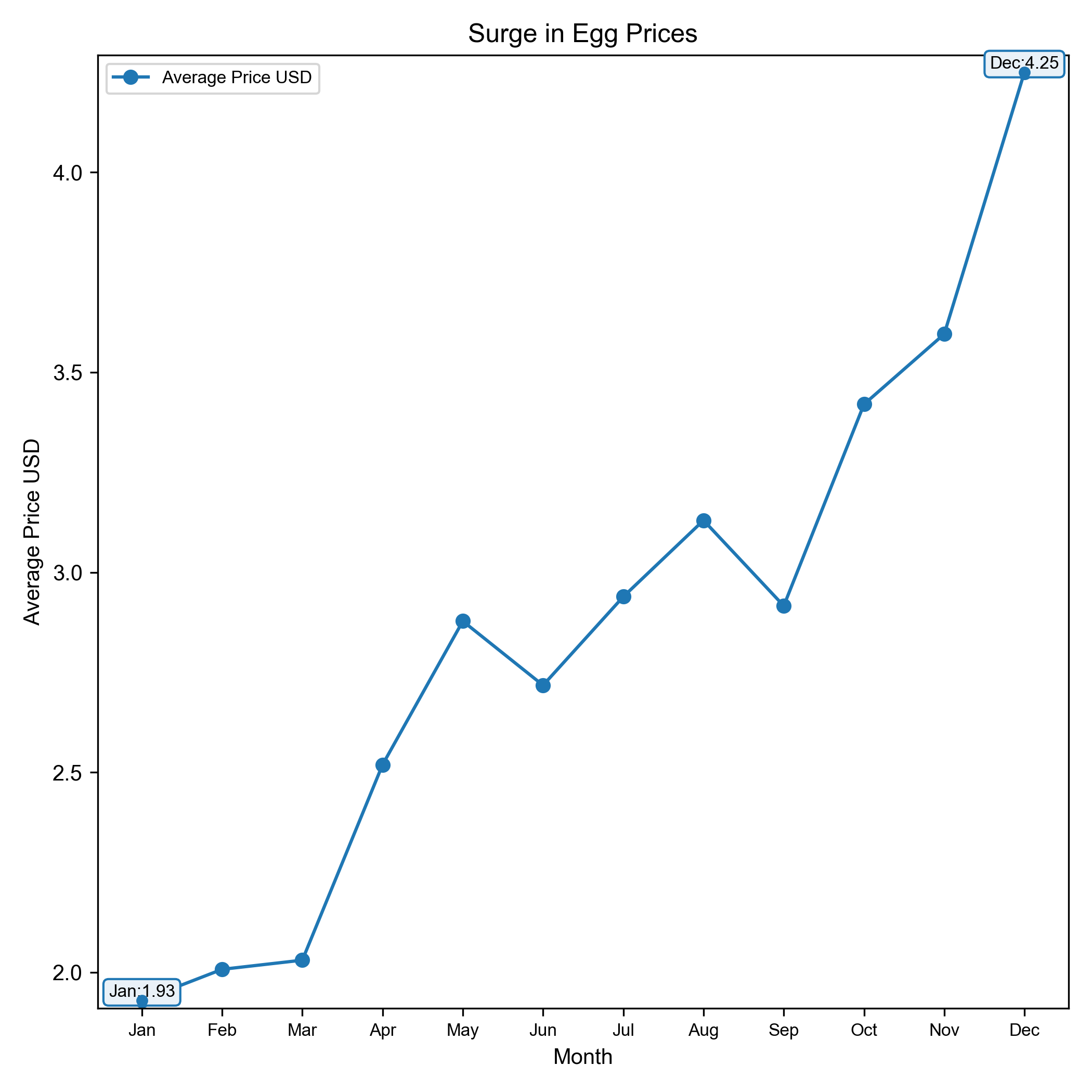}\\
    \end{minipage}
    
    \vspace{15pt}
    \textcolor{blue!60}{\textbf{Question:} }\\
    {\setlength{\spaceskip}{0.3em} \ttfamily What was the Average Price in USD during December?

    \vspace{10pt}
    * Your response should only contain the value of Average Price USD corresponding to Average Price USD in/on/at Dec.

    \vspace{10pt}
    * If there is an explicit answer in the chart, answer in exactly the same format.
    }

    \vspace{10pt}
    \textcolor{blue!60}{\textbf{Answer:} }{\ttfamily 4.25}
\end{tcolorbox}
\begin{tcolorbox}[mybox=Model's answers]
    \textbf{o4-mini:} 4.25 (Info) 
    
    \qquad\qquad4.25 (Plain)
    
    \textbf{Gemini 2.5 pro preview:} 4.25 (Info) 
    
    \hspace{107pt}4.25 (Plain)

    \textbf{InternVL 3 78B:} 4.25 (Info) 
    
    \hspace{74pt}4.25 (Plain)
    
    \textbf{TinyChart:} 4.25 (Info) 
    
    \hspace{52pt}4.25 (Plain)
\end{tcolorbox}
\newpage
\subsubsection{Sample Answers of Visual-element-based (Basic) Questions}\label{sssec:visual_basic_sample}

\begin{tcolorbox}[colframe=blue,colback=cyan!20,title=Example of visual-element-based (basic) question]
   \begin{minipage}{0.48\textwidth}
        \centering
        \includegraphics[width=\linewidth,height=160pt]{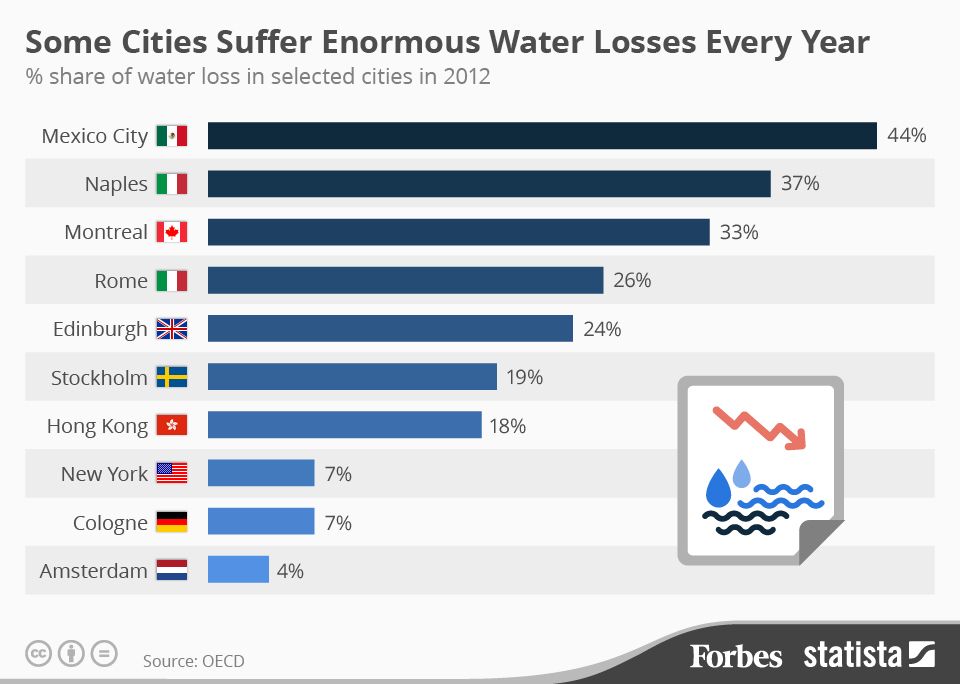}\\
    \end{minipage}
    \hspace{65pt} 
    \begin{minipage}{0.15\textwidth}
        \centering
        \includegraphics[width=\linewidth,height=80pt]{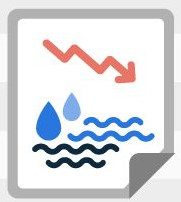}\\
    \end{minipage}
    
    \vspace{15pt}
    \textcolor{blue!60}{\textbf{Question:} }\\{\setlength{\spaceskip}{0.3em} \ttfamily Beginning with a chart and followed by its cropped segment, the images above are arranged from left to right. What is the primary reason for incorporating this imagery into the chart?

    \vspace{10pt}
    (A) To visually emphasize the chart's central theme regarding the frequency and importance of media citations.

    \vspace{10pt}
    (B) To serve primarily as a decorative background element, enhancing the overall visual appeal of the infographic.
    
    \vspace{10pt}
    (C) To specifically highlight the data column showing the number of citations, aiding in data interpretation.
    
    \vspace{10pt}
    (D) To use symbolism (water representation, downward arrow) to represent the abstract concept of 'water loss'.
    
    \vspace{10pt}
    * Your response should be the letter only (e.g., 'C'). Do not include any explanation or repeat the option text.      
    
    \vspace{10pt}
    \textcolor{blue!60}{\textbf{Answer:} }{\ttfamily D}

    }
\end{tcolorbox}
\begin{tcolorbox}[mybox=Model's answers]
    \textbf{o4-mini:} D
    
    \textbf{Gemini 2.5 pro preview:} A
    
    \textbf{InternVL 3 78B:} A
    
    \textbf{TinyChart:} 0.7
\end{tcolorbox}
\newpage

\subsubsection{Sample Answers of Visual-element-based (Metaphor) Questions}\label{sssec:visual_metaphor_sample}
\begin{tcolorbox}[colframe=blue,colback=cyan!20,title=Example of visual-element-based (metaphor) question]
   \begin{minipage}{0.48\textwidth}
        \centering
        \includegraphics[width=\linewidth,height=200pt]{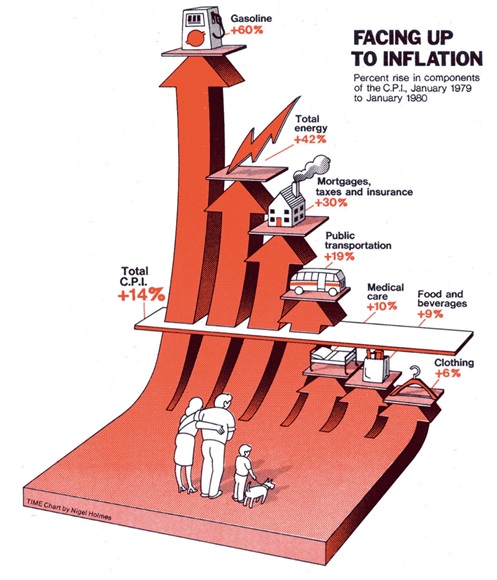}\\
    \end{minipage}
    \hfill 
    \begin{minipage}{0.48\textwidth}
        \centering
        \includegraphics[width=\linewidth,height=200pt]{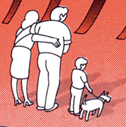}\\
    \end{minipage}
    
    \vspace{15pt}
    \textcolor{blue!60}{\textbf{Question:} }\\{\setlength{\spaceskip}{0.3em} \ttfamily Answer this question and choose the most appropriate answer (A, B, C, D).
    
    \vspace{10pt}
    According to Figure 1, the overall visual presentation aims to convey which idea, especially by drawing the figures in Figure 2:

    \vspace{10pt}
    A) The challenge of the increasing inflation needs to be addressed.
    
    \vspace{10pt}
    B) Addressing the increasing inflation needs to effort of families.
    
    \vspace{10pt}
    C) The total inflation of C.P.I is growing higher and higher over time.
    
    \vspace{10pt}
    D) High inflation is becoming a challenge for people.

    \vspace{10pt}
    
    \qquad\qquad* Your response should be the letter only (e.g., 'C'). Do not include any explanation or repeat the option text.
    }
    \vspace{10pt}
    
    \textcolor{blue!60}{\textbf{Answer:} }{\ttfamily A}
\end{tcolorbox}
\begin{tcolorbox}[mybox=Model's answers]
    \textbf{o4-mini:} D
    
    \textbf{Gemini 2.5 pro preview:} (A) The challenge of the increasing inflation needs to be addressed.
    The title "FACING UP TO INFLATION" suggests confronting a problem, which implies recognizing its severity and the need to deal with it. 
    ...
    The final answer is A.

    \textbf{InternVL 3 78B:} D
    
    \textbf{TinyChart:} A, B, C, D
\end{tcolorbox}
\newpage

\section{Licenses}\label{append:license}

In this section, we provide licenses for data and models used in our experiments.

\subsection{Licenses for Data}
 \xie{We provide the data licenses of each platform in Table~\ref{tab:app_data_license}. Since some platforms adopt license policies that may restrict data redistribution, we release our dataset in the form of URLs to mitigate copyright risks.}
\begin{table}[!hbtp]
  \centering
  \caption{Data license.}
    \begin{tabular}{lcc}
    \toprule
    \textbf{Platform} & \textbf{Data License} \\
    \midrule
    Pinterest &  \href{https://help.pinterest.com/en/article/content-rights-management}{Custom}\\
    Visual Capitalist & \href{https://licensing.visualcapitalist.com/}{Custom} \\
    Statista &  \href{https://www.statista.com/getting-started/publishing-statista-content-terms-of-use-and-publication-rights}{Custom} \\
    iStock  & \href{https://www.istockphoto.com/legal/license-agreement}{Custom}  \\
    Behance & \href{https://help.behance.net/hc/en-us/articles/7416981176603-License-Types-for-Assets}{Custom}\\
    PNGEgg & \href{https://www.pngegg.com/tos}{Custom}    \\
    information is beautiful& \href{https://informationisbeautiful.net/licensing/}{Custom}\\
    coolinfographics &  Not Applicable\\
    designweek & \href{https://www.designweek.co.uk/terms/}{Custom}\\
    marketing-partners & \href{https://www.marketing-partners.com/faqs}{Custom}\\
    datavizproject & Not Applicable\\
    ChartGalaxy & Apache-2.0 license\\
    \bottomrule
   
    \end{tabular}%
  \label{tab:app_data_license}%
\end{table}%

\subsection{Licenses for Model}
\begin{table}[!tbhp]
  \centering
  \caption{Model license.}
    \begin{tabular}{lcc}
    \toprule
    \textbf{Model} & \textbf{Model License} & \textbf{Code Licenses}\\
    \midrule
    OpenAI O4-mini &  Proprietary &    Proprietary \\
    GPT-4.1 &  Proprietary  &  Proprietary\\
    GPT-4o &  Proprietary &  Proprietary \\
    Claude 3.5 Sonnet  & Proprietary  &  Proprietary \\
    Gemini 2.5 Pro Preview &  Proprietary  &  Proprietary \\
    Gemini 2.5 Flash Preview  &  Proprietary    &  Proprietary \\
    Qwen2.5-VL-72B & \href{https://huggingface.co/Qwen/Qwen2.5-VL-72B-Instruct/blob/main/LICENSE}{Custom}  &  Apache 2.0 License \\
    Llama-4 Scout & \href{https://github.com/meta-llama/llama-models/blob/main/models/llama4/LICENSE}{Custom} & \href{https://github.com/meta-llama/llama-models/blob/main/models/llama4/LICENSE}{Custom} \\
    Intern-VL3 78B & MIT  & MIT \\
    Intern-VL3 8B & MIT & MIT \\
    Janus Pro & \href{https://github.com/deepseek-ai/DeepSeek-LLM/blob/HEAD/LICENSE-MODEL}{Custom} & MIT \\
    DeepSeek VL2 & \href{https://github.com/deepseek-ai/deepseek-vl2/blob/main/LICENSE-MODEL}{Custom}  & MIT \\
    Phi-4 & MIT & MIT \\
    LLaVA OneVision Chat 72B  &  Apache 2.0 &  Apache 2.0 \\
    LLaVA OneVision Chat 7B &  Apache 2.0 &  Apache 2.0 \\
    Pixtral & Apache-2.0 &Apache-2.0 \\
    Ovis1.6-Gemma2-9B & Apache-2.0  & Apache-2.0 \\
    ChartGemma &  GPL-3.0 & MIT\\
    TinyChart & MIT & MIT \\
    ChartInstruct-LLama2 & GPL-3.0 & MIT \\
    \bottomrule
    \end{tabular}%
  \label{tab:app_model_license}%
\end{table}%


\end{document}